\documentclass[times,twocolumn,final]{elsarticle}
\usepackage{medima_arxiv}
\usepackage{framed,multirow}
\usepackage{amssymb}
\usepackage{latexsym}
\usepackage{url}
\usepackage{cite}
\usepackage{amsmath,amssymb,amsfonts,bm}
\usepackage{booktabs}
\usepackage{arydshln}
\usepackage{tabulary}
\usepackage{graphicx,epstopdf}
\usepackage{textcomp}
\usepackage{subcaption}
\usepackage{booktabs}
\usepackage[svgnames,table]{xcolor}
\usepackage{pifont}
\usepackage{placeins}
\usepackage{fancyhdr}

\usepackage{arydshln}
\usepackage[colorlinks=true,citecolor=cyan,urlcolor=cyan]{hyperref}

\fancypagestyle{firstpagestyle}{
    \fancyhf{} 
    \fancyhead{} 
    \fancyfoot{} 
    \fancyhead[CO]{\em \fontsize{9pt}{8pt}\selectfont This article has been accepted to the Medical Image Analysis Journal, 2025.}
}

\fancypagestyle{default}{
    \fancyhf{}
    \fancyhead[R]{\thepage} 
    \fancyhead[LO]{\em \fontsize{9pt}{8pt}\selectfont This article has been accepted to the International Conference on Information Processing in Computer-Assisted Interventions, 2024.}
}

\definecolor{newcolor}{rgb}{.8,.349,.1}
\newcommand{\triplet}[1]{\textlangle{\textit{#1}}\textrangle{}}

\begin{document}


\begin{frontmatter}

\title{Learning multi-modal representations by watching hundreds of surgical video lectures}%

\author[1,3]{Kun \snm{Yuan}}
\author[1,2]{Vinkle \snm{Srivastav}}
\author[1]{Tong \snm{Yu}}
\author[2,4]{Jo\"el L. \snm{Lavanchy}}
\author[6]{Jacques \snm{Marescaux}}
\author[2,5]{Pietro \snm{Mascagni}}
\author[3]{Nassir \snm{Navab}}
\author[1,2]{Nicolas \snm{Padoy}}
\cortext[cor1]{Corresponding author: 
  Tel.: +33-390413530}
\ead{npadoy@unistra.fr}

\address[1]{University of Strasbourg, CNRS, INSERM, ICube, UMR7357, Strasbourg, France}
\address[2]{IHU Strasbourg, Strasbourg, France}
\address[3]{CAMP, Technische Universit\"at M\"unchen, Munich, Germany}
\address[4]{University Digestive Health Care Center – Clarunis, 4002 Basel, Switzerland}
\address[5]{Fondazione Policlinico Universitario A. Gemelli IRCCS, Rome, Italy}
\address[6]{IRCAD, Strasbourg, Strasbourg, France}

\received{XXX}
\finalform{XXX}
\accepted{XXX}
\availableonline{XXX}
\communicated{XXX}

\begin{abstract}
{Recent advancements in surgical computer vision applications have been driven by vision-only models, which do not explicitly integrate the rich semantics of language into their design.} These methods rely on manually annotated surgical videos to predict a fixed set of object categories, limiting their generalizability to unseen surgical procedures and downstream tasks. In this work, we put forward the idea that the surgical video lectures available through open surgical e-learning platforms can provide effective  {vision and language } supervisory signals for multi-modal representation learning without relying on manual annotations. We address the surgery-specific linguistic challenges present in surgical video lectures by employing multiple complementary automatic speech recognition systems to generate text transcriptions. We then present a novel method, \textit{SurgVLP} - Surgical Vision Language Pre-training, for multi-modal representation learning. \textit{SurgVLP} constructs a new contrastive learning objective to align video clip embeddings with the corresponding multiple text embeddings by bringing them together within a joint latent space. To effectively demonstrate the representational capability of the learned joint latent space, we introduce several vision-and-language surgical tasks and evaluate various vision-only tasks specific to surgery, e.g., surgical tool, phase, and triplet recognition.
{
Extensive experiments across diverse surgical procedures and tasks demonstrate that the multi-modal representations learned by \textit{SurgVLP} exhibit strong transferability and adaptability in surgical video analysis. Furthermore, our zero-shot evaluations highlight \textit{SurgVLP}’s potential as a general-purpose foundation model for surgical workflow analysis, reducing the reliance on extensive manual annotations for downstream tasks, and facilitating adaptation methods such as few-shot learning to build a scalable and data-efficient solution for various downstream surgical applications.} The code is available at \url{https://github.com/CAMMA-public/SurgVLP}.

\end{abstract}

\begin{keyword}
Multi-modal representation learning \sep Surgical video lectures\sep Self-supervision \sep Vision-and-language
\end{keyword}

\end{frontmatter}


\thispagestyle{firstpagestyle}

\section{Introduction}
\label{sec:introduction}

\begin{figure*}[!t]
\centerline{\includegraphics[width=2\columnwidth]{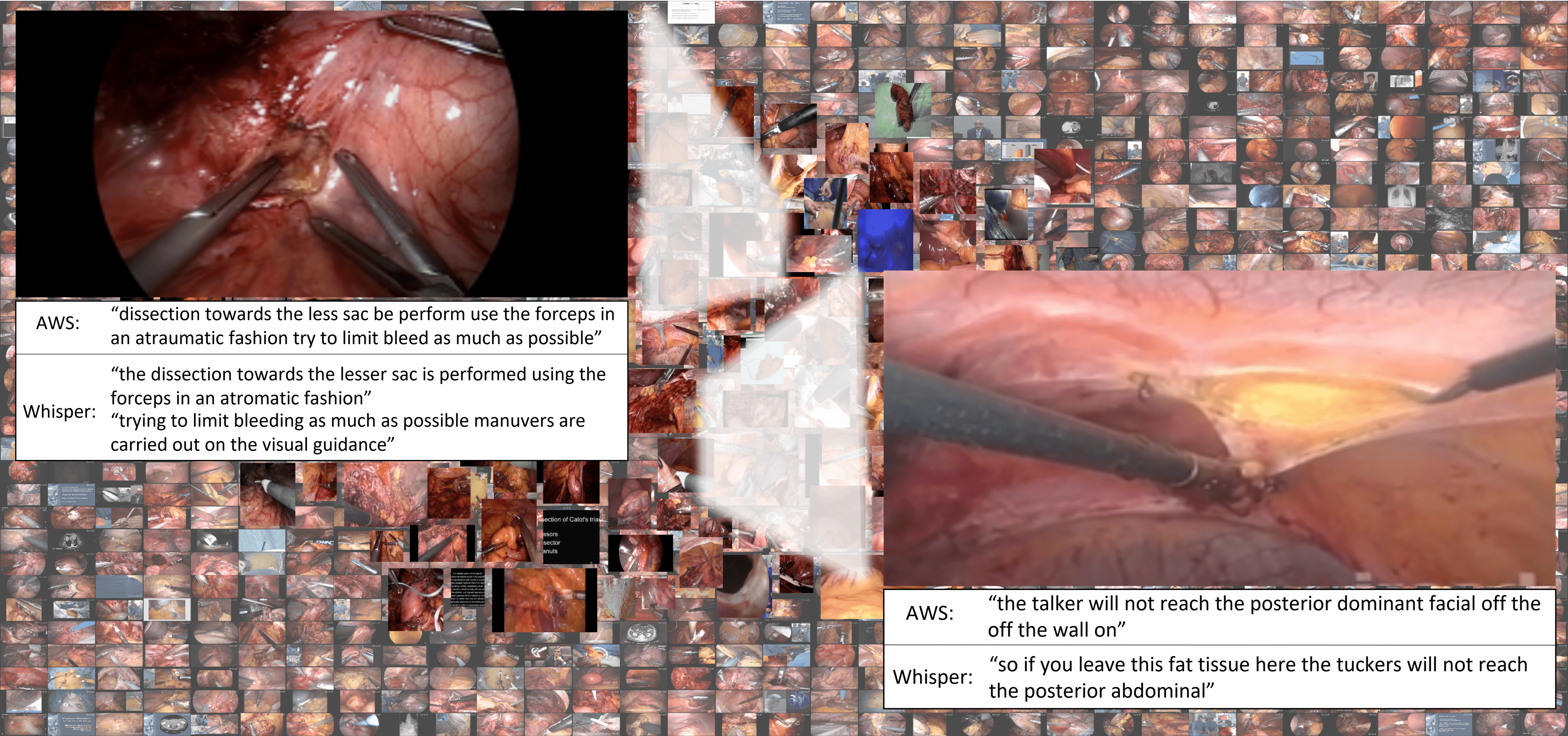}}
\caption{Examples of video clip-text pairs from SVL dataset. The video clip-text pairs are pairs of video clips and their corresponding transcripts. We generate transcripts for hundreds of surgical video lectures using two ASR systems, {i.e., AWS Medical Transcribe~\citep{AWS} and Whisper~\citep{radford2022robust}}. The transcripts usually illustrate the essential concept of surgical anatomies, instruments and events. We use large-scale video clip-text pairs to learn joint multi-modal representations.}
\label{fig1}
\end{figure*}

Recent progress in surgical computer vision has started to pave the way for a new wave of AI-assisted support systems for the operating room (OR)~\citep{maier2017surgical,maier2022surgical,ward2021computer,mascagni2022computer,madani2020artificial,yuan2021surgical}. The field has seen substantial advancements, evolving from coarse-level surgical workflow recognition~\citep{blum2008modeling,blum2010modeling,padoy2012statistical,twinanda2016endonet,dergachyova2016automatic} to fine-grained surgical scene understanding through surgical action triplet~\citep{nwoye2021rendezvous}, pixel-level scene segmentation~\citep{allan20192017,alapatt2021temporally}, and surgical scene reconstruction~\citep{wang2022neural,pfeiffer2019generating,rivoir2021long}. Nevertheless, the current advancements present {three} major limitations. First, these approaches have largely been focused on building task-specific fully-supervised deep-learning models that demand a significant effort from clinical experts to generate labeled ground truth. Second, the efficacy of these approaches has primarily been demonstrated on a limited number of mono-centric, procedure-specific surgical video datasets, which are not representative enough to encompass the complex intricacies of the overall surgical workflow~\citep{eisenmann2022biomedical}. {Third, these approaches do not explicitly incorporate the rich semantics of natural language texts into their design. The natural language text, incorporating broad visual concepts, can serve as natural supervision for visual models, ensuring high generality and usability to a diverse set of downstream tasks.} Approaches that can extend to multiple downstream tasks using {minimally labeled natural language supervision}, while leveraging large-scale multi-procedural surgical videos, will be instrumental in scaling up the approaches for widespread adoption.

In the general computer vision domain, \textit{multi-modal representation learning}~\citep{radford2021learning,miech2020end} that combines visual and free-form natural text information is emerging as a viable alternative to circumvent the need to collect labeled training data for different downstream tasks~\citep{radford2021learning,jia2021scaling}. These approaches aim to learn a low-dimensional joint latent space by pre-training two parallel encoders - one for vision and one for text - on large-scale paired visual-textual inputs. The shared latent space of the two modalities enables zero-shot transfer learning, i.e., the ability of the pre-trained visual and text encoders to adapt to different downstream tasks without fine-tuning using task-specific labels. This breakthrough has led to promising results in a wide range of general computer vision applications, including zero-shot image classification~\citep{radford2021learning}, image captioning~\citep{nukrai2022text}, semantic image retrieval~\citep{sain2023clip}, and text-to-shape generation~\citep{sanghi2022clip}.

Considering this impressive advancement in multi-modal representation learning, a natural question arises: can such high-level joint representations be learned for surgical computer vision? If possible, this could be a significant step forward in the progress of surgical data science~\citep{maier2022surgical}. By obtaining such representations, it would not only enable us to perform existing surgical video analysis tasks, such as coarse-grained to fine-grained surgical workflow recognition~\citep{twinanda2016endonet,nwoye2021rendezvous},  without using task-specific labels, but it would also open up new avenues for scalable and intelligent cognitive aids in the OR. These include vision-and-language applications, such as surgical visual question answering~\citep{seenivasan2022surgical}, surgical report generation~\citep{xu2021class}, and facilitating interactive communication between clinicians and surgical devices.

This work introduces \textit{SurgVLP}, Surgical Vision Language Pre-training, a deep learning approach to perform large-scale multi-modal representation learning for surgical computer vision. Developing such an approach is not without its unique challenges. One of the primary obstacles is the unavailability of large-scale multi-modal multi-procedural surgical datasets compared to the millions of multi-modal visual-textual pairs available in the general computer vision domain~\citep{radford2021learning,grauman2022ego4d,miech2019howto100m}. For instance, a recently developed Ego4D~\citep{grauman2022ego4d} dataset collects $3,000$ hours of activity videos and manually narrates them. Such methods are unattainable in the surgical field due to the significant human efforts required in collecting and annotating surgical videos. As our first contribution, we propose to employ surgical video lectures available through open surgical e-learning platforms such as WebSurg~\citep{WebSurgt99:online} and EAES~\citep{eaes_2023}, and online video sharing platforms such as YouTube~\citep{1YouTube5:online}, for visual-textual multi-modal learning. Compared to the manually labeled medical imaging reports~\citep{chen2022multi} or surgical instructions~\citep{rojas2020daisi}, we propose to employ unprocessed and possibly noisy audio as the primary source of supervision for multi-modal representation learning. We leverage recent advances in audio speech recognition (ASR)~\citep{mehrish2023review} to transcribe the lecture audio into sentences and link them to the corresponding video segment to construct large amounts of video clip-text pairs, as shown in Fig. \ref{fig1}. The resulting surgical video lecture (SVL) dataset contains diverse descriptions of surgical events, instrument usage, and anatomical status across various surgical procedures, thereby providing sufficient supervision to enable multi-modal representation learning for surgery.

Multi-modal representation learning using the SVL dataset nonetheless poses several linguistic challenges. First, the surgical concepts described in these videos use domain-specific knowledge and scientific terminology not typically encountered in general computer vision. For instance, \textit{``grasping the neck of the gallbladder and retracting it towards the left lower quadrant to open up the hepatocystic triangle''} and \textit{``dissect above an imaginary safety line connecting Rouviere's sulcus and the base of the fourth liver segment''} are surgery-specific descriptions commonly found in the surgical video lectures for the laparoscopic cholecystectomy procedure. Furthermore, there could be a semantic misalignment between the surgical video clips and the corresponding textual descriptions. In fact, the lecturer describing the surgical procedure might divert from the case at hand and recall a similar case with a bleeding event, even if this is not shown in the associated video. Additionally, these videos have long-range dependencies. For example, the lecturer might comment on the importance of an adequate dissection to obtain tension-free anastomosis, even though that dissection step was shown at the procedure's beginning or edited out. Finally, while recent ASR models~\citep{chen2022maestro,radford2022robust} can effectively transcribe day-to-day speech, their performance is suboptimal in surgical scenarios due to the surgery-specific linguistic challenges, as described before. For example, the state-of-the-art ASR Whisper model~\citep{radford2022robust} can understand the sentence structure and common words but struggles with surgical-specific terms (e.g., transcribing ``jejunostomy'' as ``egenostomy''). Commercial medical-specific solutions, such as AWS~\citep{AWS}, is considerably better at transcribing medical terminology but often fail to capture the overall structure and boundaries of the sentences.

We propose two key techniques for developing surgery-specific multi-modal representation learning. First, we employ text transcription from two noisy but complementary ASR systems, i.e., Whisper~\citep{radford2022robust} and AWS~\citep{AWS}, to obtain improved supervisory signals for the learning process, {as shown in Fig. ~\ref{fig1}}, effectively mitigating the limitations and inaccuracies associated with each individual system. Second, we propose a new contrastive learning objective that utilizes these dual text transcriptions from the ASR systems with the corresponding video clip. The proposed contrastive learning objective aims to encourage the embedding vectors of the video clip and the corresponding dual textual transcriptions to be close in the joint latent space. By doing so, the learned multi-modal representation retains common semantics present in the noisy ASR transcripts, enabling a more effective fusion of visual and textual information.

To effectively showcase the representation capability of the learned joint latent space, we introduce various vision-and-language tasks for surgery to serve as multi-modal benchmarks for evaluation. These tasks include \textit{text-based video retrieval}, \textit{temporal activity grounding}, and \textit{video captioning}. The text-based video retrieval task aims to associate a given text query to various video clips, while the temporal activity grounding task involves localizing a given text query to a specific video segment in the entire video. These two tasks examine how well the joint latent space captures the underlying relationship inherent in surgical visual information and its textual descriptions. The video captioning task aims to generate captions for a given surgical video clip. Since this is a generative task, it entails using a text decoder to produce coherent textual output. We propose an approach to construct a \textit{text decoder} and append it to our pre-trained encoders, seamlessly repurposing our pre-trained model to function as a video captioner. The whole process requires only the textual data to train the text decoder model. We demonstrate a notable improvement over the baseline methods for all the vision-and-language tasks.

Next, we assess the robustness and adaptability of our approach when applied to unseen surgical datasets and tasks. Specifically, we examine its performance in traditional vision-only surgical tasks, including surgical tool, phase, and action triplet recognition~\citep{twinanda2016endonet,nwoye2021rendezvous}. We evaluate our approach as a zero-shot transfer learning by processing the category labels (tool, phase, or action triplet) into textual form and classifying the video frames based on the similarity of visual and textual latent vectors. The results show that the general surgical concepts learned by our multi-modal joint representations from various surgical procedures can benefit a specific surgical procedure, such as laparoscopic cholecystectomy. {To the best of our knowledge, this is the first work to demonstrate self-supervised multi-modal pretraining for recognizing surgical tools, phases, and action triplets without annotation. While our zero-shot performance trails fully supervised baselines, particularly in tasks requiring fine-grained anatomical reasoning, the results highlight SurgVLP’s potential to serve as a foundation backbone to reduce annotation costs for downstream tasks.} Finally, we conduct extensive ablation studies to shed light on the different components of our approach and their impacts on the results. The contributions of our work can be succinctly summarized in the following four key aspects:
\begin{itemize}
    \item We propose to harness the knowledge from surgical video lectures accessible via open surgical e-learning platforms for visual-textual multi-modal representation learning. To this end, we introduce a large-scale dataset of surgical video lectures (SVL) comprising 1.4k procedural videos.  
    \item We propose to leverage text transcription from two complementary ASR systems, Whisper and AWS, {to enhance the representation learning process by addressing the linguistically inaccurate sentences produced by these ASR systems. }
    \item We propose a novel contrastive learning objective that leverages dual text transcriptions from ASR systems and the corresponding video clip, aiming to encourage close proximity of embedding vectors in the joint latent space.
    \item {We demonstrate the zero-shot transferability of our proposed framework in multiple vision-language and vision-only tasks.} 
\end{itemize}

\section{Related Works}

\subsection{{Advances in representation learning methods}}

Traditional fully-supervised approaches learn representations from datasets of images or videos with manually annotated labels. However, manual annotation is laborious and expensive, especially when creating labeled datasets for new problems. Therefore, several works have studied representation learning by deriving supervision from unlabeled visual data, i.e., self-supervised learning (SSL) methods. 

\textbf{Self-supervised learning}. Typically, SSL methods rely on heuristics-based pretext tasks by pre-training the model to solve handcrafted tasks having some degree of relevance to downstream tasks. These pretext tasks focus on the consistency of visual entities, including predicting the geometric transformations~\citep{jing2018self}, the future representation~\citep{vondrick2016anticipating} as well as colorizing videos~\citep{vondrick2018tracking}. However, these methods heavily rely on the quality of the pretext tasks to show improvement in the downstream tasks. Recently, contrastive learning methods have emerged as an alternative. These methods regulate the distribution of feature vectors within the embedding space. This regulation process is accomplished by generating \textit{positive} and \textit{negative} pairs and then applying a discriminative loss function. This function brings the embeddings of \textit{positive} pairs closer together while simultaneously distancing the embeddings of \textit{negative} pairs. Specifically, MoCo~\citep{he2020momentum} proposes to store embeddings of \textit{negative} samples in a queue-based memory bank which is continuously updated during the course of contrastive learning. MoCo v2~\citep{chen2020improved} and SimCLR~\citep{chen2020simple} are subsequent works that proposed to improve representation learning by using more advanced data augmentations.

In the surgical computer vision field, SSL methods have also been studied to support different surgical downstream tasks ranging from estimating remaining surgery duration to surgical activity recognition ~\citep{ross2018exploiting,yengera2018less,funke2018temporal,ramesh2022dissecting,rivoir2019unsupervised}. Specifically, these approaches employ SSL methods to pre-train the model to obtain robust and effective representations and then fine-tune the pre-trained model on different surgical downstream tasks to achieve superior performance.

Our approach distinguishes itself from these vision-only SSL methods by performing representation learning using supervisory guidance derived from language. Language supervision is produced by applying automatic speech recognition (ASR) systems to surgical video lectures. We focus on learning the multi-modal representations exclusively from uncurated surgical lecture videos, presenting a more realistic and scalable learning scenario. 

\textbf{Language in representation learning}. The potential of language in deep learning has been harnessed through advancements in natural language processing (NLP). Transformer-based models like Llama~\citep{touvron2023llama} and T5~\citep{raffel2020exploring} have set new standards in tasks such as translation and summarization. Moreover, transformer encoders like Bart~\citep{lewis2019bart} and RoBerta~\citep{liu2019roberta} have improved bidirectional language understanding, enhancing natural language task performance. In the medical field, models like SciBert~\citep{beltagy2019scibert} and BioClinicalBert~\citep{huang2019clinicalbert}, trained on large-scale biomedical data and electronic health records, respectively, have significantly improved biomedical text mining and healthcare prognostics. 

Different from the conventional visual-only SSL methods as discussed before, recent representation learning methods have started to explore language as an alternative semantic supervision. Numerous works have proposed to use the paired image/video and text datasets to learn a joint latent space where the visual and textual data are close if they are semantically similar~\citep{changpinyo2021conceptual,krishna2017visual,chen2015microsoft}. However, these methods still rely on human annotators to describe the visual content. To avoid labeling images, several works~\citep{mahajan2018exploring} have leveraged image titles, descriptions, and hashtag metadata to provide language supervision. The recently proposed CLIP model utilizes a dataset with $400$ million image-text pairs for multi-modal representation learning, achieving superior zero-shot performance on various downstream tasks~\citep{radford2021learning}. Also, De-CLIP leverages various supervisions, including self-supervision within each modality and multi-view supervision across modalities to efficiently optimize the model using large-scale image-text paired datasets~\citep{li2021supervision}.

{\textbf{Multi-modal representation learning framework}. When incorporating the additional language modality in representation learning, several frameworks are commonly used for modality integration. These frameworks can be classified into three types~\citep{guo2019deep}: joint representation learning, coordinated representation learning, and encoder-decoder modeling. Each has a distinct architecture and approach to integrating multi-modal features.}

{Joint representation learning aims to extract a single feature vector that fuses complementary information from uni-modal feature vectors. Recent methods employ a transformer architecture with attention mechanisms after the uni-modal encoders to merge cross-modal features~\citep{luo2020univl,liu2016multimodal,wu2014zero,habibian2016video2vec,poria2016fusing}. Utilizing heterogeneous features from different modalities has shown promising results in discriminative tasks, such as video and image classification~\citep{li2019visualbert,liu2016multimodal,jiang2017exploiting}.}

{Coordinated representation learning seeks to learn separate but constrained representations for each modality within a coordinated space. Specifically, it learns a joint embedding space where embedding vectors from different modalities can be compared by semantic similarities. Recent works leverage a dual-branch model trained using contrastive learning on large-scale image-text pairs~\citep{radford2021learning,xu2021videoclip,miech2020end,li2022blip}, resulting in impressive zero-shot performance in image classification and cross-modal retrieval tasks. By learning joint representations, this framework preserves the unique and useful characteristics of each modality. However, its dual-branch nature limits its application to cross-modal generative tasks. Our SurgVLP framework falls under coordinated representation learning, focusing on discriminative surgical downstream tasks such as phase recognition. We use a dual-branch model to generate uni-modal embeddings for video and text, learning their semantic correspondences by adding distance constraints.}

{Encoder-decoder modeling is used to learn an intermediate representation that maps one modality to another. The encoder converts the source modality into a latent vector, which the decoder then uses to generate a novel sample of the target modality. This approach is particularly beneficial for generative tasks such as text-to-image synthesis~\citep{reed2016generative,rombach2022high} and image captioning~\citep{xu2015show,liang2017recurrent}. }

\textbf{Audio in representation learning}. While exciting works have been proposed using natural language supervision, the application of natural language supervision for video representation learning has received comparatively less attention. A significant hurdle is the high cost of annotating textual descriptions for video content. Some recent works adopt audio to learn a self-supervised multisensory representation~\citep{owens2018audio,alayrac2020self}. These approaches train a neural network to predict whether video frames and audio are temporally aligned. However, these works operate on the event audio signals, such as instrument or environmental audio, rather than spoken audio.  As a result, these approaches are limited to specific audio contexts, and they may not perform optimally in scenarios where spoken language is the primary focus. 

Spoken audio has gained more research interest with the rapid development of audio speech recognition (ASR) methods~\citep{seide2011feature,amodei2016deep,narayanan2018toward}. These ASR methods achieve a much lower word error rate when trained on the $2,000$ hours Switchboard dataset~\citep{seide2011feature}. By leveraging the semi-supervised strategy, authors in ~\citep{narayanan2018toward} have expanded the training dataset size to $162,000$ hours of labeled audio, thereby further decreasing the word error. The recently introduced Whisper model has set the new state-of-the-art performance in the general ASR domain by training a transformer decoder using a large-scale dataset in a weakly supervised manner~\citep{radford2022robust}. 

Powered by the emergence of ASR, HowTo100M~\citep{miech2019howto100m,miech2020end} constructs a large-scale paired video and text dataset by transcribing audio into textual annotations, thereby providing large-scale language supervision for representation learning. YT-Temporal~\citep{zellers2021merlot} creates a dataset for learning multimodal knowledge from textual graphs derived from 6 million public YouTube videos. It applies multiple filtering and denoising strategies to obtain clean and better-aligned audio transcripts. By utilizing audio from a wide range of instructional videos, these datasets facilitate the creation of scalable and robust models capable of understanding natural language from diverse, real-world scenarios.

While Whisper has shown promising transcription results in the general ASR domain, it faces challenges in accurately transcribing domain-specific content, such as surgical video lectures. Whisper is primarily trained on extensive multilingual general domain datasets, which are less likely to include a substantial number of surgery-specific words. To address this issue, we propose to use two ASR systems to generate video clip-text pairs from surgical video lectures. Specifically, we use AWS Transcribe Medical~\citep{AWS} as a complementary ASR system to Whisper. AWS Transcribe Medical is a service provided by Amazon Web Services and is specifically designed to transcribe medical keywords accurately. Combining these two ASR systems addresses the limitations associated with the individual ASR system, thus helping to create effective transcriptions for surgical video lectures.

\subsection{Downstream tasks}

The quality of representation learning is typically assessed by applying the learned representations to various downstream tasks. In our context, we categorize these downstream tasks into two main groups: vision-and-language and vision-only tasks. Vision-and-language tasks are at the intersection of vision and language modalities and can be classified into two categories: visual-textual understanding and generation tasks. Visual-textual understanding tasks aim to correlate the information across different modalities, such as text-based video retrieval. Visual-textual generation tasks aim to generate one modality output from another, such as video captioning. The objective of vision-only tasks is to understand and interpret the visual data from a single vision modality. A few examples of these tasks include image recognition, object detection, semantic segmentation, and pose estimation. In the following, we describe different vision-and-language and vision-only tasks. 

\textbf{Vision-and-language tasks}. With the substantial progress in NLP~\citep{devlin2018bert,brown2020language} and multi-modal self-supervised learning~\citep{radford2021learning}, research has been shifting towards multi-modal vision-and-language tasks. Text-based video retrieval~\citep{gabeur2020multi} enables efficient video content searching and understanding using natural language queries. Temporal activity grounding~\citep{gao2017tall} seeks to identify and localize specific action steps within a video sequence given a text description. In the context of surgical vision-and-language tasks, SurgVQA~\citep{seenivasan2022surgical} proposes a visual question-answering system designed to answer the questions related to surgical scenes, with the aim to offer real-time support to junior residents. Surgical report generation is also explored to describe current surgical events and objects~\citep{xu2022rethinking}. Similarly, DAISI~\citep{rojas2020daisi} has introduced a method for generating surgical instructions for different surgical procedures.

\textbf{Vision-only tasks}. Vision-only tasks have made substantial advancements using fully-supervised deep learning methods. These advancements have shown state-of-the-art performance in object detection~\citep{girshick2015fast}, action segmentation~\citep{tang2019coin,farha2019ms}, image classification~\citep{deng2009imagenet,he2016deep}, and action recognition~\citep{soomro2012ucf101}. These successes are primarily attributed to the power of deep representations in capturing complex visual patterns. 

The surgical computer vision community has embraced these advancements, exploring numerous studies on surgical tool and phase recognition~\citep{twinanda2016endonet, dergachyova2016automatic, al2018monitoring,shah2023glsformer}.  Following this, researchers proposed more refined spatiotemporal architectures to capture the dynamic and complex nature of surgical tasks~\citep{jin2017sv,czempiel2020tecno,czempiel2021opera,jin2021temporal, rivoir2022pitfalls}

For a more comprehensive understanding, a recent survey provides an overview of surgical phase recognition approaches~\citep{garrow2021machine}. For a more fine-grained surgical activity recognition, authors in \citep{nwoye2021rendezvous} have introduced the challenging action triplet recognition task for laparoscopic cholecystectomy procedures. They propose a dataset to classify instrument, verb, and target components from surgical scenes. However, all previous methods are vision-only, fully supervised methods that rely heavily on human-annotated curated data. 

In this work, we investigate both vision-and-language and vision-only surgical downstream tasks. We apply our approach to vision-and-language tasks, including text-based video retrieval, temporal activity grounding, and video captioning, to assess how well the model integrates and correlates information across modalities. We also apply the model to vision-only tasks, including surgical tool, phase, and action triplet recognition, to investigate whether our representation learning from uncurated data benefits a specific procedure, for example, laparoscopic cholecystectomy. We take a first step towards generic representation learning for surgery by performing zero-shot recognition on these surgical computer vision tasks. By evaluating our models on both vision-and-language and vision-only tasks, we aim to provide a comprehensive view of the capabilities and limitations of our approach.

\begin{figure*}[!t]
\centerline{\includegraphics[width=2.05\columnwidth]{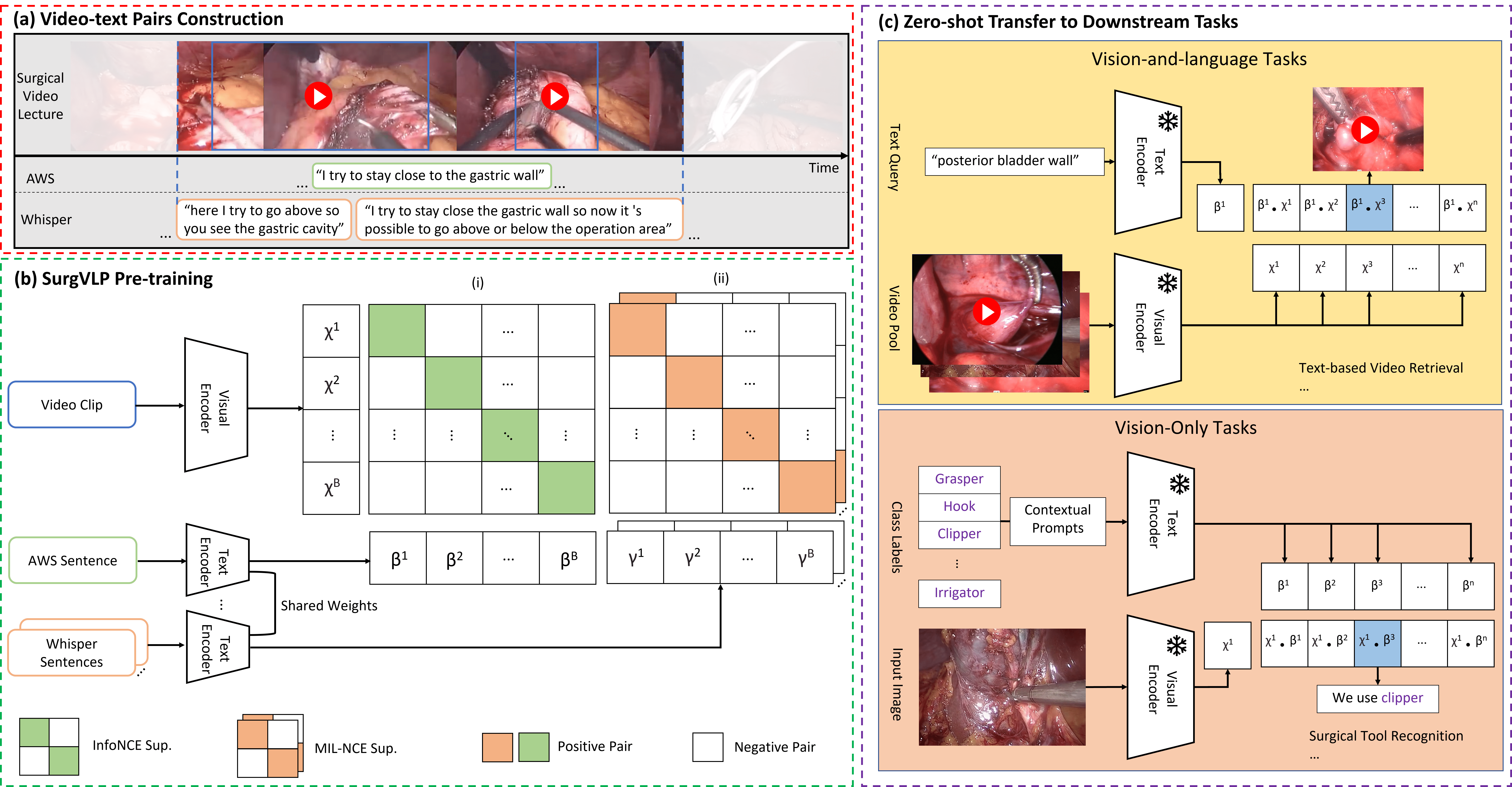}}
\caption{{Pipeline of proposed SurgVLP. Figure (a) shows examples of video clip-text pairs and their construction process. We have two text views and we pair them to random lengths of video clips. Figure (b) presents the contrastive learning objective with AWS sentences and Whisper sentences.} \textit{SurgVLP} utilizes the Info-NCE and MIL-NCE losses for AWS and Whisper sentences, respectively. Figure (c) illustrates how to perform downstream tasks in the zero-shot setting. We show the vision-and-language tasks, e.g., text-based video retrieval and temporal activity grounding, at the top and the vision-only tasks at the bottom.}
\label{pipeline}
\end{figure*}

\section{Approach}
\label{sec:approach}

This section outlines our \textit{SurgVLP} approach for learning joint multi-modal representations from surgical video lectures and its adaptability to multiple downstream tasks. We first introduce the process of constructing a large-scale video clip-text pair dataset from surgical video lectures. Then, we present our dual-branch architecture consisting of one branch for the visual encoder and another for the text encoder to extract the latent vectors from the video clip and {the corresponding multiple views of texts from the ASR systems}, respectively. {We follow CLIP~\citep{radford2021learning} and use a dual-branch architecture to conduct the multi-modal representation learning}. We then describe our contrastive learning objective to optimize the dual-branch architecture using the joint latent vectors. Finally, we describe a variety of vision-and-language and vision-only surgical downstream tasks to evaluate learned multi-modal representations.

\subsection{Video clip-text pairs}
\label{video-text}

We utilize the recent advancements in audio speech recognition (ASR) to transcribe lecture audio into textual sentences. In particular, we use two different ASR systems to generate complementary text views, namely the AWS Medical Transcribe ASR system~\citep{AWS} and the Whisper ASR system~\citep{radford2022robust}, as we observe that ASR systems perform differently on the specific surgical domain. 
In particular, AWS Medical Transcribe is used to understand medical terminology and surgery-specific terms, and Whisper is used to understand the overall sentence structure and common words. {As shown in Table.~\ref{tab:aws_whisper}, we have included a direct comparison of text generated by AWS and Whisper to illustrate their respective strengths in ASR for surgical language understanding. Our observations confirm that AWS excels in recognizing specific surgical terms, making it particularly useful for capturing domain-specific vocabulary accurately. For example, in the first case, AWS correctly transcribes ``trocar,'' which is an important surgical instrument, while Whisper misinterprets it as ``choker.'' However, Whisper provides a more structured and coherent transcription, as seen in the second example, where it accurately reconstructs the full sentence with proper context, compared to AWS’s fragmented output. These differences highlight a key trade-off: AWS is more precise in recognizing surgical terms, whereas Whisper generates more fluent and contextually complete sentences. This distinction is crucial when designing multi-view models, as combining outputs from both ASR systems helps leverage AWS's terminology accuracy and Whisper's sentence coherence, leading to more effective language supervision for surgical video-language pretraining.}

\begin{table}[h]
    \centering
    \renewcommand{\arraystretch}{1.2}
    \begin{tabular}{p{0.45\linewidth} | p{0.45\linewidth}}
        \toprule
        \textbf{AWS} & \textbf{Whisper} \\
        \midrule
        this be possible through the use of a \textbf{trocar} connect to a glove & this is possible through the use of a \textbf{choker} connected to a glove a tip that has been described numerous times in the literature \\ 
        \midrule
        a tube \newline
        the plication be perform start on the right side of the esophagus \newline
        on use any thigh limb \newline
        two oh stitch 
        & a two-way fund application is performed starting on the right side of the esophagus and using ethybon 2 0 stitches \\
        \bottomrule
    \end{tabular}
    \caption{{Comparison of transcriptions generated by AWS and Whisper ASR systems.}}
    \label{tab:aws_whisper}
\end{table}

To obtain the first text view, we start with the AWS ASR system and use it to convert lecture audio into text transcriptions. We then apply text preprocessing to truncate the transcription into sentences based on the pre-defined stop symbols. Then, we apply filtering strategies to keep the sentences containing the surgical terminologies and remove meaningless and extremely short sentences. Thus, for a given surgical lecture video, these operations extract multiple AWS sentences, where each sentence is accompanied by its start and end timestamps. Due to the filtering, the AWS sentences contain surgery-specific terms but are sparsely distributed along the video time axis. More details on text pre-processing are given in section \ref{pretrain-dataset}.

To obtain the second text view, we use the Whisper ASR system to generate the text transcription. We apply the same text preprocessing strategy as above to truncate the transcription into sentences. We do not apply heavy filtering because the Whisper ASR system generates meaningful sentences with clear stop symbols. Therefore, the processed Whisper sentences possess different punctuation patterns than AWS sentences and are densely distributed along the video time axis. This variation in punctuation and distribution leads to potential overlapping between AWS and Whisper sentences, as shown in {Fig.~\ref{pipeline} (a)}. Thus, one AWS sentence usually corresponds to $M \geq 1$ overlapped Whisper sentences. To extract the corresponding video clip from given multiple text views, we propose the following steps: 
\begin{itemize}
    \item sample an AWS sentence to ensure that the sentence contains surgery-specific term; 
    \item find $M$ overlapped Whisper sentences;
    \item merge Whisper sentences and sample a timestamp, called \textit{center timestamp}, within the boundary of merged Whisper sentences;
    \item grow a video clip of random duration (up to $10$ seconds) from this \textit{center timestamp} as the center for the video clip.
\end{itemize}
Below, we give our intuitions behind the choices made for the video clip sampling. 

\textbf{Better clip sampling with key surgical terms.} Randomly sampling a video clip from a surgical lecture video increases the chances of having a video clip that either does not align with the relevant text or has no correspondence with any text at all. Therefore, sampling the AWS sentences to locate video clips ensures that the sampled video clip contains surgical terms of interest.

\textbf{Better video clip boundary.} The start and end timestamps of AWS sentences provide a rough range for video clip duration. However, sampling video clips based on the exact timestamps of AWS transcribed sentences could produce less relevant video clips. This is because AWS transcribed sentences miss explicit sentence boundaries, thereby producing weak alignment to the visual content. To address this, we propose to sample video clips within the boundary of merged Whisper sentences because they provide better and more explicit activity boundaries. We also sample video clips of different lengths to encourage fine-grained association. Thus, a video clip may have a better chance of being aligned or supervised by nearby text. {When using only one text view (either AWS or Whisper), video cutting is based on the timestamps obtained from the transcripts of the respective ASR system.}

These steps produce a set of video clips, each associated with one AWS sentence and $M$ Whisper sentences. This constitutes our \textit{video clip-text pair} dataset, as illustrated in Fig.~\ref{fig1} and Fig.~\ref{pipeline} (a).

\subsection{Dual-branch model}
Given the \textit{video clip-text pair} dataset, we denote a video clip as $v$ and its corresponding AWS sentence and Whisper sentences as $a$ and {$[w^m], m \in \{1,2,...M\}$}, respectively. Here, $m$ indexes over each Whisper sentence. Given all paired video clips and transcribed sentences {$\{v_i, a_i, [w^m_i]\}_{i=1}^{K}$}, \textit{SurgVLP} aims to learn a joint latent space that correlates semantically similar video clips with the corresponding multiple view texts. Here, $K$ is the total number of video clip-text pairs in the dataset. Formally, we aim to learn two parameterized mappings: $\mathcal{F}: v \rightarrow \mathbb{R}^d$ maps a video clip into a $d$ dimensional latent vector, and $\mathcal{G}: a / w^m \rightarrow \mathbb{R}^d$ maps an AWS or a Whisper sentence into a $d$ dimensional latent vector. To learn these two mappings, we employ a dual-branch model with a vision branch $\mathcal{F}$ using a CNN-based visual encoder and a text branch $\mathcal{G}$ using a transformer-based text encoder, described in the following.

\subsubsection{Visual encoder}
We employ the ResNet-50 model~\citep{he2016deep} pre-trained on ImageNet as the base architecture for the visual encoder $\mathcal{F}$. We replace the last layer of ResNet-50 with a linear layer to project the incoming global-averaged pooled vector into the $d$ dimensional vector. For a given video clip $v$, we first sample a fixed number of frames $\{z_0,...,z_T\}$ and then feed them through $\mathcal{F}$ to obtain the latent vectors for each frame, i.e., $\mathcal{F}(z_i) \in \mathbb{R}^d$. Subsequently, we perform average pooling across frames to obtain the final latent vector $\chi \in \mathbb{R}^{d}$, as shown in the following: 
\begin{equation}
\centering
\chi = \frac{1}{T} \sum^{T}_{i} \mathcal{F}(z_i).
\label{eq1}
\end{equation}

{In this work, we adopt ResNet-50 with ImageNet initialization as our visual encoder, prioritizing spatial feature extraction in surgical videos. While Vision Transformers (ViTs) have demonstrated strong capabilities in modeling long-range dependencies, their effectiveness heavily relies on large-scale pretraining datasets, which are often unavailable in specialized domains like surgical vision-language tasks. Moreover, previous works in surgical workflow analysis predominantly utilize CNN-based architectures, making ResNet-50 a fair baseline for comparison. Additionally, ViTs exhibit higher initialization sensitivity, leading to less stable performance in our vision-language pretraining framework. Since temporal modeling is beyond the scope of this work, we focus on extracting strong spatial representations, where ResNet-50 proves to be a robust and efficient choice for surgical video understanding.}

\subsubsection{Text encoder}

As mentioned above, we use AWS and Whisper ASR systems to transcribe surgical audio into surgical texts. The surgical texts generated using these ASR systems are inherently noisy, missing either the overall sentence structure or domain-specific scientific terminologies. Thus, a robust and domain-specific text encoder is required to handle the complexity of the surgical texts. We propose to use a transformer-based text encoder to encode the transcribed surgical sentences into representative latent vectors. In particular, we employ BioClinicalBert~\citep{huang2019clinicalbert} as our text encoder $\mathcal{G}$. It is a base-size Bert model~\citep{devlin2018bert}, containing $12$ encoders with $12$ bidirectional self-attention heads totaling $110$ million parameters. It is pre-trained on medical texts and generates more representative features than other text encoders, such as SciBert~\citep{beltagy2019scibert} and Bert~\citep{devlin2018bert}, see section \ref{ablation-study} for a comparison. 

Given an AWS or Whisper sentence, we first split each word into $n_i$ subwords based on the word-piece tokenizer. Therefore, we generate a total of $N = \sum_{i=1}^L n_i$ subwords for a sentence with $L$ words. For the sentences of different lengths, we truncate or pad them to the fixed length of $N$ subwords. The word-piece tokenizer addresses out-of-vocabulary words and typo errors by splitting them into subword units. It maintains linguistic meaning and minimizes the adverse effects of unknown surgical terminologies. We consider each subword as a token $t_i$ and pass its token ID to the tokenizer's lookup embedding table to get the corresponding input embedding vectors of size $\mathbb{R}^{N \times 768}$. Then, input embedding vectors are passed through multiple transformer layers of the text encoder $\mathcal{G}$, resulting in a feature vector of size $\mathbb{R}^{N \times 768}$. Then, $\mathcal{G}$ performs the global average pooling to aggregate $N$ feature vectors into one vector. Finally, a linear layer (MLP) projects this vector into the $d$ dimensional output latent vector as follows: 
\begin{equation}
\label{eq3}
\centering
\beta / \gamma^m \ = MLP(\frac{1}{N} \sum_{i}^N \mathcal{G}(t_i)).
\end{equation}
Here, $\beta$ and $\gamma^m$ represent a $d$ dimensional latent vector for each sentence transcribed from AWS and Whisper ASR systems, respectively. {In the learned joint embedding space, the visual and textual embedding vectors are represented in $\mathbb{R}^{N \times 768}$ to largely preserve the textual semantics and contextual information extracted from BioClinicalBert~\citep{huang2019clinicalbert}.}

\subsection{Multiple text-views contrastive supervision}

In this section, we describe our multiple text-views contrastive learning objective. Given the latent vectors {$\{\chi_i, \beta_i, [\gamma^m_i]\}_{i=1}^{K}, m \in \{1,2,...M\}$} from video clip-text pairs {$\{v_i, a_i, [w^m_i]\}_{i=1}^{K}, m \in \{1,2,...M\}$}, we first consider the case when we have text supervision from only one ASR, i.e., supervision from AWS text latent vectors ($\beta_i$). In this case, we employ the InfoNCE loss, which is a contrastive loss function utilized in self-supervised learning~\citep{oord2018representation, he2020momentum, chen2020simple} and multi-modal representation learning~\citep{radford2021learning,li2021supervision}; here NCE stands for Noise Contrastive Estimation. By utilizing the InfoNCE loss, denoted as $\mathcal{L}_{InfoNCE}$, we align the visual latent space with the corresponding AWS textual latent space, as follows: 
\begin{equation}
    \centering
    \mathcal{L}_{InfoNCE} =- \frac{1}{B} \sum_{i=1}^{B} \log \frac{\exp(sim(\chi_i, \beta_i)/\tau)}{\sum^B_{j=1} \exp(sim(\chi_i, \beta_j)/\tau)}.
    \label{InfoNCE}
\end{equation}
Here, the numerator $sim(\chi_i, \beta_i)$ is the cosine similarity score between visual and corresponding AWS text latent vectors, i.e., \textit{positive} pairs. The denominator $sim(\chi_i, \beta_j)$ is the cosine similarity between the visual latent vector and all other AWS text latent vectors in the batch, i.e., \textit{negative} pairs, as illustrated in Fig.~\ref{pipeline} (b). Also, $B$ is the batch size, and $\tau$ is the temperature hyper-parameter to control the probability distribution over the positive and negative pairs in the embedding space~\citep{wu2018unsupervised}. A high-temperature value smooths out the distribution, allowing the model to consider a large number of negative pairs. A low-temperature value makes the distribution sharper, focusing more on the positive pairs and reducing the influence of negative examples.

While AWS sentences contain medical and surgery-specific terms, they however generate incomplete sentence fractions, eventually providing insufficient supervision when utilizing the InfoNCE learning objective. To address this issue, as our key contribution, we propose to exploit text supervision from multiple complementary ASR sources. In addition to the text supervision from the AWS text latent vectors ($\beta_i$), we utilize text supervision from Whisper latent vectors ($\gamma^m_i$) that can recognize overall sentence structure, thus enhancing the alignment between text and video clips.
 
As mentioned in section~\ref{video-text}, we link $M \geq 1$ Whisper sentences to a video clip. Therefore, we propose to extend the InfoNCE learning objective with the MIL-NCE~\citep{miech2020end} learning objective, where MIL stands for Multiple Instance Learning. The MIL-NCE learning objective, denoted as $\mathcal{L}_{MIL-NCE}$, aims to align the visual latent vector $\chi$ with multiple Whisper latent vectors $\gamma^m$, as follows:
\begin{equation}
    \centering
    \mathcal{L}_{MIL-NCE} =- \frac{1}{B} \sum_{i=1}^{B} \log \frac{\sum^M_{m=1}\exp(sim(\chi_i, \gamma^m_i)/\tau)}{\sum^B_{j=1} \sum^M_{m=1} \exp(sim(\chi_i, \gamma^m_j)/\tau)}.
    \label{milnce}
\end{equation}

Our final multiple text-views contrastive loss $\mathcal{L}$ combines these two loss functions, $\mathcal{L}_{InfoNCE}$ and $\mathcal{L}_{MIL-NCE}$, scaled by the weighting coefficients $\epsilon$, as shown in the following:
\begin{equation}
    \centering
    \mathcal{L} = \epsilon \mathcal{L}_{InfoNCE} + (1 - \epsilon) \mathcal{L}_{MIL-NCE}.
    \label{final loss}
\end{equation}

In summary, the $\mathcal{L}_{InfoNCE}$ loss is responsible for aligning surgery-specific terms from AWS sentences with video clips, and the $\mathcal{L}_{MIL-NCE}$ loss is responsible for addressing the misalignment issue that the lecturers might talk about something before or after they actually demonstrate it. Both loss functions aim to maximize the alignment between \textit{positive} pairs of video clips and texts compared to random \textit{negative} pairs in the batch $B$. 

\begin{table*}[t]
	\centering
	\caption{\small{Manually designed contextual prompts for the class names of the surgical phase and tool recognition tasks. The main action of scissors is cutting, but this action can be performed by many other instruments, such as hook. Therefore, we use ``I use scissors'' as the context prompt for the ``Scissors'' class.}}
	\scalebox{0.96}{
		\begin{tabular}{ll|ll}
			\toprule
			\textbf{{Phase Labels}} & \textbf{Prompts} & \textbf{Tool Labels} & \textbf{Prompts} \\
			\midrule
			\emph{Preparation}    & \begin{tabular}[c]{@{}l@{}} In preparation phase I insert trocars to patient \\abdomen cavity \end{tabular}& \emph{Grasper} & \begin{tabular}[c]{@{}l@{}}I use grasper or cautery \\ forcep to grasp it\end{tabular}    \\
            \hline
			\emph{CalotTriangleDissection}    & \begin{tabular}[c]{@{}l@{}} In calot triangle dissection phase I use grasper \\ to hold gallbladder and use hook to expose the \\ hepatic triangle area and cystic duct and cystic \\ artery \end{tabular} & \emph{Bipolar} & \begin{tabular}[c]{@{}l@{}}I use bipolar to coagulate \\ and clean the bleeding \end{tabular}    \\
            \hline
            \emph{ClippingCutting}    & \begin{tabular}[c]{@{}l@{}} In clip and cut phase I use clipper to clip the \\ cystic duct and artery then use scissor to cut them\end{tabular} & \emph{Hook} & \begin{tabular}[c]{@{}l@{}}I use hook to dissect it \end{tabular}    \\
            \hline
            \emph{GallbladderDissection}    & \begin{tabular}[c]{@{}l@{}} In dissection phase I use the hook to dissect the \\ connective tissue between gallbladder and liver\end{tabular} & \emph{Scissors} & \begin{tabular}[c]{@{}l@{}}I use scissor \end{tabular}    \\
            \hline
            \emph{GallbladderPacking}    & \begin{tabular}[c]{@{}l@{}} In packaging phase I put the gallbladder \ into the \\ specimen bag \end{tabular}& \emph{SpecimenBag} & \begin{tabular}[c]{@{}l@{}} I use specimenbag \\ to wrap it \end{tabular}    \\   
            \hline
            \emph{CleaningCoagulation}    & \begin{tabular}[c]{@{}l@{}} In clean and coagulation phase I use suction and \\ irrigation to clear the surgical field and coagulate \\ bleeding vessels \end{tabular} & \emph{Irrigator} & \begin{tabular}[c]{@{}l@{}} I use irrigator to suck it \end{tabular}    \\      
            \hline
            \emph{GallbladderRetraction}    & \begin{tabular}[c]{@{}l@{}} In retraction phase I grasp the specimen bag \\  and remove it from trocar \end{tabular} & \emph{Clipper} & \begin{tabular}[c]{@{}l@{}} I use clipper to clip it \end{tabular} \\\bottomrule\hline
		\end{tabular}
	}
	\label{table:prompt}
	\vspace{-1mm}
\end{table*}

\section{Downstream tasks and SurgVLP's adaptation}
\textit{SurgVLP} learns generic multi-modal representations from different surgical procedures to correlate the semantically similar video clips and texts. Therefore, in order to assess its learned joint latent representations in handling the multi-modal data, we introduce various \textit{vision-and-language} surgical tasks. These tasks include text-based video retrieval, temporal activity grounding, and video captioning to serve as comprehensive multi-modal benchmarks.

Furthermore, we investigate the zero-shot transferability of \textit{SurgVLP} to the traditional \textit{vision-only} surgical tasks. This evaluation is crucial to assess whether the learned joint latent representations can effectively generalize to an unseen surgical dataset without using task-specific labels. We evaluate \textit{SurgVLP} on Cholec80~\citep{twinanda2016endonet}, CholecT45~\citep{nwoye2021rendezvous} datasets for surgical phase, tool, and action triplet recognition. {In the following}, we provide a detailed explanation of both vision-and-language and vision-only surgical tasks.  

\subsection{Vision-and-language surgical tasks}
\label{vision-and-language surgical tasks}

\textbf{Text-based video retrieval}. Given a textual query using natural language, text-based video retrieval aims to retrieve matched video clips from a large video repository. This task allows users to search for videos using free-form language queries, which is a flexible and intuitive way of querying. An ideal retrieving system needs to handle novel queries that were not seen during the training process. Also, text-based video retrieval implies the degree of understanding across different data modalities, i.e., video and text, which is a challenging and important problem in surgical computer vision. In this work, we perform this task in a zero-shot manner as evaluation methods relying on model fine-tuning or linear probing require task-specific labels and complex hyper-parameter tuning, which may not directly reflect the true performance of representation learning. We freeze the visual and text encoders after the pre-training. We use the visual encoder to generate the visual latent vectors of the video clips from the test database. Then, given a language query, we use the text encoder to generate a text latent vector that is matched against the visual latent vectors using cosine similarity. We retrieve top-K highest-matched video clips. 

Unlike the pre-training phase, where video clips of random durations are generated, we follow a different process to generate video clips for evaluation. We manually correct Whisper transcriptions on a subset of the dataset with the help of a clinical expert. Subsequently, each individual sentence is aligned with its corresponding video clip, ensuring a precise correspondence between the textual description and the visual content. This process results in a video clip-text pair test dataset that enables a proper multi-modal evaluation of \textit{SurgVLP}.

\textbf{Temporal activity grounding}. Similar to the text-based video retrieval task, temporal activity grounding is a retrieval-based task. However, rather than retrieving from a large pool of video clips, it aims to localize a given textual query to a relevant segment within a video. It therefore requires more fine-grained video text understanding and could be useful in many applications, such as identifying critical events in a given surgical video using textual queries and surgical video editing. We perform the temporal activity grounding in a zero-shot way by dividing a long lecture video into multiple video clips and retrieving the relevant video clip based on the cosine similarity with the given textual query. We utilize the same evaluation dataset as described in the previous section.

\begin{figure}[!t]
\centerline{\includegraphics[width=1\columnwidth]{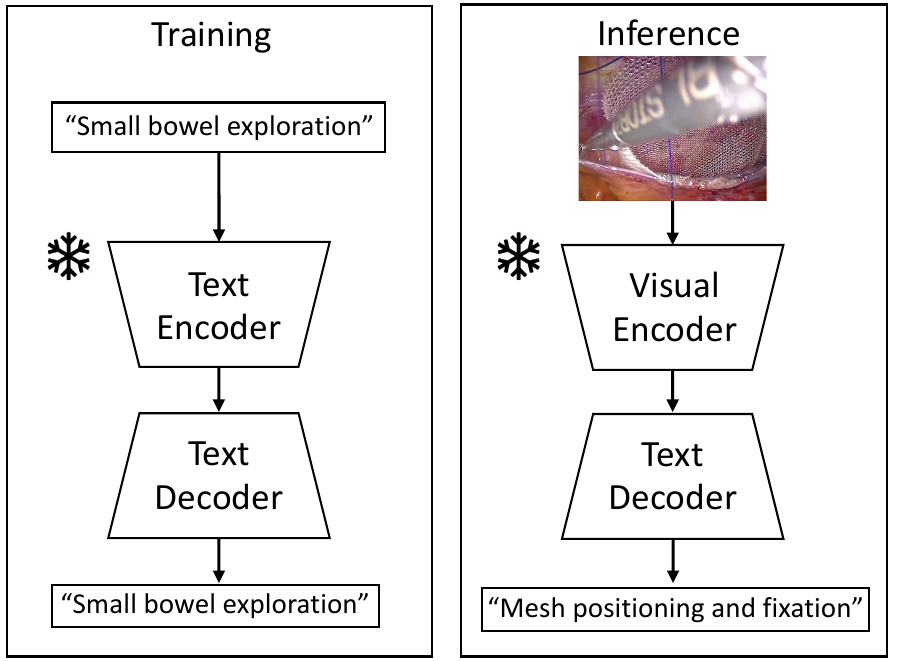}}
\caption{Text-only-training for video captioning: We use the learned joint embedding space where text is encoded in a representation close to the ones of its corresponding video clips. During training, we train the text decoder to generate captions from text embeddings. During inference, the visual embeddings are fed to the visual encoder and then to the text decoder to generate the text captions.}
\label{zero shot caption}
\end{figure}

\textbf{Text-only training for video captioning}. The above retrieval-based tasks essentially aim to measure \textit{SurgVLP}'s multi-modal representations by its ability to understand and correlate the latent vectors of two different modalities, i.e., video and text. However, it is unknown if the learned representations can be adapted to a textual generation task, such as image or video captioning. 

The traditional computer vision approaches for image or video captioning require paired visual and textual input to train the model~\citep{chen2020generating}. In this work, we however propose to exploit the multi-modal knowledge of \textit{SurgVLP} for the task of video captioning by performing a text-only training process. This approach requires the minimal modifications to our pre-trained model and does not require any visual input for training. 

Our approach solely relies on textual data for training, achieved through the construction of a trainable \textit{text decoder} ~\citep{nukrai2022text}. During the training phase, this text decoder is appended to the frozen pre-trained text encoder and is trained as an encoder-decoder architecture to reconstruct the textual sentences from the textual latent vectors, which are generated from the \textit{SurgVLP}'s pre-trained \textit{text encoder}. During the inference phase, the visual latent vector from a given video clip is generated using the frozen pre-trained visual encoder and fed into the trained text decoder to generate the corresponding text captions. Thanks to the multi-modal knowledge learned through \textit{SurgVLP}, it allows us to effectively pass the latent vectors of video clips to the text decoder. The training and inference phase is illustrated in Fig.~\ref{zero shot caption}. 

To generate meaningful captions for video clips, instead of using inherently noisy transcribed textual sentences from pre-training, we opt for using \textit{metadata sentences} from the surgical video lectures. These metadata sentences are uploaded by the content creator and typically describe different phases of the surgical procedure in a given video. These sentences are shorter and more abstract such as \textit{``creation of the gastric pouch''} or \textit{``trocar and port placement''} instead of a transcribed sentence such as \textit{``I think the important thing is that we have a good view at the beginning I perform the dissection with electrocautery''}. 

These metadata sentences show a significantly different textual distribution than the pre-training sentences. Therefore, using this task, we aim to assess the effective alignability of our learned multi-modal representations into a different textual domain. We use the metadata sentences from the videos in the pre-training dataset to train the \textit{text decoder}; for the evaluation, we use a separate set of $100$ videos containing video clips and corresponding metadata sentences.


\subsection{Vision-only surgical tasks}

We evaluate our approach on traditional vision-only surgical tasks using the Cholec80 ~\citep{twinanda2016endonet} and CholecT45 datasets~\citep {nwoye2021rendezvous}. Through these evaluations, we aim to assess the domain generalization and robustness to distribution shift of \textit{SurgVLP}, as these datasets contain only laparoscopic cholecystectomy videos. In the following, we describe these tasks and outline our approach to solving them without relying on any ground-truth labels, i.e., in a zero-shot manner.  

\textbf{Surgical tool, phase and triplet recognition}.\label{zero-shot-workflow} We evaluate three recognition downstream tasks of different granularities: surgical tool and phase recognition as coarse-grained tasks and surgical action triplet recognition as fine-grained task. 

\textit{SurgVLP} is pre-trained to predict if a video clip and a text snippet are paired together in the joint latent space. We reuse this capability by processing the name of all classes as potential text snippets and predicting whether a video clip and class name are paired. Instead of feeding class names directly to \textit{SurgVLP}, we manually create contextual prompts for the names of all classes in Cholec80 and CholecT45. Given the contextual prompts, we compute their text latent vectors using the frozen text encoder. We feed the input testing image to the visual encoder to compute the visual latent vector and compare it to the text latent vectors using cosine similarity. We assign the most probable class to the image using the one with the highest cosine similarity score, as shown in Fig.~\ref{pipeline}(c).

The zero-shot evaluation on these datasets presents numerous challenges. First, manually defining the class names for a coarse-grained surgical phase introduces ambiguity. For example, the Cholec80 dataset manually defines the class name \textit{``clipping and cutting''}, which does not accurately describe the surgical definition. In the clipping and cutting phase, the surgeon uses the clipper to clip the cystic duct and artery, then use scissors to cut them. Therefore, directly feeding class names into our pre-trained model increases confusion and degrades performance. Second, there exists a significant textual gap between the downstream and the pre-training datasets. Specifically, annotators usually define class as a single word in the downstream dataset. However, pre-training text transcriptions are usually complete sentences. A video clip rarely accompanies a single word in our pre-training dataset.

To address the above issues, we perform prompt engineering. We create contextual prompts from category names. For surgical phase recognition, we create contextual prompts for each phase label using a sentence describing the current phase, which mainly includes instruments, anatomies, and actions, as shown in Table~\ref{table:prompt}. For surgical tool recognition, we create a phrase for each label. The phrase describes the main action performed by the instrument. For example, as shown in  Table~\ref{table:prompt}, the category label \textit{``hook''} in tool recognition is transformed to \textit{``I use hook to dissect it''}, the category label \textit{``grasper''} is transformed to \textit{``I use grasper or cautery forceps to grasp it''}, and so on.  For surgical action triplet recognition, we use a prompt template \textit{``I use \{tool\} to \{action\} \{target\}''}, where we replace the value in the bracket with corresponding labels. For example, \triplet{grasper, grasp, gut} is transformed into \textit{``I use grasper to grasp the gut''}. These generated prompts help bridge the textual gap between the pre-training dataset and the downstream dataset.

\section{Experiments}
\label{sec:experiments}

We first describe the implementation details of our method in section~\ref{impl_details} and the pre-training dataset to train our model in section~\ref{pretrain-dataset}. We present the downstream tasks and datasets used in our zero-shot evaluation in section~\ref{downstream-dataset}. We then show an ablation study emphasizing key ingredients of our approach in section~\ref{ablation-study}. We compare our \textit{SurgVLP} approach to previous self-supervised methods in section~\ref{main-result}. Finally, we show the effect of contextual prompts and different text encoders in section~\ref{prompt-textencoder}.

\begin{table*}[!t]
	\centering
	\caption{\small{{Comparison of different datasets in this work. Human: if the dataset requires intervention by human annotators. SVL-Caption and SVL-Retrieval require partial intervention because texts are not annotated from scratch by human annotators.}}}
	\scalebox{0.95}{
		\begin{tabular}{ccccccc}
			\toprule
                & Media & Language Type & \# Videos & Human & Splits & Task \\
                \midrule
                SVL-Pretrain  & Video & Transcript & 1326 & No & Train/Val  & Vision-language Pretraining  \\
                SVL-Retrieval & Video & Transcript & 20 & Yes & Test & Text-based Video Retrieval  \\
                SVL-Caption & Video & Keystep texts & 998 & Yes & Train/Val/Test & Text-only Training for Video Captioning  \\
                CholecT45 & Image & Categorical texts & 50 & Yes & Train/Val/Test & Action Triplet Recognition  \\
                Cholec80 & Image & Categorical texts & 80 & Yes & Train/Val/Test & Tool/Phase Recognition  \\

                \bottomrule\hline
		\end{tabular} 
	}
	\label{tab:data comparison}
	\vspace{-1mm}
\end{table*}

\subsection{Implementation details} 
\label{impl_details}

{We describe the multi-modal feature extraction process and the detailed hyper-parameter setup of the proposed SurgVLP in section \ref{sec:approach}.}

\textbf{Feature extraction.} We use the ResNet-50 network pre-trained on ImageNet as the backbone of the visual encoder. We uniformly sample $T=4$ frames for each video clip. We first extract the embedding vector for each frame and then aggregate these vectors by temporal average pooling. Finally, we obtain a $d=768$ dimensional vector for each video clip. For each sentence transcribed from ASR systems, we tokenize it and then pad or truncate it to obtain the length $N=77$. {This length is selected to accommodate most sentences without excessive padding, ensuring consistent input sizes for the text encoder}. We feed the tokenized sentence into the text encoder and generate a $d=768$ dimensional vector as the text embedding.

\textbf{Hyper-parameters.} {Our SurgVLP model is based on contrastive learning and influenced by hyper-parameters such as batch size, learning rate, temperature, and loss coefficient. We selected the maximum batch size that fits our GPU environment and conducted a grid search for the other hyper-parameters.} We implement our method using the Pytorch framework. We set the batch size as $B=80$ and the learning rate as $0.0001$. {The batch size $B=80$ and frames $T=4$ are chosen to optimize GPU memory usage and training speed}. The temperature parameter $\tau$ is set to $0.3$ {to moderately control the distribution sharpness, aiding in effective learning of cosine similarities}. We use {one} sentence from AWS and $M=2$ sentences from Whisper as corresponding texts for each video clip. The coefficient weight for two loss functions $\epsilon$ is set to $0.5$. {Setting $\epsilon=0.5$ provides equal weighting to the two loss functions, ensuring balanced optimization without biasing towards one loss over the other.} The pre-training takes approximately $3$ days on four A100 GPUs.

\subsection{Pre-training dataset}
\label{pretrain-dataset}

We propose the \textbf{S}urgical \textbf{V}ideo \textbf{L}ecture Pre-training dataset (SVL-Pretrain) for multi-modal representation learning. {As shown in Table~\ref{tab:data comparison}, the SVL-Pretrain dataset contains large amounts of laparoscopic surgical video lectures.} Specifically, we query the videos based on keywords from three online platforms, i.e., Websurg~\citep{WebSurgt99:online}, EAES~\citep{eaes_2023}, and YouTube~\citep{1YouTube5:online}. For Websurg, we use ``intervention'' and ``laparoscopic'' as keywords to crawl $1,124$ surgical lectures. They cover the five main categories, ``Hepatobiliary and pancreatic surgery'', ``General and digestive surgery'', ``Bariatric surgery'', ``Hernia surgery'' and ``Colorectal, transanal and proctological surgery''. For EAES and YouTube, we manually collect $202$ surgical video lectures using manually designed keywords. Specifically, we compile a list of surgical keywords for specific surgical procedures, such as ``laparoscopic cholecystectomy'', ``hernia repairing'' and ``gastric bypass''. Then we form text queries by adding prefixes and suffixes to the keywords, such as ``how to...'' and ``:101''. We use these text queries to retrieve surgical video lectures from EAES and YouTube. We keep the top $50$ retrieved results and exclude videos without any audio. We also remove the video lectures that only contain slides and textbook images. In total, we collect $1,326$ lectures for the pre-training set. 

\textbf{Text pre-processing.} \textit{SurgVLP} requires video clip-text pairs for the training. Therefore, we use the AWS Medical Transcribe~\citep{AWS} system and the Whisper~\citep{radford2022robust} model to transcribe audio into textual sentences from the surgical video lectures. We apply pre-processing strategies to remove noisy and unaligned texts. Specifically, we first truncate the sentences based on the stop symbols, such as ``,'', ``;'' and so on. We then filter out sentences that do not contain any noun or verb. Since each word transcribed from AWS Medical Transcribe has a confidence score ranging from $0$ to $1$, we apply the confidence-based filter to remove sentences with a low average confidence score. We set up the threshold to $0.4$. Finally, we apply a keyword filtering strategy to remove sentences that do not contain any useful surgical word for the downstream tasks. For the sentences transcribed from Whisper, we follow the above strategy except for the confidence-based and keyword-based filtering. This difference makes the AWS sentences distribute more sparsely than Whisper sentences, providing the rationale for our hyper-parameter setup. For example, we select one AWS sentence and $M = 2$ Whisper sentences as correspondences. After pre-processing, each surgical video lecture has multiple transcribed sentences along the time axis. We then generate video clip-text pairs as mentioned in Sec. \ref{video-text}. This results in $25,578$ video clip-text pairs for the pre-training set.

\subsection{Downstream datasets and tasks}
\label{downstream-dataset}

{To show the generalizability of our learned representations, we perform the evaluation on five diverse downstream tasks using four datasets described below. As shown in Table~\ref{tab:data comparison}, we conduct both video and image-based downstream tasks. These tasks include text-based video retrieval on the SVL-Retrieval dataset, text-only training for video captioning on the SVL-Caption dataset, action triplet recognition on the CholecT45~\citep{nwoye2021rendezvous} dataset, and tool/phase recognition on the Cholec80~\citep{twinanda2016endonet} dataset. These datasets exhibit diversity in media type, language type, the number of videos, and whether human annotators are involved in annotation. These variations provide a comprehensive benchmark for testing the model's performance across both vision-and-language and vision-only tasks.}

\begin{table*}[t]
	\centering
        \setlength{\tabcolsep}{10pt}
	\caption{\small{Ablation studies. We conduct three sets of experiments to demonstrate the effect of key designs in our approach, multiple text views, clips of random lengths, and frame sampling from video clip. $\{v_i, a_i\}_{i=1}^K$: model trained with one AWS text view; $\{v_i, w_i^m\}_{i=1}^K$: model trained with one Whisper text view; $\{v_i, a_i, w_i^m\}_{i=1}^K$: model trained with both text views. {Random: Selecting a video clip with a duration randomly chosen from the range of 2 to 10 seconds.}}}
		\begin{tabular}{c|c|c|c|c}
			\toprule
	         & & \textbf{Phase Recognition} & \textbf{Triplet Recognition} & \textbf{Text-based Video Retrieval} \\ 
                & & Mean AP & Mean AP & Recall@10\\

			\midrule
                \multirow{3}{*}{Text Views} & $\{v_i, a_i\}_{i=1}^K$ & 25.5 & \textbf{11.0} & {21.6}\\
                & $\{v_i, w_i^m\}_{i=1}^K$ & 23.5 & 9.1 & {14.4} \\ 
                & $\{v_i, a_i, w_i^m\}_{i=1}^K$ & \textbf{26.7} & 10.3 & \textbf{25.1} \\ 
                \bottomrule\hline
		
                \multirow{4}{*}{Clip Length} & 2 seconds & {20.0} & 9.4 & 19.1\\
                & 4 seconds & 24.1 & {9.2} & {21.6}\\ 
                & 10 seconds & 17.3 & 9.0 & 18.8\\ 
                & Random & \textbf{26.7} & \textbf{10.3} & \textbf{25.1} \\
                \bottomrule\hline
                
                \multirow{3}{*}{Frame Sampling} & 1 frame & 19.7 & 9.5 & 18.1\\
                & 2 frames & {19.8} & {9.7} & {20.6}\\ 
                & 4 frames & \textbf{26.7} & \textbf{10.3} & \textbf{25.1}
                \\ 
                \bottomrule\hline
            \end{tabular}
	\label{tab:ablation}
	\vspace{-1mm}
\end{table*}

\begin{table*}[!t]
	\centering
	\caption{\small{Comparison of different methods in text-based video retrieval and temporal activity grounding tasks. }}
	\scalebox{1.0}{
		\begin{tabular}{cccccccc}
			\toprule
			\multirow{2}{*}{\textbf{Method}} & \multicolumn{4}{c}{\textbf{Text-based Video Retrieval}} & \multicolumn{3}{c}{\textbf{Temporal Grounding}} \\
                 \cmidrule(lr){2-5}\cmidrule(lr){6-8}
                ~ & 
                R$@$1 $(\%)$ & 
                R$@$5 $(\%)$&
                R$@$10 $(\%)$&
                Median Rank &
                R$@$1 $(\%)$ & 
                R$@$5 $(\%)$&
                R$@$10 $(\%)$
                \\
                \midrule
                 Random & 0.0 & 0.2 & 0.8 & 322 & 4.2 & 8.7 & 13.0\\
                CLIP~\citep{radford2021learning} & 0.0 & 0.9 & 2.1 & 256 & 4.2 & 7.8 & 13.9 \\
                CLIP$*$ & 0.6 & 1.2 & 2.3 & 244 & 4.3 & 9.3 & 15.3 \\
                SurgVLP & \textbf{2.8} & \textbf{11.8} & \textbf{16.1} & \textbf{70} & \textbf{8.6} & \textbf{19.9} & \textbf{29.7}\\
                \bottomrule\hline
		\end{tabular} 
	}
	\label{tab:retrieval and grounding}
	\vspace{-1mm}
\end{table*}

\subsubsection{Vision-and-language datasets and tasks}

\begin{table*}[!t]
	\centering
        \setlength{\tabcolsep}{10pt}
	\caption{\small{{SVL-Retrieval dataset. We show the categorical tags of the videos in the SVL-Retrieval testing set. Each video can belong to multiple categories, reflecting the diverse range of surgical procedures included in the testing set.}}}
	\scalebox{0.85}{
		\begin{tabular}{c|p{2.5cm}|p{2.5cm}|p{2.5cm}|p{2.5cm}|p{2.5cm}|p{2.5cm}}
			\toprule
                Tag & Laparoscopic & Stomach and duodenum & Hiatal hernia, reflux & General and digestive & Fundoplication & Upper GI surgery \\
			\midrule
                \#Videos &12 &10 &6 &4 &4 &3 \\
			\midrule

                Tag &Nissen-Rossetti fundoplication &Nissen fundoplication &Hernia surgery &Hepatobiliary and pancreatic surgery &Colorectal, transanal and proctological surgery & Colon \\
			\midrule
                \#Videos &3 &3 &3 &3 &3 &3 \\
                \midrule

                Tag &Cancer &Tumor &Robotic surgery &Liver &Laparoscopic sigmoidectomy for diverticulitis &Cyst and tumor \\
                \#Videos &3 &2 &2 &2 &2 &2 \\ 
                \midrule

                Tag &Wedge resection &Voluminous hiatal hernia &Toupet fundoplication &Sigmoiditis &Robotic cholecystectomy &Right hepatectomy \\
			\midrule
                \#Videos &1 &1 &1 &1 &1 &1 \\
                \midrule

                Tag &Peritonitis &Nissen procedure &Morbid obesit &Liver resection &Segmental gastrectomy &Gastric banding \\
			\midrule
                \#Videos &1 &1 &1 &1 &1 &1 \\
                \midrule

                Tag &Distal gastrectomy &Injury &Hinchey &Glissonian &Gastric cancer &Gastrectomy \\
			\midrule
                \#Videos &1 &1 &1 &1 &1 &1 \\
                \midrule
                
                Tag &Gallbladder &Bariatric surgery &Appendix &Appendicitis &Appendectomy & \\
                \midrule
                \#Videos &1 &1 &1 &1 &1 & \\

                \bottomrule\hline
		\end{tabular}
	}
	\label{tab:svl-retrieval}
	\vspace{-1mm}
\end{table*}

\textbf{Text-based video retrieval.} We construct a dataset, SVL-Retrieval, to evaluate \textit{SurgVLP}'s performance on text-based video retrieval tasks. Specifically, SVL-Retrieval contains $20$ surgical video lectures. Each video lecture is split into multiple video clip-text pairs, resulting in $537$ pairs. {As shown in Table~\ref{tab:svl-retrieval}, the testing videos in SVL-Retrieval are diverse, with multiple tags per video. Predominantly comprising laparoscopic surgeries, the SVL-Retrieval dataset highlights significant diversity with videos focusing on the stomach, duodenum, hernia, colon, gallbladder, and tumor surgeries. The content within each surgical type varies widely, showcasing distinct educational, technique, and case-specific aspects across the videos.} The text queries in SVL-Retrieval are initially transcribed from the Whisper ASR system. Therefore, the transcribed sentences are inherently noisy and miss surgery-specific terms. To make a representative evaluation, our clinical collaborator has validated and cleaned the text queries by correcting the typos and punctuation errors.

We follow the same evaluation protocol as described in~\citep{zhukov2019cross} to report the retrieval performance using the Recall@K metric (with $K=1,5,10$), which measures the percentage of correct video clips present within the K top-ranked clips that were retrieved (the higher, the better). {In our retrieval testing set, we have 537 video-text pairs that serve as the ground truth. For each text description, we compute the cosine similarity between the text and all 537 videos in the dataset. This process generates a ranked list of the top-K retrieved videos based on similarity. For each text description, if the ground truth video is among the top-K retrieved videos, we increment the Recall@K by one. This indicates that the correct corresponding video has been successfully identified within the top-K results. This process is repeated for all 537 text descriptions in the testing set. We compute the average Recall@K value across all 537 text descriptions to assess the overall performance of the retrieval system.} We also report the median rank of the video clips in the video pool. During the retrieval, each video clip in the SVL-Retrieval dataset receives a ranking score that indicates how close it is to the corresponding text query, compared to the other video clips in the SVL-Retrieval dataset (the lower, the better). The median rank score is the median of the ranks of the ground truth videos across all text queries. A lower median rank score indicates a stronger overall correlation between the video clips and corresponding text queries, thus demonstrating the effectiveness of the retrieval system.

\textbf{Temporal activity grounding.} We use the same SVL-Retrieval dataset to evaluate the performance of temporal activity grounding. We use the Recall@K metric (with $K=1,5,10$) to report the performance. Here, we do not apply the median rank in the temporal activity grounding task. This is because temporal grounding is a video-specific task and does not yield a large enough ranking list to represent the performance of the method accurately.

\textbf{Text-only training for video captioning.}\label{text-only} We create a video captioning dataset, called SVL-Caption, to perform text-only training for video captioning. The SVL-Caption dataset is comprised of a training set with $9,074$ \textit{metadata sentences} obtained from the videos in the SVL-Pretrain dataset. The testing set of SVL-Caption contains $734$ video clip-caption pairs from a separate set of $100$ videos obtained using the same collection process as described in section~\ref{pretrain-dataset} but not used during the pre-training. As explained in section~\ref{vision-and-language surgical tasks}, the metadata sentences, which are uploaded by the video creator, describe different phases in the surgical video lecture.

We utilize BLEU~\citep{papineni2002bleu}, METEOR~\citep{banerjee2005meteor}, and ROUGE~\citep{lin2004rouge} metrics to evaluate the quality of generated captions. The BLEU metric measures n-gram overlap, with higher scores for more matches. The METEOR metric enhances this by considering synonymy, stemming, and exact word matches. Finally, the ROUGE metric evaluates n-gram overlap, longest common subsequence, and skip-bigrams. All three metrics effectively assess the capability of the video captioning model to deliver concise, informative summaries encapsulating key information.

\subsubsection{Vision-only tasks}

\textbf{Tool recognition.} We use the publicly available Cholec80 dataset~\citep{twinanda2016endonet} to evaluate the surgical tool recognition. We use the test split of this dataset. We apply the average precision ($AP$) to measure the recognition performance for surgical tools as the area under the precision-recall curve. We report the tool-wise $AP$ and the average $AP$ across classes. 

\textbf{Phase recognition.} We use the same Cholec80 dataset test split as above to evaluate the surgical phase recognition. As precision and recall are commonly used in prior works~\citep{twinanda2016endonet,czempiel2020tecno}, we choose F1-Score as the balanced metric for precision and recall for the phase recognition task. It ranges from $0$ to $1$, where $1$ indicates perfect precision and recall.

\textbf{Triplet recognition.} We use CholecT45 dataset~\citep{nwoye2021rendezvous} to evaluate the surgical action triplet recognition. We use the test split of this dataset to make the comparison to the prior fully-supervised and self-supervised works. {For the triplet recognition task}, we utilize the average precision ($AP$) metric to recognize individual components, i.e., instrument ($AP_i$), verb ($AP_v$), and target ($AP_t$). We also evaluate the performance of instrument-tissue interactions by looking at the metrics for different sets of triplet components: the instrument-verb ($AP_{iv}$), instrument-target ($AP_{it}$), and instrument-verb-target ($AP_{ivt}$).

\subsection{Ablation studies}
\label{ablation-study}

We perform our ablation studies on the following downstream tasks: phase recognition (Mean AP), triplet recognition (Mean AP), and text-based video retrieval (Recall@10). This subset of downstream tasks has been chosen for their simplicity of evaluation and because they cover a wide range of tasks. As the hyper-parameter decisions, such as number of text views, are made based on the results of the ablation studies, we conduct the evaluation on the validation subsets for all these three tasks. In the following, we give the key takeaways from our study.

\textbf{Multiple text views introduce complementary knowledge.} Here, we remove one of the textual inputs from the video clip-text pairs $\{v_i, a_i, w_i^m\}_{i=1}^{K}$. When we train the model with sentences transcribed only from the AWS system, we obtain video clip-text pairs $\{v_i, a_i\}_{i=1}^{K}$. When we train the model with sentences transcribed only from the Whisper system, we obtain video clip-text pairs $\{v_i, w_i^m\}_{i=1}^{K}$, as shown in Table~\ref{tab:ablation}. 

As mentioned in section~\ref{video-text}, we hypothesize that combining AWS Medical Transcribe and Whisper ASR systems will introduce complementary knowledge. In Table~\ref{tab:ablation}, we prove these hypotheses by showing that the model trained from \textit{SurgVLP} outperforms its counterparts in phase recognition and text-based video retrieval tasks where the language and boundary information play the most important role. The improvement can be attributed to the fact that sentences transcribed by two ASR systems serve as two augmented versions of the ``ground truth text'', helping the model to learn transform-invariant textual features. We also find that a model trained with sentences transcribed only by AWS leads to better performance in action triplet recognition. We believe this downstream task does not require sophisticated language understanding. Instead, detecting and matching the main keywords is sufficient to solve this task.

\textbf{Different clip lengths affect different granular-level tasks.} Here, we use both AWS and Whisper text views to pre-train the model and demonstrate the effect of the different lengths for video clip sampling. We conclude that different lengths of video clips affect the performance of different downstream tasks. Also, randomly growing video clips into different lengths provide better robustness to different granular-level downstream tasks.

\begin{figure*}[!t]
\centerline{\includegraphics[width=2\columnwidth]{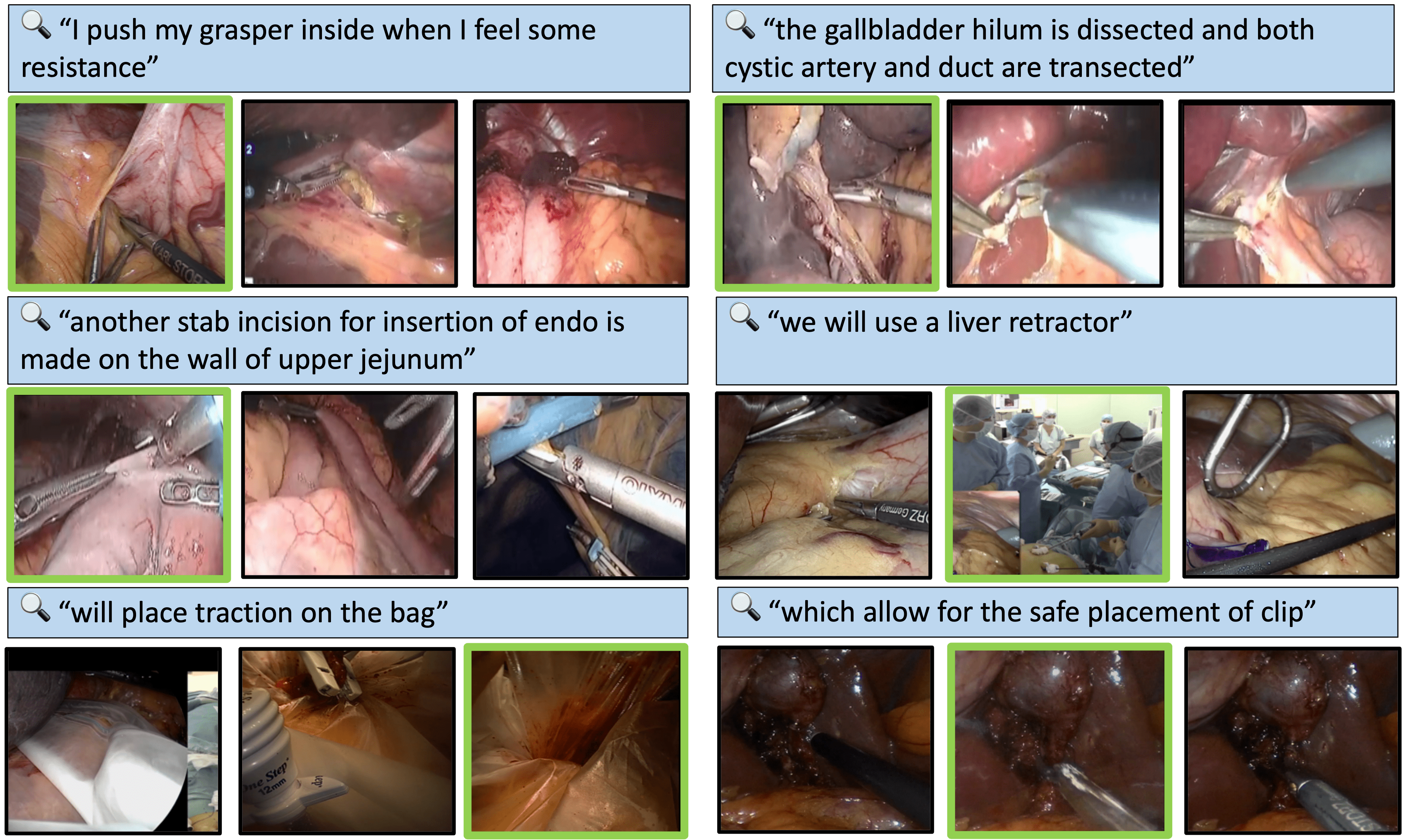}}
\caption{Qualitative results of text-based video retrieval on SVL-Retrieval dataset using \textit{SurgVLP}'s learned joint multi-modal representations. For each language query, we retrieve $3$ video clips from the repository. The ground truth video clip is framed in \textcolor{green}{green}. It is here always mentioned in the top-3 results.}
\label{retrieval fig}
\end{figure*}

\begin{figure*}[!t]
\centerline{\includegraphics[width=2\columnwidth]{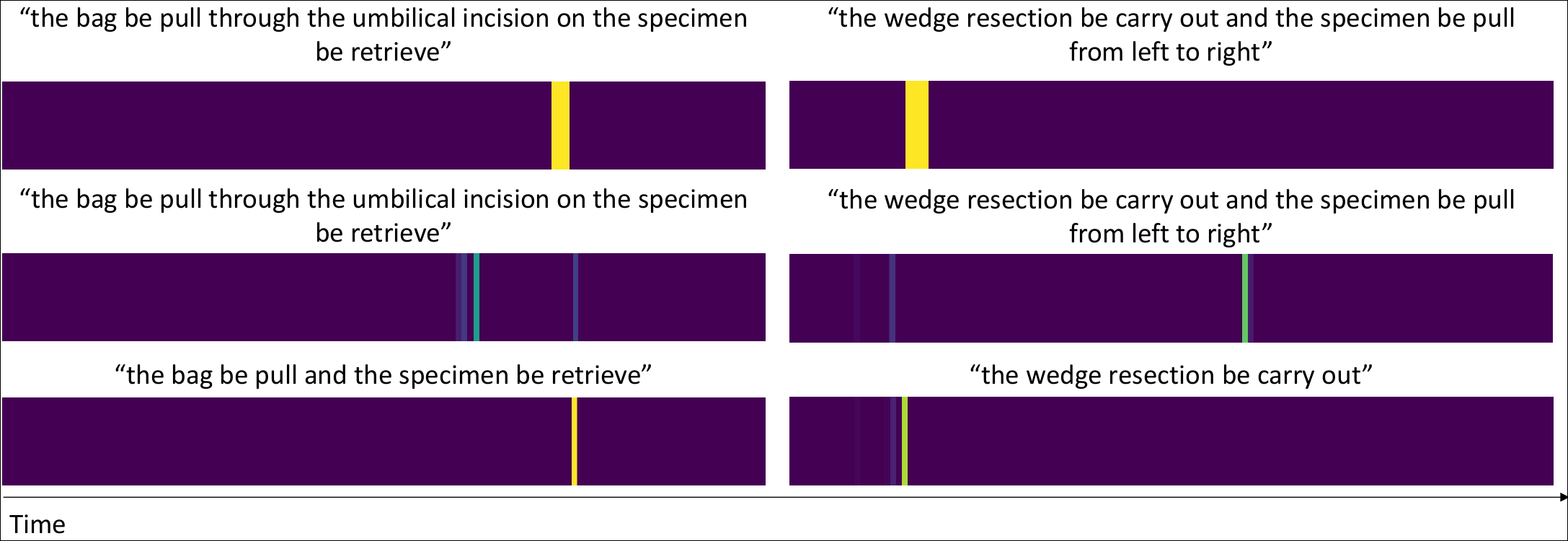}}
\caption{{Textual-visual activation maps from different sentence queries. The first row shows the ground truth. The second row shows the predicted activation map along the time axis for the raw sentence. The third row shows the newly generated activation maps conditioned by modified sentences. When the whole sentence is decomposed into sub-sentences, the \textit{SurgVLP} approach generates a focused textual-visual activation map for the sentence with clear and less ambiguous words. This shows that \textit{SurgVLP} responds to specific surgical terms rather than general terminology.}}
\label{activatioin map2}
\end{figure*}

Among all three tasks, action triplet recognition is a fine-grained task. Phase recognition and text-based video retrieval are coarse-grained. As shown in Table~\ref{tab:ablation}, longer clip lengths improve performance on coarse-grained tasks. For example, the phase recognition and text-based video retrieval tasks are significantly improved when the clip length is increased from $2s$ to $4s$. However, extremely long clip lengths bring no benefit and hinder performance. This is because longer clips provide more redundant frames that are not semantically correlated with the texts, resulting in weakly aligned video clip-text pairs. To cover the tasks with different granularities, we sample the video clips with different lengths (up to 10s), which gives the best results for all three tasks. We fix the video clip sampling strategy for the rest of the paper by growing the video clips with random lengths.

\begin{figure*}[!t]
\centerline{\includegraphics[width=2.1\columnwidth]{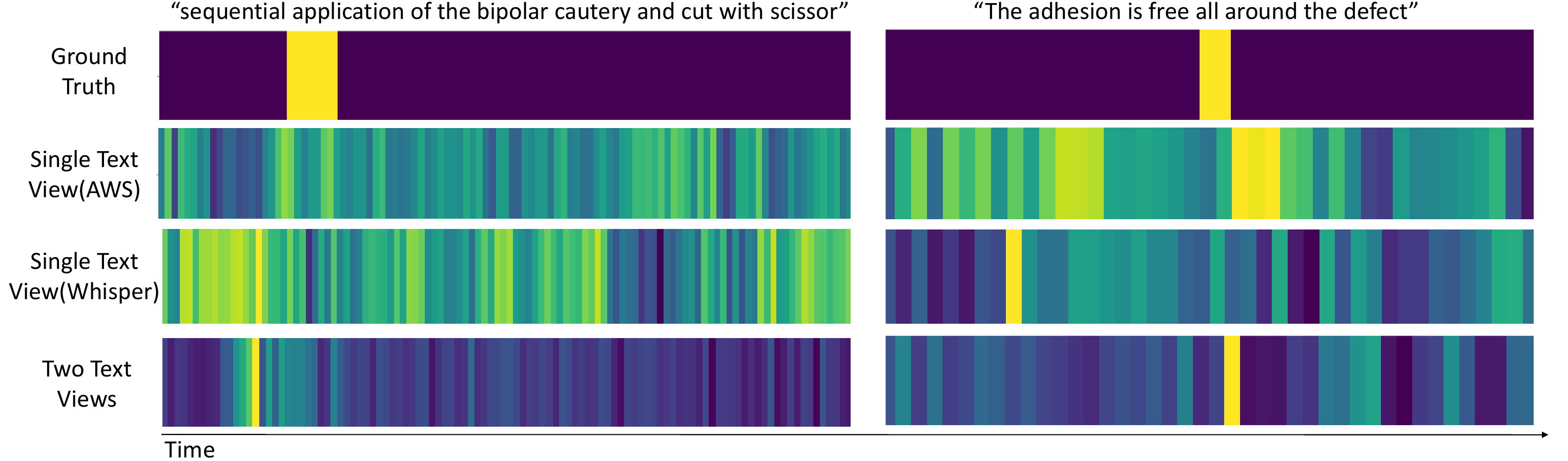}}
\caption{{Textual-visual activation maps of the \textit{SurgVLP} model, computed on two language queries from SVL-Retrieval testing set. The language queries are shown at the top of the figure, and the first row shows the ground truth activation map. The second and the third row shows the activation maps of \textit{SurgVLP} trained with one text view, i.e., AWS texts and Whisper texts, respectively. The last row shows that when the \textit{SurgVLP} model is trained on both AWS and Whisper texts, it generates more concrete activation maps with less noise.}}
\label{activation map1}
\end{figure*}

\begin{figure*}[!t]
\centering
\includegraphics[width=2\columnwidth]{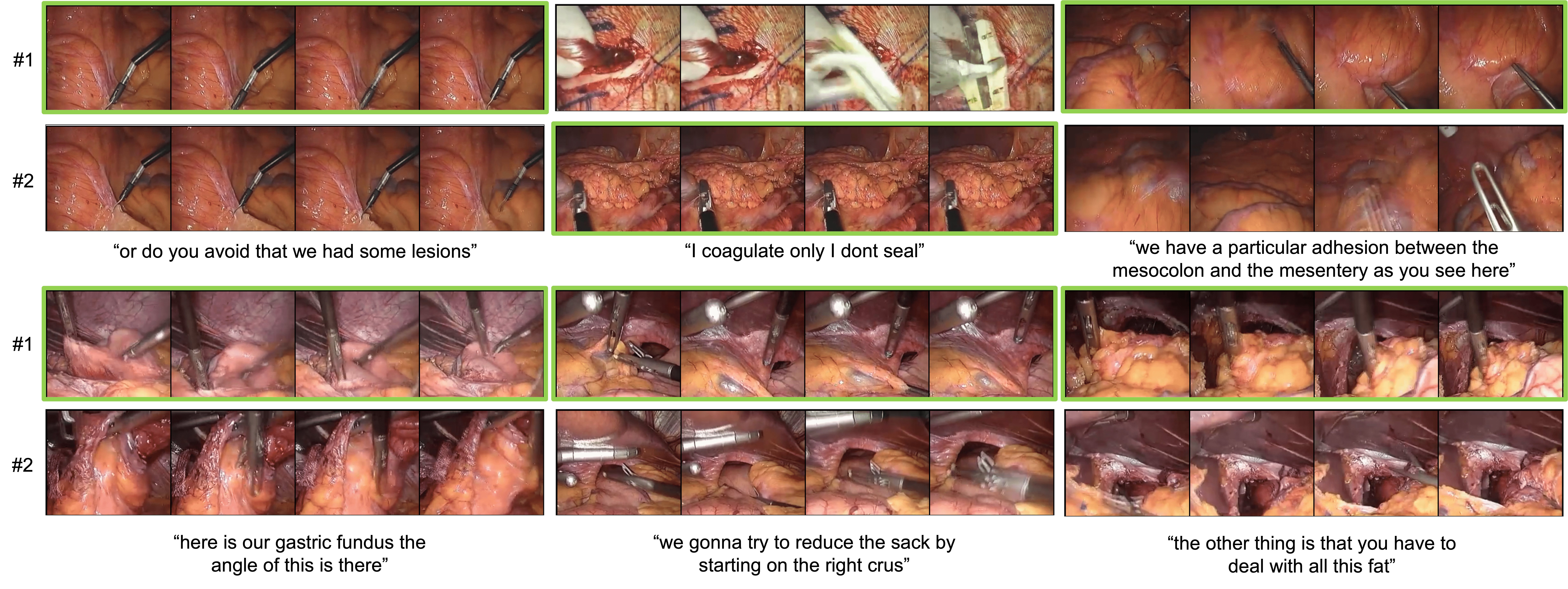}
\caption{Qualitative results of temporal activity grounding. We show the grounding results of two videos with three language queries. Each set of images represents a video clip. We show top-2 grounded clips for given text queries. Video clips framed in \textcolor{green}{green} are the ground truth to the given text. $\#1$: top-1 grounded result. $\#2$: top-2 grounded result.}
\label{temporal grounding fig}
\end{figure*}

\textbf{More frames, the better.} We continue to use two text views and the video clips of random lengths to evaluate the quality of our multi-modal representations trained with different numbers of frames. As shown in Table~\ref{tab:ablation}, performance is improved when the model includes more frames from the clip. Specifically, the performance of triplet recognition increases from $9.5$ to $10.3$ when three additional frames are used in the pre-training. This suggests that including more visual information from video clips could further improve the performance of the joint representation. Given the limited computational resources, we sample up to $4$ frames for each video clip in the current work.

\subsection{Main results and comparisons}
\label{main-result}

We build three baselines for comparisons with our \textit{SurgVLP} approach from different perspectives. One trivial system is a randomly initialized \textit{SurgVLP}, denoted as \textbf{Random}, to present it as a lower baseline. We initialize it randomly $3$ times and present the average results. We also take the publicly available \textbf{CLIP}~\citep{radford2021learning} model as a competitive baseline because it is trained on $400$ million of image-text pairs and shows promising zero-shot image classification results on the ImageNet benchmark. We further load the CLIP model's weight and fine-tune it with our SVL-Pretrain dataset. We denote the resulting model as \textbf{CLIP$*$}.

\subsubsection{Vision-and-language tasks}

This section analyzes \textit{SurgVLP}'s zero-shot transfer to multiple vision-and-language downstream tasks.

\textbf{Text-based video retrieval.} Quantitatively, as shown in Table~\ref{tab:retrieval and grounding},  \textit{SurgVLP} surpasses the baselines by a large margin. Compared to the publicly available CLIP~\citep{radford2021learning} model, multi-modal representations from our \textit{SurgVLP} approach correlate the surgical narrative description to video clips. It also links the instrument, action, and activity descriptions of different granularities to the corresponding video clips. Qualitatively, Fig.~\ref{retrieval fig} illustrates the top three retrieved video clips from our proposed SVL-Retrieval dataset. The ground truth is framed in \textcolor{green}{green}. It shows that \textit{SurgVLP} can capture specific medical terms like bag and liver retractor.

\begin{figure*}[!t]
\centerline{\includegraphics[width=2\columnwidth]{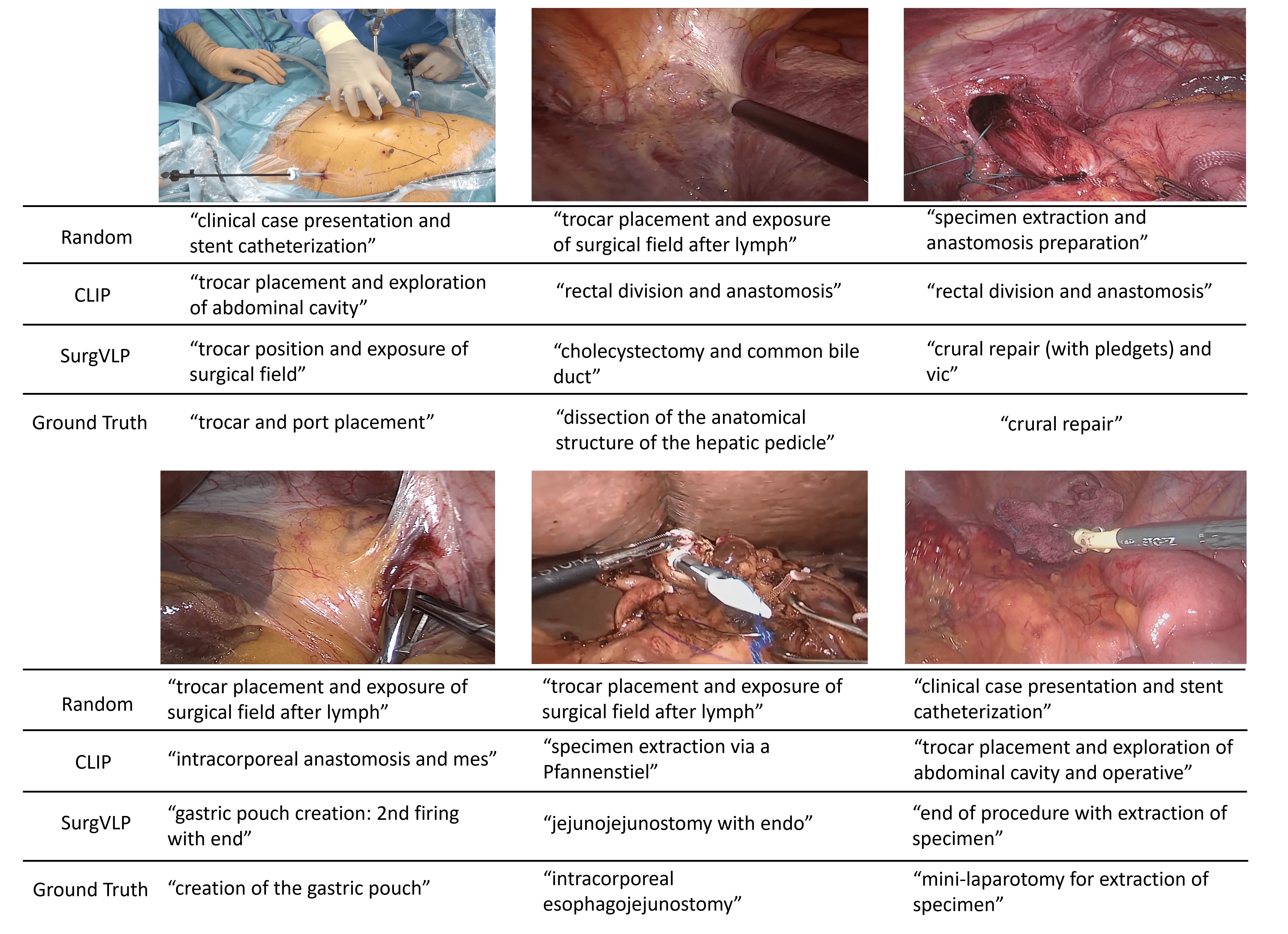}}
\caption{Caption results from text-only training for video captioning. Random: randomly initialized \textit{SurgVLP}. CLIP~\citep{radford2021learning}: publicly available joint embedding space from OpenAI pre-trained CLIP model. \textit{SurgVLP} shows more reliable captioning results with more overlap to the ground truth sentence. Also, the \textit{SurgVLP} approach can generate detailed captions with the surgical instrument mentioned, e.g. ``pledgets'' in the top row last column.}
\label{caption results}
\end{figure*}

\begin{table*}[!t]
	\centering
        \setlength{\tabcolsep}{10pt}
	\caption{\small{Quantitative results of text-only training for video captioning. We report $6$ conventional metrics to measure the similarity between generated text and ground text. Our proposed \textit{SurgVLP} significantly outperforms previous work, especially for ROUGE, which requires an accurate representation of not only individual words but also their correct order.}}
	\scalebox{1.0}{
		\begin{tabular}{c|cccccc}
			\toprule
			\textbf{Methods} & \textbf{BLEU-1} & \textbf{BLEU-2} & \textbf{BLEU-3} & \textbf{BLEU-4} & \textbf{METEOR} & \textbf{ROUGE}\\
			\midrule
                Random & 0.19 & 0.02 & 0.01 & 0 & 0.04 & 0.22 \\
                CLIP~\citep{radford2021learning}  & 0.21 & 0.04 & 0.02 & 0.01 & 0.04 & 0.24 \\
                SurgVLP & \textbf{0.36} & \textbf{0.22} & \textbf{0.17} & \textbf{0.12} & \textbf{0.18} & \textbf{0.33}\\\bottomrule\hline
		\end{tabular}
	}
	\label{tab:caption quati}
	\vspace{-1mm}
\end{table*}

\textbf{Temporal activity grounding.} We then evaluate our learned multi-modal representations on the temporal activity grounding task in Table~\ref{tab:retrieval and grounding}. It shows that \textit{SurgVLP} outperforms the public CLIP model despite its use of $400$ million images for pre-training. We believe that this improved localization ability is due to the fact that our model is trained on surgical narrative descriptions with temporally grounded video clips as opposed to image-level annotations in CLIP~\citep{radford2021learning}.

Fig.~\ref{activatioin map2} shows the impact of novel text queries on the temporal activity grounding task. Specifically, we simulate the open-vocabulary scenario by decomposing the sentence into sub-sentences. {For example, we construct a clear and concise text query by removing the sub-sentence \textit{``through umbilical incision''} from the complete sentence \textit{``the bag be pull through the umbilical incision on the specimen be retrieve''}, resulting in text query \textit{``the bag be pull and the specimen be retrieve''}. We plot the visual-textual activation map by comparing the text embedding to all visual frame embeddings of the video. Also, as shown in Fig.~\ref{activatioin map2} right, the combination of two sentences will generate the decentralized activation map that spreads the attention to different timestamps, indicating that the \textit{SurgVLP} model is triggered to the unambiguous semantics, and removing the redundant text query can improve the temporal activity grounding performance.} As shown in Fig.~\ref{activation map1}, \textit{SurgVLP} trained with two text views generates a focused attention map, highlighting relevant frames and ignoring adjacent ones. This demonstrates that multiple text view supervision effectively connects the visual and the textual content. Compared to the AWS text view, the model pretrained with Whisper text view generates the attention map with focused areas when the text query is about the long-term video activity, such as ``sequential application of bipolar cautery and cut with scissor''. {Fig.~\ref{activation map1} also demonstrates that incorporating multiple text views helps generate textual activation maps with reduced noise, suggesting a more refined multi-modal embedding space that better aligns images with semantically similar language descriptions. However, the observed misalignment with ground truth highlights a key limitation: temporal grounding depends on accurate boundary detection, which our pretrained video-text pairs cannot fully provide. This discrepancy arises because spoken descriptions often precede corresponding visual actions, creating an inherent latency between audio transcriptions and visual content. As a result, the activation maps may exhibit shifts in alignment, reflecting this temporal gap.}

We also show the qualitative results for the temporal activity grounding task. As shown in the first column of the top row of Fig.~\ref{temporal grounding fig}, the system clearly understands the concept of \textit{``lesion''} to locate the correct video clip. As shown in the second column of the top row, it grounds the \textit{``seal''} action correctly but fails to ground the \textit{``coagulate''} action. 

\begin{table*}[!t]
	\centering
        \setlength{\tabcolsep}{10pt}
	\caption{\small{Zero-shot tool recognition on Cholec80. T1: grasper; T2: bipolar; T3: hook; T4: scissor; T5: clipper; T6: irrigator; T7: specimen bag. Fully-supervised: ResNet50 model with full supervision.}}
	\scalebox{1.0}{
		\begin{tabular}{c|cccccccc}
			\toprule
			\textbf{Methods}     & \textbf{T1}   & \textbf{T2}   & \textbf{T3}   & \textbf{T4}   & \textbf{T5}   & \textbf{T6}   & \textbf{T7}   &\textbf{Mean} \\ 
			\midrule
                Fully-supervised      & 86.9 & 79.6 & 99.2 & 81.2 & 94.9 & 61.3 & 93.8 & 85.3 \\
                \cdashline{1-9}
                Random      & 48.3 & 6.1  & 57.7 & 2.4  & 5.0  & 4.6  & 11.5  & 19.3 \\
                CLIP~\citep{radford2021learning} & 60.7 & 4.3 & 56.0 & 1.9 & 2.5 & 4.6 & 2.8 & 22.6 \\
                CLIP$*$& \textbf{61.7} & 5.1 & 57.2 & 1.6 & 2.7 & 3.5 & 7.5 & 23.2\\

                SurgVLP     & 55.2 & \textbf{21.3} & \textbf{61.4} & \textbf{7.8}  & \textbf{16.2} & \textbf{7.9}  & \textbf{64.9} & \textbf{33.3} \\
                \bottomrule\hline
		\end{tabular}
	}
	\label{zero_shot_c80_tool}
	\vspace{-1mm}
\end{table*}

\begin{table*}[!t]
	\centering
        \setlength{\tabcolsep}{10pt}

	\caption{\small{Zero-shot phase recognition on Cholec80. P1: preparation; P2: calot triangle dissection; P3: clipping and cutting; P4: gallbladder dissection; P5: gallbladder packing; P6: cleaning and coagulation; P7: gallbladder extraction. F1-Score is used as the evaluation metric. Fully-supervised: ResNet50 model with full supervision.}}
	\scalebox{1.0}{
		\begin{tabular}{c|cccccccc}
			\toprule
			\textbf{Methods}     & \textbf{P1}   & \textbf{P2}   & \textbf{P3}   & \textbf{P4}   & \textbf{P5}   & \textbf{P6}   & \textbf{P7}   &\textbf{Mean} \\ 
			\midrule
                Fully-supervised      & 62.1 & 84.5 & 74.4 & 82.2 & 62.7 & 52.2 & 52.2 & 67.3 \\
                \cdashline{1-9}
                Random      & 5.6    & 0.3    & 4.7    & 3.4    & 6.5    & 1.0    & 4.4 & 1.8  \\
                CLIP~\citep{radford2021learning} & 10.3 & 38.0 & 0 & 20.1 & 0.3 & 3.7 & 0.4 & 10.4 \\
                CLIP$*$& 9.0 & 38.7 & 0 & \textbf{20.3} & 2.3 & 3.6 & 14.7 & 11.5\\
                
                SurgVLP  & \textbf{30.9} & \textbf{58.2} & \textbf{11.6}  & 11.9 & \textbf{34.5} & \textbf{6.0}  & \textbf{15.5} & \textbf{24.0}\\

                \bottomrule\hline
		\end{tabular}
	}
	\label{zero_shot_c80_phase}
	\vspace{-1mm}
\end{table*}

\begin{table*}[!t]
	\centering
        \setlength{\tabcolsep}{13pt}
	\caption{\small{Zero-shot triplet recognition results. We report the average precision for each component and the combination of the components. i: instrument, v: verb, t: target, iv: instrument-verb, it: instrument-target, ivt: instrument-verb-target triplet.}}
	\scalebox{1.0}{
		\begin{tabular}{c|cccccc}
			\toprule
			\textbf{mAP (\%)}     & \textbf{i}   & \textbf{v}   & \textbf{t}   & \textbf{iv}   & \textbf{it}   & \textbf{ivt} \\ 
			\midrule
                    Fully-supervised~\citep{nwoye2021rendezvous} & 89.5 & 63.0 & 45.2 & 38.7 & 37.5 & 30.8\\
                    \cdashline{1-7}
                    Random & 22.7 & 14.9 & 10.9 & 5.0 & 4.3 & 3.2\\
                    
                    CLIP~\citep{radford2021learning} & 24.4 & 15.3& 10.3 & 7.2 & 4.0& 3.1 \\
                    CLIP$*$ & 25.7 & 15.8 & 11.0 & 7.7 & 4.9 & 3.7 \\
                    
                    SurgVLP & \textbf{32.6} & \textbf{22.8} & \textbf{17.1} & \textbf{10.8} & \textbf{8.6} & \textbf{7.0}\\\bottomrule\hline
                    
		\end{tabular}
	}
	\label{zero-shot triplet}
	\vspace{-1mm}
\end{table*}

\textbf{Text-only training for video captioning.} We conduct text-only training for video captioning, demonstrating that the model pre-trained from the \textit{SurgVLP} approach can be applied for text generation tasks. We compare \textit{SurgVLP} to two baselines, the randomly initialized \textit{SurgVLP} and CLIP. Here, all methods are trained and evaluated using the SVL-Caption dataset. As shown in Table~\ref{tab:caption quati}, \textit{SurgVLP} performs better than the other baselines, as it is trained with surgical video clip-text pairs. Random initialization yields the lowest results across all metrics because the visual and textual embeddings are not correlated. When a semantically similar video clip is given to this model, the model does not align it to the corresponding textual embedding, thus generating incorrect text output. Our proposed \textit{SurgVLP} approach scores highest in all the metrics, suggesting the effect of its learned multi-modal embedding space. We also show qualitative results in Fig.~\ref{caption results} to demonstrate that \textit{SurgVLP} can accurately generate abstract metadata sentences from a given surgical video clip.

\subsubsection{Vision-only tasks}
This section analyzes \textit{SurgVLP}'s zero-shot transfer to different vision-only surgical downstream tasks such as surgical tool, phase, and triplet recognition on laparoscopic cholecystectomy.

\textbf{Tool recognition}. As depicted in Table~\ref{zero_shot_c80_tool}, \textit{SurgVLP} exhibits promising tool recognition performance, especially in identifying tools with unique shapes like the T7 (specimen bag) with a $64.9\%$ average precision (AP). This zero-shot performance is close to a fully supervised ResNet-50 model, as shown in the first row. Nevertheless, there remains ample opportunity for improvement in recognizing more challenging tools, such as scissors and irrigators. These tools are rarely mentioned in the SVL-Pretrain dataset. Also, we find that CLIP$*$ gives a slight improvement but still lags behind \textit{SurgVLP}. This demonstrates that multi-modal pre-training on natural images and texts hampers surgical tool recognition tasks due to the large domain gap between natural and surgical objects.

\begin{figure*}[!t]
\centerline{\includegraphics[width=2\columnwidth]{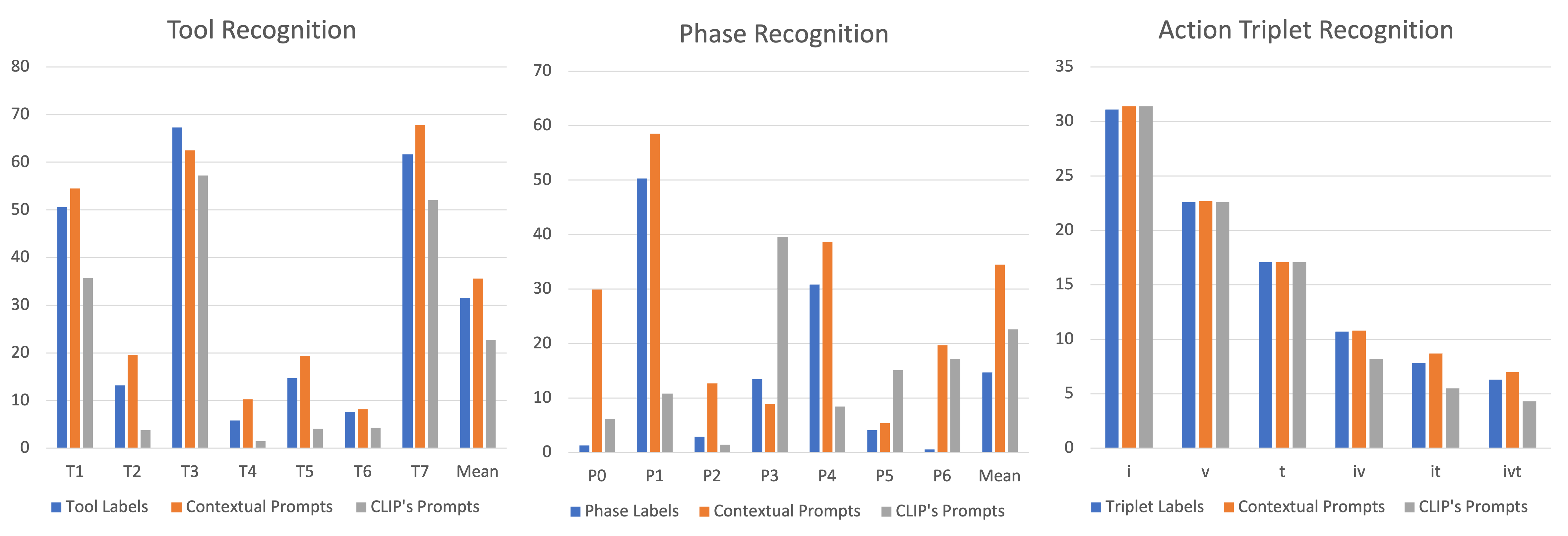}}
\caption{Effect of our designed contextual prompts to the zero-shot transfer of vision-only downstream tasks. Our contextual prompts outperform their counterparts by encoding more specific action and anatomy information, thus boosting phase recognition and instrument-verb recognition.}
\label{prompt_effect}
\end{figure*}

\begin{figure}[!t]
\centering
\includegraphics[width=1\columnwidth]{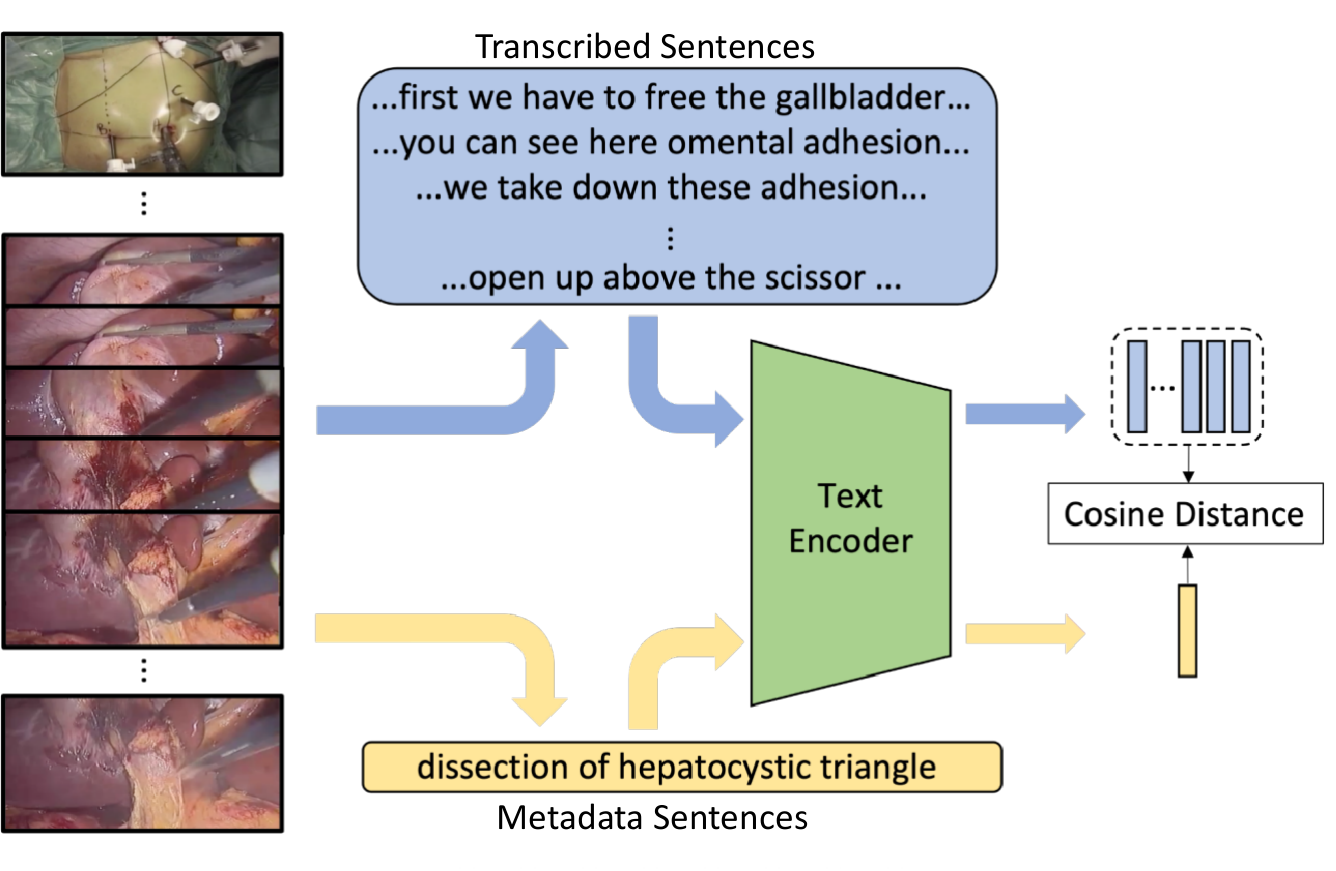}
\caption{Text architecture selection. We calculate the cosine similarity score between the transcript texts from ASR and pre-segment texts from metadata to measure which text encoder retains the semantic information between these two texts.}
\label{text evaluate}
\end{figure}

\begin{table}[t!]
	\centering
	\caption{\small{Cosine similarity scores comparison among different text encoders.}}
	\scalebox{0.85}{
		\begin{tabular}{c|c}
			\toprule
			\textbf{Models} & \textbf{Similarity Score} \\
			\midrule
                Word2Vec Meanpooling & 0.11 \\
                Word2Vec Maxpooling & 0.56 \\
                Bert~\citep{devlin2018bert} & 0.63 \\
                CLIP's Text Encoder~\citep{radford2021learning} & 0.69 \\
                SciBert~\citep{beltagy2019scibert} & 0.80 \\                BioClinicalBert~\citep{huang2019clinicalbert} & 0.86\\\bottomrule\hline
		\end{tabular}
	}
	\label{table text encoder}
	\vspace{-1mm}
\end{table}

\textbf{Phase recognition}. As shown in Table~\ref{zero_shot_c80_phase}, \textit{SurgVLP} outperforms the baseline methods by a large margin. This indicates that our surgery-specific pre-training method helps to recognize surgical phases without explicitly training for them. Also, \textit{SurgVLP} and CLIP$*$ largely outperform the vanilla CLIP~\citep{radford2021learning} at P7 (gallbladder extraction) and P5 (gallbladder packing), which involves specimen-bag during the operation. This is consistent with the improvement in recognizing T7 (specimen bag), indicating that T7 contains generic knowledge that can be transferred across different procedures. {We assess the statistical significance between our \textit{SurgVLP} and CLIP$*$ models by comparing their F1-scores. We train \textit{SurgVLP} three times with different random seeds and perform inference on the test set. An independent t-test comparing the F1-scores of \textit{SurgVLP} and CLIP$*$ yields a t-statistic of $5.03$ and a p-value of $0.0037$, indicating a significant difference (p $< 0.05$). These results confirm that the performance improvement of \textit{SurgVLP} over the baseline is statistically significant and unlikely due to chance.}

\textbf{Triplet recognition}. As mentioned in~\citep{cole2022does}, contrastive learning lags far behind supervised learning on ﬁne-grained visual classiﬁcation tasks. Therefore, we conduct the action triplet recognition task to study a similar question for surgical computer vision. We aim to assess whether the multi-modal features of \textit{SurgVLP} are only effective for ``easy'' classiﬁcation tasks or are also useful for more challenging
``ﬁne-grained'' visual concepts. As shown in Table~\ref{zero-shot triplet}, \textit{SurgVLP} outperforms the baseline methods in zero-shot triplet recognition by enhancing the recognition of combinations of instruments, verbs, and targets in cholecystectomy. Specifically, we find that \textit{SurgVLP} significantly improves verb and target recognition performance. This suggests that the model effectively learns better anatomical and action knowledge from our SVL-Pretrain dataset and can transfer the learned knowledge to procedure-specific downstream tasks. The zero-shot performance is still lagging behind the fully-supervised method by a large margin, and we expect this could be addressed by scaling the dataset.

\subsubsection{Text encoder and prompt analysis}
\label{prompt-textencoder}

Finally, we present the results of two experiments. First, we conduct experiments to identify an appropriate text encoder by exploring various text encoders for their ability to understand surgery-specific texts. Second, we study the impact of different types of prompts on the vision-only downstream tasks for zero-shot evaluations. In the following, we give two key takeaways from these two experiments. 

\textbf{BioClinicalBert text encoder excels at encoding surgery-specific texts.} Conventional text encoders like Word2Vec~\citep{mikolov2013efficient} show difficulties in handling the out-of-vocabulary surgical terms. Also, a large textual domain gap exists between the surgical datasets and conventional computer vision datasets. Therefore, we provide a study demonstrating which text encoders can generate representative text embeddings for multi-modal representation learning in the surgical field. Specifically, we consider Bert~\citep{lu2019vilbert} pre-trained on a large-scale web corpus, SciBert~\citep{beltagy2019scibert} pre-trained on the bio-medical scientific papers, CLIP's text encoder~\citep{radford2021learning} pre-trained on large-scale natural image-caption pairs, and Word2Vec~\citep{mikolov2013efficient} with different pooling strategies, to compare with \textit{SurgVLP}'s text encoder, i.e., BioClinicalBert~\citep{huang2019clinicalbert}.

We select the video clips from the SVL-Caption dataset. Each clip is associated with a metadata sentence. Given the start and end timestamps of this sentence, we extract the transcribed sentences within the video clip. We assume that the metadata sentences and their corresponding transcribed sentences are paired and have high similarity. Therefore, we feed the sentences into the text encoders mentioned above to generate text embeddings. We then calculate the cosine similarity score between the corresponding metadata and transcription text embeddings, as reported in Table~\ref{table text encoder}. Overall, BioClinicalBert achieves the best results compared to the other text encoders. We show that the bag-of-words approach, e.g., Word2Vec, achieves the lowest score because transcribed sentences from the SVL-Pretrain dataset require complex textual understanding, which Word2Vec fails to capture. Also, as CLIP's text encoder is trained on a web corpus that contains large amounts of natural texts, it shows the difficulty in handling surgical-specific texts. We use BioClinicalBert in our \textit{SurgVLP} approach as it outperforms all other text encoders in this experiment.

\textbf{Contextual prompts aligned to downstream tasks result in better performance.} To bridge the textual gap between the pre-training and downstream datasets, we manually design the contextual prompts for class names. We use these prompts as text inputs to perform the zero-shot transfer experiments on vision-only downstream tasks. We compare our contextual prompts, explained in section \ref{zero-shot-workflow}, to two variants: the textual form of the class names and CLIP's default prompts. The CLIP's default prompt follows a template \textit{``the photo of \{class name\}''} where the class name is replaced with the phase, tool, or triplet class name. As shown in Fig.~\ref{prompt_effect}, our designed contextual prompts improve the performance on all three tasks. For example, the $P0$ phase, which is ``preparation'', is significantly improved. Our contextual prompts also outperform CLIP's prompts in terms of instrument-verb, instrument-target, and instrument-verb-target recognition. This shows that our proposed contextual prompts encode more action and anatomical linguistic information to benefit fine-grained vision-only downstream tasks.

\section{Discussion and Conclusion}

\subsection{Discussion}

\subsubsection{Future Work}
This work shows that the proposed SurgVLP achieves superior zero-shot performance than the state-of-the-art methods~\citep{radford2021learning} from the general computer vision field. The better performance is enabled due to the constructed large-scale surgical vision-language dataset and pretraining strategy with multiple text views. However, the zero-shot adaptation of SurgVLP is not supervised by any annotated data, leading to a suboptimal performance compared to the fully supervised works~\citep{twinanda2016endonet,czempiel2020tecno}. A potential improvement toward real-world application is to adapt the learned multi-modal representations of the pretrained SurgVLP to the downstream tasks using less labeled data with fully-supervised finetuning. Specifically, the SurgVLP's dual-branch architecture can encode domain-specific textual knowledge while capturing detailed visual patterns from the surgical scene~\citep{kan2023knowledge}. Consequently, the feature extractor of fully supervised methods~\citep{twinanda2016endonet,czempiel2020tecno} could be boosted by the complementary information from the textual side. Another future work line is to explore ``cheaper'' self-supervision signals within the visual and textual modalities. Typical work includes building the external knowledge base~\citep{shen2022k} and performing retrieval-augmented vision-language pretraining~\citep{xie2023ra}. Witnessing the recent emergence of large language models~\citep{touvron2023llama} and its encoded clinical knowledge, exploring the usage of these language models by eliciting their knowledge can help to merge the domain gap. Furthermore, the current work overlooks the hierarchical structure inherent in surgical videos. To address this, hierarchical multi-modal pretraining can be incorporated to further improve the performance for the surgical downstream tasks requiring long temporal context for the prediction.

\subsubsection{{Limitation}}
{While SurgVLP reduces annotation burdens, its zero-shot performance remains below fully supervised methods, particularly in complex tasks requiring fine-grained anatomical reasoning or handling intraoperative variability. Also, reliance on error-prone ASR systems introduces transcription inaccuracies in surgical terminology and temporal misalignments between narrated steps and visual content, magnified by domain shifts between lecture demonstrations and real-world surgical environments. In addition, pretraining data lacks diversity for rare or specialized surgeries, while conventional accuracy metrics inadequately reflect clinical risks, e.g., error impacts on patient safety. Finally, computational costs may hinder deployment in resource-limited settings. These limitations position SurgVLP as a pretraining strategy to augment, but not replace, fully-supervised learning, requiring targeted fine-tuning and domain-adaptation methods for clinical applicability.}

\subsection{Conclusion}
The expensive and laborious process of creating manually annotated datasets has been a main hindrance in developing scalable surgical computer vision AI systems. In this work, we argue that surgical video lectures available through open surgical e-learning platforms can provide a wealth of multi-modal knowledge to train a scalable system for multi-modal representation learning. We have harnessed this knowledge by creating a multi-modal and multi-procedural dataset comprising 1.4k surgical video lectures. In order to derive effective supervisory signals without manual annotations, we utilize the recent advancements in automatic speech recognition (ASR) systems to transcribe the audio from these videos into textual descriptions. This automated process has resulted in a visual-textual multi-modal surgical dataset consisting of descriptions of surgical events, instrument usage, and anatomical status across various surgical procedures. 

In order to tackle the surgery-specific linguistic challenges inherently present in these videos, we utilize text transcriptions from two complementary ASR models, namely Whisper and AWS. The AWS model {captures} specific surgical terms, whereas the Whisper model {captures} the overall sentence structure. By combining the complementary knowledge of these two systems, we overcome inherent limitations and inaccuracies present in each ASR system. We then propose a novel contrastive learning objective for multi-modal representation learning. Our approach, called \textit{SurgVLP}, learns effective multi-modal representations by bringing embeddings of multiple text transcriptions and video clip to close proximity in the joint latent space.

To demonstrate the efficacy of the learned joint latent space, we present a range of vision-and-language tasks tailored for surgical computer vision. These tasks include text-based video retrieval, temporal activity grounding, and video captioning, serving as benchmarks for evaluating the multi-modal representation capability of \textit{SurgVLP}. We demonstrate that the learned multi-modal representations are not only useful for these vision-and-language tasks but can also be seamlessly applied to traditional vision-only surgical downstream tasks. We show promising results on these vision-only surgical tasks, namely surgical tool, phase, and triplet recognition, without using any manual annotations.

\section*{Acknowledgment}
This work has received funding from the European Union (ERC, CompSURG, 101088553). Views and opinions expressed are however those of the authors only and do not necessarily reflect those of the European Union or the European Research Council. Neither the European Union nor the granting authority can be held responsible for them. This work was also partially supported by French state funds managed by the ANR under Grants ANR-20-CHIA-0029-01 and ANR-10-IAHU-02. This work was granted access to the HPC resources of IDRIS under the allocation AD011013704R1 and AD011011631R2 made by GENCI. Jo\"el L. Lavanchy was funded by the Swiss National Science Foundation ($P500PM_206724$, $P5R5PM_217663$).

We also appreciate the tremendous effort put forth by the open education platforms Websurg (IRCAD), EAES, and YouTube to provide high-quality educational content and make it freely accessible to learners around the world. We are incredibly grateful to the clinicians who have devoted their time and expertise to create and share content on these platforms, making this work possible.

\bibliographystyle{model2-names.bst}\biboptions{authoryear}
\bibliography{refs}

\begin{thebibliography}{111}
\expandafter\ifx\csname natexlab\endcsname\relax\def\natexlab#1{#1}\fi
\providecommand{\url}[1]{\texttt{#1}}
\providecommand{\href}[2]{#2}
\providecommand{\path}[1]{#1}
\providecommand{\DOIprefix}{doi:}
\providecommand{\ArXivprefix}{arXiv:}
\providecommand{\URLprefix}{URL: }
\providecommand{\Pubmedprefix}{pmid:}
\providecommand{\doi}[1]{\href{http://dx.doi.org/#1}{\path{#1}}}
\providecommand{\Pubmed}[1]{\href{pmid:#1}{\path{#1}}}
\providecommand{\bibinfo}[2]{#2}
\ifx\xfnm\relax \def\xfnm[#1]{\unskip,\space#1}\fi
\bibitem[{Al~Hajj et~al.(2018)Al~Hajj, Lamard, Conze, Cochener and
  Quellec}]{al2018monitoring}
\bibinfo{author}{Al~Hajj, H.}, \bibinfo{author}{Lamard, M.},
  \bibinfo{author}{Conze, P.H.}, \bibinfo{author}{Cochener, B.},
  \bibinfo{author}{Quellec, G.}, \bibinfo{year}{2018}.
\newblock \bibinfo{title}{Monitoring tool usage in surgery videos using boosted
  convolutional and recurrent neural networks}.
\newblock \bibinfo{journal}{Medical image analysis} \bibinfo{volume}{47},
  \bibinfo{pages}{203--218}.
\bibitem[{Alapatt et~al.(2021)Alapatt, Mascagni, Vardazaryan, Garcia, Okamoto,
  Mutter, Marescaux, Costamagna, Dallemagne and Padoy}]{alapatt2021temporally}
\bibinfo{author}{Alapatt, D.}, \bibinfo{author}{Mascagni, P.},
  \bibinfo{author}{Vardazaryan, A.}, \bibinfo{author}{Garcia, A.},
  \bibinfo{author}{Okamoto, N.}, \bibinfo{author}{Mutter, D.},
  \bibinfo{author}{Marescaux, J.}, \bibinfo{author}{Costamagna, G.},
  \bibinfo{author}{Dallemagne, B.}, \bibinfo{author}{Padoy, N.},
  \bibinfo{year}{2021}.
\newblock \bibinfo{title}{Temporally constrained neural networks (tcnn): A
  framework for semi-supervised video semantic segmentation}.
\newblock \bibinfo{journal}{arXiv preprint arXiv:2112.13815} .
\bibitem[{Alayrac et~al.(2020)Alayrac, Recasens, Schneider, Arandjelovi{\'c},
  Ramapuram, De~Fauw, Smaira, Dieleman and Zisserman}]{alayrac2020self}
\bibinfo{author}{Alayrac, J.B.}, \bibinfo{author}{Recasens, A.},
  \bibinfo{author}{Schneider, R.}, \bibinfo{author}{Arandjelovi{\'c}, R.},
  \bibinfo{author}{Ramapuram, J.}, \bibinfo{author}{De~Fauw, J.},
  \bibinfo{author}{Smaira, L.}, \bibinfo{author}{Dieleman, S.},
  \bibinfo{author}{Zisserman, A.}, \bibinfo{year}{2020}.
\newblock \bibinfo{title}{Self-supervised multimodal versatile networks}.
\newblock \bibinfo{journal}{Advances in Neural Information Processing Systems}
  \bibinfo{volume}{33}, \bibinfo{pages}{25--37}.
\bibitem[{Allan et~al.(2019)Allan, Shvets, Kurmann, Zhang, Duggal, Su, Rieke,
  Laina, Kalavakonda, Bodenstedt et~al.}]{allan20192017}
\bibinfo{author}{Allan, M.}, \bibinfo{author}{Shvets, A.},
  \bibinfo{author}{Kurmann, T.}, \bibinfo{author}{Zhang, Z.},
  \bibinfo{author}{Duggal, R.}, \bibinfo{author}{Su, Y.H.},
  \bibinfo{author}{Rieke, N.}, \bibinfo{author}{Laina, I.},
  \bibinfo{author}{Kalavakonda, N.}, \bibinfo{author}{Bodenstedt, S.}, et~al.,
  \bibinfo{year}{2019}.
\newblock \bibinfo{title}{2017 robotic instrument segmentation challenge}.
\newblock \bibinfo{journal}{arXiv preprint arXiv:1902.06426} .
\bibitem[{Amodei et~al.(2016)Amodei, Ananthanarayanan, Anubhai, Bai,
  Battenberg, Case, Casper, Catanzaro, Cheng, Chen et~al.}]{amodei2016deep}
\bibinfo{author}{Amodei, D.}, \bibinfo{author}{Ananthanarayanan, S.},
  \bibinfo{author}{Anubhai, R.}, \bibinfo{author}{Bai, J.},
  \bibinfo{author}{Battenberg, E.}, \bibinfo{author}{Case, C.},
  \bibinfo{author}{Casper, J.}, \bibinfo{author}{Catanzaro, B.},
  \bibinfo{author}{Cheng, Q.}, \bibinfo{author}{Chen, G.}, et~al.,
  \bibinfo{year}{2016}.
\newblock \bibinfo{title}{Deep speech 2: End-to-end speech recognition in
  english and mandarin}, in: \bibinfo{booktitle}{International conference on
  machine learning}, \bibinfo{organization}{PMLR}. pp.
  \bibinfo{pages}{173--182}.
\bibitem[{AWS(2023)}]{AWS}
\bibinfo{author}{AWS}, \bibinfo{year}{2023}.
\newblock \bibinfo{title}{Amazon transcribe medical}.
\newblock \URLprefix \url{https://aws.amazon.com/transcribe/medical/}.
\bibitem[{Banerjee and Lavie(2005)}]{banerjee2005meteor}
\bibinfo{author}{Banerjee, S.}, \bibinfo{author}{Lavie, A.},
  \bibinfo{year}{2005}.
\newblock \bibinfo{title}{Meteor: An automatic metric for mt evaluation with
  improved correlation with human judgments}, in:
  \bibinfo{booktitle}{Proceedings of the acl workshop on intrinsic and
  extrinsic evaluation measures for machine translation and/or summarization},
  pp. \bibinfo{pages}{65--72}.
\bibitem[{Beltagy et~al.(2019)Beltagy, Lo and Cohan}]{beltagy2019scibert}
\bibinfo{author}{Beltagy, I.}, \bibinfo{author}{Lo, K.},
  \bibinfo{author}{Cohan, A.}, \bibinfo{year}{2019}.
\newblock \bibinfo{title}{Scibert: A pretrained language model for scientific
  text}.
\newblock \bibinfo{journal}{arXiv preprint arXiv:1903.10676} .
\bibitem[{Blum et~al.(2010)Blum, Feu{\ss}ner and Navab}]{blum2010modeling}
\bibinfo{author}{Blum, T.}, \bibinfo{author}{Feu{\ss}ner, H.},
  \bibinfo{author}{Navab, N.}, \bibinfo{year}{2010}.
\newblock \bibinfo{title}{Modeling and segmentation of surgical workflow from
  laparoscopic video}, in: \bibinfo{booktitle}{Medical Image Computing and
  Computer-Assisted Intervention--MICCAI 2010: 13th International Conference,
  Beijing, China, September 20-24, 2010, Proceedings, Part III 13},
  \bibinfo{organization}{Springer}. pp. \bibinfo{pages}{400--407}.
\bibitem[{Blum et~al.(2008)Blum, Padoy, Feu{\ss}ner and
  Navab}]{blum2008modeling}
\bibinfo{author}{Blum, T.}, \bibinfo{author}{Padoy, N.},
  \bibinfo{author}{Feu{\ss}ner, H.}, \bibinfo{author}{Navab, N.},
  \bibinfo{year}{2008}.
\newblock \bibinfo{title}{Modeling and online recognition of surgical phases
  using hidden markov models}, in: \bibinfo{booktitle}{Medical Image Computing
  and Computer-Assisted Intervention--MICCAI 2008: 11th International
  Conference, New York, NY, USA, September 6-10, 2008, Proceedings, Part II
  11}, \bibinfo{organization}{Springer}. pp. \bibinfo{pages}{627--635}.
\bibitem[{Brown et~al.(2020)Brown, Mann, Ryder, Subbiah, Kaplan, Dhariwal,
  Neelakantan, Shyam, Sastry, Askell et~al.}]{brown2020language}
\bibinfo{author}{Brown, T.}, \bibinfo{author}{Mann, B.},
  \bibinfo{author}{Ryder, N.}, \bibinfo{author}{Subbiah, M.},
  \bibinfo{author}{Kaplan, J.D.}, \bibinfo{author}{Dhariwal, P.},
  \bibinfo{author}{Neelakantan, A.}, \bibinfo{author}{Shyam, P.},
  \bibinfo{author}{Sastry, G.}, \bibinfo{author}{Askell, A.}, et~al.,
  \bibinfo{year}{2020}.
\newblock \bibinfo{title}{Language models are few-shot learners}.
\newblock \bibinfo{journal}{Advances in neural information processing systems}
  \bibinfo{volume}{33}, \bibinfo{pages}{1877--1901}.
\bibitem[{Changpinyo et~al.(2021)Changpinyo, Sharma, Ding and
  Soricut}]{changpinyo2021conceptual}
\bibinfo{author}{Changpinyo, S.}, \bibinfo{author}{Sharma, P.},
  \bibinfo{author}{Ding, N.}, \bibinfo{author}{Soricut, R.},
  \bibinfo{year}{2021}.
\newblock \bibinfo{title}{Conceptual 12m: Pushing web-scale image-text
  pre-training to recognize long-tail visual concepts}, in:
  \bibinfo{booktitle}{Proceedings of the IEEE/CVF Conference on Computer Vision
  and Pattern Recognition}, pp. \bibinfo{pages}{3558--3568}.
\bibitem[{Chen et~al.(2020a)Chen, Kornblith, Norouzi and
  Hinton}]{chen2020simple}
\bibinfo{author}{Chen, T.}, \bibinfo{author}{Kornblith, S.},
  \bibinfo{author}{Norouzi, M.}, \bibinfo{author}{Hinton, G.},
  \bibinfo{year}{2020}a.
\newblock \bibinfo{title}{A simple framework for contrastive learning of visual
  representations}, in: \bibinfo{booktitle}{International conference on machine
  learning}, \bibinfo{organization}{PMLR}. pp. \bibinfo{pages}{1597--1607}.
\bibitem[{Chen et~al.(2020b)Chen, Fan, Girshick and He}]{chen2020improved}
\bibinfo{author}{Chen, X.}, \bibinfo{author}{Fan, H.},
  \bibinfo{author}{Girshick, R.}, \bibinfo{author}{He, K.},
  \bibinfo{year}{2020}b.
\newblock \bibinfo{title}{Improved baselines with momentum contrastive
  learning}.
\newblock \bibinfo{journal}{arXiv preprint arXiv:2003.04297} .
\bibitem[{Chen et~al.(2015)Chen, Fang, Lin, Vedantam, Gupta, Doll{\'a}r and
  Zitnick}]{chen2015microsoft}
\bibinfo{author}{Chen, X.}, \bibinfo{author}{Fang, H.}, \bibinfo{author}{Lin,
  T.Y.}, \bibinfo{author}{Vedantam, R.}, \bibinfo{author}{Gupta, S.},
  \bibinfo{author}{Doll{\'a}r, P.}, \bibinfo{author}{Zitnick, C.L.},
  \bibinfo{year}{2015}.
\newblock \bibinfo{title}{Microsoft coco captions: Data collection and
  evaluation server}.
\newblock \bibinfo{journal}{arXiv preprint arXiv:1504.00325} .
\bibitem[{Chen et~al.(2022a)Chen, Du, Hu, Liu, Li, Wan and
  Chang}]{chen2022multi}
\bibinfo{author}{Chen, Z.}, \bibinfo{author}{Du, Y.}, \bibinfo{author}{Hu, J.},
  \bibinfo{author}{Liu, Y.}, \bibinfo{author}{Li, G.}, \bibinfo{author}{Wan,
  X.}, \bibinfo{author}{Chang, T.H.}, \bibinfo{year}{2022}a.
\newblock \bibinfo{title}{Multi-modal masked autoencoders for medical
  vision-and-language pre-training}, in: \bibinfo{booktitle}{Medical Image
  Computing and Computer Assisted Intervention--MICCAI 2022: 25th International
  Conference, Singapore, September 18--22, 2022, Proceedings, Part V},
  \bibinfo{organization}{Springer}. pp. \bibinfo{pages}{679--689}.
\bibitem[{Chen et~al.(2020c)Chen, Song, Chang and Wan}]{chen2020generating}
\bibinfo{author}{Chen, Z.}, \bibinfo{author}{Song, Y.}, \bibinfo{author}{Chang,
  T.H.}, \bibinfo{author}{Wan, X.}, \bibinfo{year}{2020}c.
\newblock \bibinfo{title}{Generating radiology reports via memory-driven
  transformer}, in: \bibinfo{booktitle}{Proceedings of the 2020 Conference on
  Empirical Methods in Natural Language Processing (EMNLP)}, pp.
  \bibinfo{pages}{1439--1449}.
\bibitem[{Chen et~al.(2022b)Chen, Zhang, Rosenberg, Ramabhadran, Moreno, Bapna
  and Zen}]{chen2022maestro}
\bibinfo{author}{Chen, Z.}, \bibinfo{author}{Zhang, Y.},
  \bibinfo{author}{Rosenberg, A.}, \bibinfo{author}{Ramabhadran, B.},
  \bibinfo{author}{Moreno, P.}, \bibinfo{author}{Bapna, A.},
  \bibinfo{author}{Zen, H.}, \bibinfo{year}{2022}b.
\newblock \bibinfo{title}{Maestro: Matched speech text representations through
  modality matching}.
\newblock \bibinfo{journal}{arXiv preprint arXiv:2204.03409} .
\bibitem[{Cole et~al.(2022)Cole, Yang, Wilber, Mac~Aodha and
  Belongie}]{cole2022does}
\bibinfo{author}{Cole, E.}, \bibinfo{author}{Yang, X.},
  \bibinfo{author}{Wilber, K.}, \bibinfo{author}{Mac~Aodha, O.},
  \bibinfo{author}{Belongie, S.}, \bibinfo{year}{2022}.
\newblock \bibinfo{title}{When does contrastive visual representation learning
  work?}, in: \bibinfo{booktitle}{Proceedings of the IEEE/CVF Conference on
  Computer Vision and Pattern Recognition}, pp. \bibinfo{pages}{14755--14764}.
\bibitem[{Czempiel et~al.(2020)Czempiel, Paschali, Keicher, Simson, Feussner,
  Kim and Navab}]{czempiel2020tecno}
\bibinfo{author}{Czempiel, T.}, \bibinfo{author}{Paschali, M.},
  \bibinfo{author}{Keicher, M.}, \bibinfo{author}{Simson, W.},
  \bibinfo{author}{Feussner, H.}, \bibinfo{author}{Kim, S.T.},
  \bibinfo{author}{Navab, N.}, \bibinfo{year}{2020}.
\newblock \bibinfo{title}{Tecno: Surgical phase recognition with multi-stage
  temporal convolutional networks}, in: \bibinfo{booktitle}{Medical Image
  Computing and Computer Assisted Intervention--MICCAI 2020: 23rd International
  Conference, Lima, Peru, October 4--8, 2020, Proceedings, Part III 23},
  \bibinfo{organization}{Springer}. pp. \bibinfo{pages}{343--352}.
\bibitem[{Czempiel et~al.(2021)Czempiel, Paschali, Ostler, Kim, Busam and
  Navab}]{czempiel2021opera}
\bibinfo{author}{Czempiel, T.}, \bibinfo{author}{Paschali, M.},
  \bibinfo{author}{Ostler, D.}, \bibinfo{author}{Kim, S.T.},
  \bibinfo{author}{Busam, B.}, \bibinfo{author}{Navab, N.},
  \bibinfo{year}{2021}.
\newblock \bibinfo{title}{Opera: Attention-regularized transformers for
  surgical phase recognition}, in: \bibinfo{booktitle}{Medical Image Computing
  and Computer Assisted Intervention--MICCAI 2021: 24th International
  Conference, Strasbourg, France, September 27--October 1, 2021, Proceedings,
  Part IV 24}, \bibinfo{organization}{Springer}. pp. \bibinfo{pages}{604--614}.
\bibitem[{Deng et~al.(2009)Deng, Dong, Socher, Li, Li and
  Fei-Fei}]{deng2009imagenet}
\bibinfo{author}{Deng, J.}, \bibinfo{author}{Dong, W.},
  \bibinfo{author}{Socher, R.}, \bibinfo{author}{Li, L.J.},
  \bibinfo{author}{Li, K.}, \bibinfo{author}{Fei-Fei, L.},
  \bibinfo{year}{2009}.
\newblock \bibinfo{title}{Imagenet: A large-scale hierarchical image database},
  in: \bibinfo{booktitle}{2009 IEEE conference on computer vision and pattern
  recognition}, \bibinfo{organization}{Ieee}. pp. \bibinfo{pages}{248--255}.
\bibitem[{Dergachyova et~al.(2016)Dergachyova, Bouget, Huaulm{\'e}, Morandi and
  Jannin}]{dergachyova2016automatic}
\bibinfo{author}{Dergachyova, O.}, \bibinfo{author}{Bouget, D.},
  \bibinfo{author}{Huaulm{\'e}, A.}, \bibinfo{author}{Morandi, X.},
  \bibinfo{author}{Jannin, P.}, \bibinfo{year}{2016}.
\newblock \bibinfo{title}{Automatic data-driven real-time segmentation and
  recognition of surgical workflow}.
\newblock \bibinfo{journal}{International journal of computer assisted
  radiology and surgery} \bibinfo{volume}{11}, \bibinfo{pages}{1081--1089}.
\bibitem[{Devlin et~al.(2018)Devlin, Chang, Lee and Toutanova}]{devlin2018bert}
\bibinfo{author}{Devlin, J.}, \bibinfo{author}{Chang, M.W.},
  \bibinfo{author}{Lee, K.}, \bibinfo{author}{Toutanova, K.},
  \bibinfo{year}{2018}.
\newblock \bibinfo{title}{Bert: Pre-training of deep bidirectional transformers
  for language understanding}.
\newblock \bibinfo{journal}{arXiv preprint arXiv:1810.04805} .
\bibitem[{EAES(2023)}]{eaes_2023}
\bibinfo{author}{EAES}, \bibinfo{year}{2023}.
\newblock \bibinfo{title}{The european association of endoscopic surgery}.
\newblock \URLprefix \url{https://eaes.eu/}.
\bibitem[{Eisenmann et~al.(2022)Eisenmann, Reinke, Weru, Tizabi, Isensee,
  Adler, Godau, Cheplygina, Kozubek, Ali et~al.}]{eisenmann2022biomedical}
\bibinfo{author}{Eisenmann, M.}, \bibinfo{author}{Reinke, A.},
  \bibinfo{author}{Weru, V.}, \bibinfo{author}{Tizabi, M.D.},
  \bibinfo{author}{Isensee, F.}, \bibinfo{author}{Adler, T.J.},
  \bibinfo{author}{Godau, P.}, \bibinfo{author}{Cheplygina, V.},
  \bibinfo{author}{Kozubek, M.}, \bibinfo{author}{Ali, S.}, et~al.,
  \bibinfo{year}{2022}.
\newblock \bibinfo{title}{Biomedical image analysis competitions: The state of
  current participation practice}.
\newblock \bibinfo{journal}{arXiv preprint arXiv:2212.08568} .
\bibitem[{Farha and Gall(2019)}]{farha2019ms}
\bibinfo{author}{Farha, Y.A.}, \bibinfo{author}{Gall, J.},
  \bibinfo{year}{2019}.
\newblock \bibinfo{title}{Ms-tcn: Multi-stage temporal convolutional network
  for action segmentation}, in: \bibinfo{booktitle}{Proceedings of the IEEE/CVF
  conference on computer vision and pattern recognition}, pp.
  \bibinfo{pages}{3575--3584}.
\bibitem[{Funke et~al.(2018)Funke, Jenke, Mees, Weitz, Speidel and
  Bodenstedt}]{funke2018temporal}
\bibinfo{author}{Funke, I.}, \bibinfo{author}{Jenke, A.},
  \bibinfo{author}{Mees, S.T.}, \bibinfo{author}{Weitz, J.},
  \bibinfo{author}{Speidel, S.}, \bibinfo{author}{Bodenstedt, S.},
  \bibinfo{year}{2018}.
\newblock \bibinfo{title}{Temporal coherence-based self-supervised learning for
  laparoscopic workflow analysis}, in: \bibinfo{booktitle}{OR 2.0 Context-Aware
  Operating Theaters, Computer Assisted Robotic Endoscopy},
  \bibinfo{organization}{Springer}. pp. \bibinfo{pages}{85--93}.
\bibitem[{Gabeur et~al.(2020)Gabeur, Sun, Alahari and Schmid}]{gabeur2020multi}
\bibinfo{author}{Gabeur, V.}, \bibinfo{author}{Sun, C.},
  \bibinfo{author}{Alahari, K.}, \bibinfo{author}{Schmid, C.},
  \bibinfo{year}{2020}.
\newblock \bibinfo{title}{Multi-modal transformer for video retrieval}, in:
  \bibinfo{booktitle}{European Conference on Computer Vision},
  \bibinfo{organization}{Springer}. pp. \bibinfo{pages}{214--229}.
\bibitem[{Gao et~al.(2017)Gao, Sun, Yang and Nevatia}]{gao2017tall}
\bibinfo{author}{Gao, J.}, \bibinfo{author}{Sun, C.}, \bibinfo{author}{Yang,
  Z.}, \bibinfo{author}{Nevatia, R.}, \bibinfo{year}{2017}.
\newblock \bibinfo{title}{Tall: Temporal activity localization via language
  query}, in: \bibinfo{booktitle}{Proceedings of the IEEE international
  conference on computer vision}, pp. \bibinfo{pages}{5267--5275}.
\bibitem[{Garrow et~al.(2021)Garrow, Kowalewski, Li, Wagner, Schmidt,
  Engelhardt, Hashimoto, Kenngott, Bodenstedt, Speidel
  et~al.}]{garrow2021machine}
\bibinfo{author}{Garrow, C.R.}, \bibinfo{author}{Kowalewski, K.F.},
  \bibinfo{author}{Li, L.}, \bibinfo{author}{Wagner, M.},
  \bibinfo{author}{Schmidt, M.W.}, \bibinfo{author}{Engelhardt, S.},
  \bibinfo{author}{Hashimoto, D.A.}, \bibinfo{author}{Kenngott, H.G.},
  \bibinfo{author}{Bodenstedt, S.}, \bibinfo{author}{Speidel, S.}, et~al.,
  \bibinfo{year}{2021}.
\newblock \bibinfo{title}{Machine learning for surgical phase recognition: a
  systematic review}.
\newblock \bibinfo{journal}{Annals of surgery} \bibinfo{volume}{273},
  \bibinfo{pages}{684--693}.
\bibitem[{Girshick(2015)}]{girshick2015fast}
\bibinfo{author}{Girshick, R.}, \bibinfo{year}{2015}.
\newblock \bibinfo{title}{Fast r-cnn}, in: \bibinfo{booktitle}{Proceedings of
  the IEEE international conference on computer vision}, pp.
  \bibinfo{pages}{1440--1448}.
\bibitem[{Grauman et~al.(2022)Grauman, Westbury, Byrne, Chavis, Furnari,
  Girdhar, Hamburger, Jiang, Liu, Liu et~al.}]{grauman2022ego4d}
\bibinfo{author}{Grauman, K.}, \bibinfo{author}{Westbury, A.},
  \bibinfo{author}{Byrne, E.}, \bibinfo{author}{Chavis, Z.},
  \bibinfo{author}{Furnari, A.}, \bibinfo{author}{Girdhar, R.},
  \bibinfo{author}{Hamburger, J.}, \bibinfo{author}{Jiang, H.},
  \bibinfo{author}{Liu, M.}, \bibinfo{author}{Liu, X.}, et~al.,
  \bibinfo{year}{2022}.
\newblock \bibinfo{title}{Ego4d: Around the world in 3,000 hours of egocentric
  video}, in: \bibinfo{booktitle}{Proceedings of the IEEE/CVF Conference on
  Computer Vision and Pattern Recognition}, pp. \bibinfo{pages}{18995--19012}.
\bibitem[{Guo et~al.(2019)Guo, Wang and Wang}]{guo2019deep}
\bibinfo{author}{Guo, W.}, \bibinfo{author}{Wang, J.}, \bibinfo{author}{Wang,
  S.}, \bibinfo{year}{2019}.
\newblock \bibinfo{title}{Deep multimodal representation learning: A survey}.
\newblock \bibinfo{journal}{Ieee Access} \bibinfo{volume}{7},
  \bibinfo{pages}{63373--63394}.
\bibitem[{Habibian et~al.(2016)Habibian, Mensink and
  Snoek}]{habibian2016video2vec}
\bibinfo{author}{Habibian, A.}, \bibinfo{author}{Mensink, T.},
  \bibinfo{author}{Snoek, C.G.}, \bibinfo{year}{2016}.
\newblock \bibinfo{title}{Video2vec embeddings recognize events when examples
  are scarce}.
\newblock \bibinfo{journal}{IEEE transactions on pattern analysis and machine
  intelligence} \bibinfo{volume}{39}, \bibinfo{pages}{2089--2103}.
\bibitem[{He et~al.(2020)He, Fan, Wu, Xie and Girshick}]{he2020momentum}
\bibinfo{author}{He, K.}, \bibinfo{author}{Fan, H.}, \bibinfo{author}{Wu, Y.},
  \bibinfo{author}{Xie, S.}, \bibinfo{author}{Girshick, R.},
  \bibinfo{year}{2020}.
\newblock \bibinfo{title}{Momentum contrast for unsupervised visual
  representation learning}, in: \bibinfo{booktitle}{Proceedings of the IEEE/CVF
  conference on computer vision and pattern recognition}, pp.
  \bibinfo{pages}{9729--9738}.
\bibitem[{He et~al.(2016)He, Zhang, Ren and Sun}]{he2016deep}
\bibinfo{author}{He, K.}, \bibinfo{author}{Zhang, X.}, \bibinfo{author}{Ren,
  S.}, \bibinfo{author}{Sun, J.}, \bibinfo{year}{2016}.
\newblock \bibinfo{title}{Deep residual learning for image recognition}, in:
  \bibinfo{booktitle}{Proceedings of the IEEE conference on computer vision and
  pattern recognition}, pp. \bibinfo{pages}{770--778}.
\bibitem[{Huang et~al.(2019)Huang, Altosaar and
  Ranganath}]{huang2019clinicalbert}
\bibinfo{author}{Huang, K.}, \bibinfo{author}{Altosaar, J.},
  \bibinfo{author}{Ranganath, R.}, \bibinfo{year}{2019}.
\newblock \bibinfo{title}{Clinicalbert: Modeling clinical notes and predicting
  hospital readmission}.
\newblock \bibinfo{journal}{arXiv preprint arXiv:1904.05342} .
\bibitem[{Jia et~al.(2021)Jia, Yang, Xia, Chen, Parekh, Pham, Le, Sung, Li and
  Duerig}]{jia2021scaling}
\bibinfo{author}{Jia, C.}, \bibinfo{author}{Yang, Y.}, \bibinfo{author}{Xia,
  Y.}, \bibinfo{author}{Chen, Y.T.}, \bibinfo{author}{Parekh, Z.},
  \bibinfo{author}{Pham, H.}, \bibinfo{author}{Le, Q.}, \bibinfo{author}{Sung,
  Y.H.}, \bibinfo{author}{Li, Z.}, \bibinfo{author}{Duerig, T.},
  \bibinfo{year}{2021}.
\newblock \bibinfo{title}{Scaling up visual and vision-language representation
  learning with noisy text supervision}, in: \bibinfo{booktitle}{International
  Conference on Machine Learning}, \bibinfo{organization}{PMLR}. pp.
  \bibinfo{pages}{4904--4916}.
\bibitem[{Jiang et~al.(2017)Jiang, Wu, Wang, Xue and
  Chang}]{jiang2017exploiting}
\bibinfo{author}{Jiang, Y.G.}, \bibinfo{author}{Wu, Z.}, \bibinfo{author}{Wang,
  J.}, \bibinfo{author}{Xue, X.}, \bibinfo{author}{Chang, S.F.},
  \bibinfo{year}{2017}.
\newblock \bibinfo{title}{Exploiting feature and class relationships in video
  categorization with regularized deep neural networks}.
\newblock \bibinfo{journal}{IEEE transactions on pattern analysis and machine
  intelligence} \bibinfo{volume}{40}, \bibinfo{pages}{352--364}.
\bibitem[{Jin et~al.(2017)Jin, Dou, Chen, Yu, Qin, Fu and Heng}]{jin2017sv}
\bibinfo{author}{Jin, Y.}, \bibinfo{author}{Dou, Q.}, \bibinfo{author}{Chen,
  H.}, \bibinfo{author}{Yu, L.}, \bibinfo{author}{Qin, J.},
  \bibinfo{author}{Fu, C.W.}, \bibinfo{author}{Heng, P.A.},
  \bibinfo{year}{2017}.
\newblock \bibinfo{title}{Sv-rcnet: workflow recognition from surgical videos
  using recurrent convolutional network}.
\newblock \bibinfo{journal}{IEEE transactions on medical imaging}
  \bibinfo{volume}{37}, \bibinfo{pages}{1114--1126}.
\bibitem[{Jin et~al.(2021)Jin, Long, Chen, Zhao, Dou and
  Heng}]{jin2021temporal}
\bibinfo{author}{Jin, Y.}, \bibinfo{author}{Long, Y.}, \bibinfo{author}{Chen,
  C.}, \bibinfo{author}{Zhao, Z.}, \bibinfo{author}{Dou, Q.},
  \bibinfo{author}{Heng, P.A.}, \bibinfo{year}{2021}.
\newblock \bibinfo{title}{Temporal memory relation network for workflow
  recognition from surgical video}.
\newblock \bibinfo{journal}{IEEE Transactions on Medical Imaging}
  \bibinfo{volume}{40}, \bibinfo{pages}{1911--1923}.
\bibitem[{Jing et~al.(2018)Jing, Yang, Liu and Tian}]{jing2018self}
\bibinfo{author}{Jing, L.}, \bibinfo{author}{Yang, X.}, \bibinfo{author}{Liu,
  J.}, \bibinfo{author}{Tian, Y.}, \bibinfo{year}{2018}.
\newblock \bibinfo{title}{Self-supervised spatiotemporal feature learning via
  video rotation prediction}.
\newblock \bibinfo{journal}{arXiv preprint arXiv:1811.11387} .
\bibitem[{Kan et~al.(2023)Kan, Wang, Lu, Zhen, Guan and
  Zheng}]{kan2023knowledge}
\bibinfo{author}{Kan, B.}, \bibinfo{author}{Wang, T.}, \bibinfo{author}{Lu,
  W.}, \bibinfo{author}{Zhen, X.}, \bibinfo{author}{Guan, W.},
  \bibinfo{author}{Zheng, F.}, \bibinfo{year}{2023}.
\newblock \bibinfo{title}{Knowledge-aware prompt tuning for generalizable
  vision-language models}, in: \bibinfo{booktitle}{Proceedings of the IEEE/CVF
  International Conference on Computer Vision}, pp.
  \bibinfo{pages}{15670--15680}.
\bibitem[{Krishna et~al.(2017)Krishna, Zhu, Groth, Johnson, Hata, Kravitz,
  Chen, Kalantidis, Li, Shamma et~al.}]{krishna2017visual}
\bibinfo{author}{Krishna, R.}, \bibinfo{author}{Zhu, Y.},
  \bibinfo{author}{Groth, O.}, \bibinfo{author}{Johnson, J.},
  \bibinfo{author}{Hata, K.}, \bibinfo{author}{Kravitz, J.},
  \bibinfo{author}{Chen, S.}, \bibinfo{author}{Kalantidis, Y.},
  \bibinfo{author}{Li, L.J.}, \bibinfo{author}{Shamma, D.A.}, et~al.,
  \bibinfo{year}{2017}.
\newblock \bibinfo{title}{Visual genome: Connecting language and vision using
  crowdsourced dense image annotations}.
\newblock \bibinfo{journal}{International journal of computer vision}
  \bibinfo{volume}{123}, \bibinfo{pages}{32--73}.
\bibitem[{Lewis et~al.(2019)Lewis, Liu, Goyal, Ghazvininejad, Mohamed, Levy,
  Stoyanov and Zettlemoyer}]{lewis2019bart}
\bibinfo{author}{Lewis, M.}, \bibinfo{author}{Liu, Y.}, \bibinfo{author}{Goyal,
  N.}, \bibinfo{author}{Ghazvininejad, M.}, \bibinfo{author}{Mohamed, A.},
  \bibinfo{author}{Levy, O.}, \bibinfo{author}{Stoyanov, V.},
  \bibinfo{author}{Zettlemoyer, L.}, \bibinfo{year}{2019}.
\newblock \bibinfo{title}{Bart: Denoising sequence-to-sequence pre-training for
  natural language generation, translation, and comprehension}.
\newblock \bibinfo{journal}{arXiv preprint arXiv:1910.13461} .
\bibitem[{Li et~al.(2022)Li, Li, Xiong and Hoi}]{li2022blip}
\bibinfo{author}{Li, J.}, \bibinfo{author}{Li, D.}, \bibinfo{author}{Xiong,
  C.}, \bibinfo{author}{Hoi, S.}, \bibinfo{year}{2022}.
\newblock \bibinfo{title}{Blip: Bootstrapping language-image pre-training for
  unified vision-language understanding and generation}, in:
  \bibinfo{booktitle}{International conference on machine learning},
  \bibinfo{organization}{PMLR}. pp. \bibinfo{pages}{12888--12900}.
\bibitem[{Li et~al.(2019)Li, Yatskar, Yin, Hsieh and Chang}]{li2019visualbert}
\bibinfo{author}{Li, L.H.}, \bibinfo{author}{Yatskar, M.},
  \bibinfo{author}{Yin, D.}, \bibinfo{author}{Hsieh, C.J.},
  \bibinfo{author}{Chang, K.W.}, \bibinfo{year}{2019}.
\newblock \bibinfo{title}{Visualbert: A simple and performant baseline for
  vision and language}.
\newblock \bibinfo{journal}{arXiv preprint arXiv:1908.03557} .
\bibitem[{Li et~al.(2021)Li, Liang, Zhao, Cui, Ouyang, Shao, Yu and
  Yan}]{li2021supervision}
\bibinfo{author}{Li, Y.}, \bibinfo{author}{Liang, F.}, \bibinfo{author}{Zhao,
  L.}, \bibinfo{author}{Cui, Y.}, \bibinfo{author}{Ouyang, W.},
  \bibinfo{author}{Shao, J.}, \bibinfo{author}{Yu, F.}, \bibinfo{author}{Yan,
  J.}, \bibinfo{year}{2021}.
\newblock \bibinfo{title}{Supervision exists everywhere: A data efficient
  contrastive language-image pre-training paradigm}.
\newblock \bibinfo{journal}{arXiv preprint arXiv:2110.05208} .
\bibitem[{Liang et~al.(2017)Liang, Hu, Zhang, Gan and
  Xing}]{liang2017recurrent}
\bibinfo{author}{Liang, X.}, \bibinfo{author}{Hu, Z.}, \bibinfo{author}{Zhang,
  H.}, \bibinfo{author}{Gan, C.}, \bibinfo{author}{Xing, E.P.},
  \bibinfo{year}{2017}.
\newblock \bibinfo{title}{Recurrent topic-transition gan for visual paragraph
  generation}, in: \bibinfo{booktitle}{Proceedings of the IEEE international
  conference on computer vision}, pp. \bibinfo{pages}{3362--3371}.
\bibitem[{Lin(2004)}]{lin2004rouge}
\bibinfo{author}{Lin, C.Y.}, \bibinfo{year}{2004}.
\newblock \bibinfo{title}{Rouge: A package for automatic evaluation of
  summaries}, in: \bibinfo{booktitle}{Text summarization branches out}, pp.
  \bibinfo{pages}{74--81}.
\bibitem[{Liu et~al.(2016)Liu, Feng and Zhou}]{liu2016multimodal}
\bibinfo{author}{Liu, Y.}, \bibinfo{author}{Feng, X.}, \bibinfo{author}{Zhou,
  Z.}, \bibinfo{year}{2016}.
\newblock \bibinfo{title}{Multimodal video classification with stacked
  contractive autoencoders}.
\newblock \bibinfo{journal}{Signal Processing} \bibinfo{volume}{120},
  \bibinfo{pages}{761--766}.
\bibitem[{Liu et~al.(2019)Liu, Ott, Goyal, Du, Joshi, Chen, Levy, Lewis,
  Zettlemoyer and Stoyanov}]{liu2019roberta}
\bibinfo{author}{Liu, Y.}, \bibinfo{author}{Ott, M.}, \bibinfo{author}{Goyal,
  N.}, \bibinfo{author}{Du, J.}, \bibinfo{author}{Joshi, M.},
  \bibinfo{author}{Chen, D.}, \bibinfo{author}{Levy, O.},
  \bibinfo{author}{Lewis, M.}, \bibinfo{author}{Zettlemoyer, L.},
  \bibinfo{author}{Stoyanov, V.}, \bibinfo{year}{2019}.
\newblock \bibinfo{title}{Roberta: A robustly optimized bert pretraining
  approach}.
\newblock \bibinfo{journal}{arXiv preprint arXiv:1907.11692} .
\bibitem[{Lu et~al.(2019)Lu, Batra, Parikh and Lee}]{lu2019vilbert}
\bibinfo{author}{Lu, J.}, \bibinfo{author}{Batra, D.}, \bibinfo{author}{Parikh,
  D.}, \bibinfo{author}{Lee, S.}, \bibinfo{year}{2019}.
\newblock \bibinfo{title}{Vilbert: Pretraining task-agnostic visiolinguistic
  representations for vision-and-language tasks}.
\newblock \bibinfo{journal}{Advances in neural information processing systems}
  \bibinfo{volume}{32}.
\bibitem[{Luo et~al.(2020)Luo, Ji, Shi, Huang, Duan, Li, Li, Bharti and
  Zhou}]{luo2020univl}
\bibinfo{author}{Luo, H.}, \bibinfo{author}{Ji, L.}, \bibinfo{author}{Shi, B.},
  \bibinfo{author}{Huang, H.}, \bibinfo{author}{Duan, N.}, \bibinfo{author}{Li,
  T.}, \bibinfo{author}{Li, J.}, \bibinfo{author}{Bharti, T.},
  \bibinfo{author}{Zhou, M.}, \bibinfo{year}{2020}.
\newblock \bibinfo{title}{Univl: A unified video and language pre-training
  model for multimodal understanding and generation}.
\newblock \bibinfo{journal}{arXiv preprint arXiv:2002.06353} .
\bibitem[{Madani et~al.(2020)Madani, Namazi, Altieri, Hashimoto, Rivera,
  Pucher, Navarrete-Welton, Sankaranarayanan, Brunt, Okrainec
  et~al.}]{madani2020artificial}
\bibinfo{author}{Madani, A.}, \bibinfo{author}{Namazi, B.},
  \bibinfo{author}{Altieri, M.S.}, \bibinfo{author}{Hashimoto, D.A.},
  \bibinfo{author}{Rivera, A.M.}, \bibinfo{author}{Pucher, P.H.},
  \bibinfo{author}{Navarrete-Welton, A.}, \bibinfo{author}{Sankaranarayanan,
  G.}, \bibinfo{author}{Brunt, L.M.}, \bibinfo{author}{Okrainec, A.}, et~al.,
  \bibinfo{year}{2020}.
\newblock \bibinfo{title}{Artificial intelligence for intraoperative guidance:
  using semantic segmentation to identify surgical anatomy during laparoscopic
  cholecystectomy}.
\newblock \bibinfo{journal}{Annals of surgery} .
\bibitem[{Mahajan et~al.(2018)Mahajan, Girshick, Ramanathan, He, Paluri, Li,
  Bharambe and Van Der~Maaten}]{mahajan2018exploring}
\bibinfo{author}{Mahajan, D.}, \bibinfo{author}{Girshick, R.},
  \bibinfo{author}{Ramanathan, V.}, \bibinfo{author}{He, K.},
  \bibinfo{author}{Paluri, M.}, \bibinfo{author}{Li, Y.},
  \bibinfo{author}{Bharambe, A.}, \bibinfo{author}{Van Der~Maaten, L.},
  \bibinfo{year}{2018}.
\newblock \bibinfo{title}{Exploring the limits of weakly supervised
  pretraining}, in: \bibinfo{booktitle}{Proceedings of the European conference
  on computer vision (ECCV)}, pp. \bibinfo{pages}{181--196}.
\bibitem[{Maier-Hein et~al.(2022)Maier-Hein, Eisenmann, Sarikaya, M{\"a}rz,
  Collins, Malpani, Fallert, Feussner, Giannarou, Mascagni
  et~al.}]{maier2022surgical}
\bibinfo{author}{Maier-Hein, L.}, \bibinfo{author}{Eisenmann, M.},
  \bibinfo{author}{Sarikaya, D.}, \bibinfo{author}{M{\"a}rz, K.},
  \bibinfo{author}{Collins, T.}, \bibinfo{author}{Malpani, A.},
  \bibinfo{author}{Fallert, J.}, \bibinfo{author}{Feussner, H.},
  \bibinfo{author}{Giannarou, S.}, \bibinfo{author}{Mascagni, P.}, et~al.,
  \bibinfo{year}{2022}.
\newblock \bibinfo{title}{Surgical data science--from concepts toward clinical
  translation}.
\newblock \bibinfo{journal}{Medical image analysis} \bibinfo{volume}{76},
  \bibinfo{pages}{102306}.
\bibitem[{Maier-Hein et~al.(2017)Maier-Hein, Vedula, Speidel, Navab, Kikinis,
  Park, Eisenmann, Feussner, Forestier, Giannarou et~al.}]{maier2017surgical}
\bibinfo{author}{Maier-Hein, L.}, \bibinfo{author}{Vedula, S.S.},
  \bibinfo{author}{Speidel, S.}, \bibinfo{author}{Navab, N.},
  \bibinfo{author}{Kikinis, R.}, \bibinfo{author}{Park, A.},
  \bibinfo{author}{Eisenmann, M.}, \bibinfo{author}{Feussner, H.},
  \bibinfo{author}{Forestier, G.}, \bibinfo{author}{Giannarou, S.}, et~al.,
  \bibinfo{year}{2017}.
\newblock \bibinfo{title}{Surgical data science for next-generation
  interventions}.
\newblock \bibinfo{journal}{Nature Biomedical Engineering} \bibinfo{volume}{1},
  \bibinfo{pages}{691--696}.
\bibitem[{Mascagni et~al.(2022)Mascagni, Alapatt, Sestini, Altieri, Madani,
  Watanabe, Alseidi, Redan, Alfieri, Costamagna et~al.}]{mascagni2022computer}
\bibinfo{author}{Mascagni, P.}, \bibinfo{author}{Alapatt, D.},
  \bibinfo{author}{Sestini, L.}, \bibinfo{author}{Altieri, M.S.},
  \bibinfo{author}{Madani, A.}, \bibinfo{author}{Watanabe, Y.},
  \bibinfo{author}{Alseidi, A.}, \bibinfo{author}{Redan, J.A.},
  \bibinfo{author}{Alfieri, S.}, \bibinfo{author}{Costamagna, G.}, et~al.,
  \bibinfo{year}{2022}.
\newblock \bibinfo{title}{Computer vision in surgery: from potential to
  clinical value}.
\newblock \bibinfo{journal}{npj Digital Medicine} \bibinfo{volume}{5},
  \bibinfo{pages}{163}.
\bibitem[{Mehrish et~al.(2023)Mehrish, Majumder, Bhardwaj and
  Poria}]{mehrish2023review}
\bibinfo{author}{Mehrish, A.}, \bibinfo{author}{Majumder, N.},
  \bibinfo{author}{Bhardwaj, R.}, \bibinfo{author}{Poria, S.},
  \bibinfo{year}{2023}.
\newblock \bibinfo{title}{A review of deep learning techniques for speech
  processing}.
\newblock \bibinfo{journal}{arXiv preprint arXiv:2305.00359} .
\bibitem[{Miech et~al.(2020)Miech, Alayrac, Smaira, Laptev, Sivic and
  Zisserman}]{miech2020end}
\bibinfo{author}{Miech, A.}, \bibinfo{author}{Alayrac, J.B.},
  \bibinfo{author}{Smaira, L.}, \bibinfo{author}{Laptev, I.},
  \bibinfo{author}{Sivic, J.}, \bibinfo{author}{Zisserman, A.},
  \bibinfo{year}{2020}.
\newblock \bibinfo{title}{End-to-end learning of visual representations from
  uncurated instructional videos}, in: \bibinfo{booktitle}{Proceedings of the
  IEEE/CVF Conference on Computer Vision and Pattern Recognition}, pp.
  \bibinfo{pages}{9879--9889}.
\bibitem[{Miech et~al.(2019)Miech, Zhukov, Alayrac, Tapaswi, Laptev and
  Sivic}]{miech2019howto100m}
\bibinfo{author}{Miech, A.}, \bibinfo{author}{Zhukov, D.},
  \bibinfo{author}{Alayrac, J.B.}, \bibinfo{author}{Tapaswi, M.},
  \bibinfo{author}{Laptev, I.}, \bibinfo{author}{Sivic, J.},
  \bibinfo{year}{2019}.
\newblock \bibinfo{title}{Howto100m: Learning a text-video embedding by
  watching hundred million narrated video clips}, in:
  \bibinfo{booktitle}{Proceedings of the IEEE/CVF International Conference on
  Computer Vision}, pp. \bibinfo{pages}{2630--2640}.
\bibitem[{Mikolov et~al.(2013)Mikolov, Chen, Corrado and
  Dean}]{mikolov2013efficient}
\bibinfo{author}{Mikolov, T.}, \bibinfo{author}{Chen, K.},
  \bibinfo{author}{Corrado, G.}, \bibinfo{author}{Dean, J.},
  \bibinfo{year}{2013}.
\newblock \bibinfo{title}{Efficient estimation of word representations in
  vector space}.
\newblock \bibinfo{journal}{arXiv preprint arXiv:1301.3781} .
\bibitem[{Narayanan et~al.(2018)Narayanan, Misra, Sim, Pundak, Tripathi,
  Elfeky, Haghani, Strohman and Bacchiani}]{narayanan2018toward}
\bibinfo{author}{Narayanan, A.}, \bibinfo{author}{Misra, A.},
  \bibinfo{author}{Sim, K.C.}, \bibinfo{author}{Pundak, G.},
  \bibinfo{author}{Tripathi, A.}, \bibinfo{author}{Elfeky, M.},
  \bibinfo{author}{Haghani, P.}, \bibinfo{author}{Strohman, T.},
  \bibinfo{author}{Bacchiani, M.}, \bibinfo{year}{2018}.
\newblock \bibinfo{title}{Toward domain-invariant speech recognition via large
  scale training}, in: \bibinfo{booktitle}{2018 IEEE Spoken Language Technology
  Workshop (SLT)}, \bibinfo{organization}{IEEE}. pp. \bibinfo{pages}{441--447}.
\bibitem[{Nukrai et~al.(2022)Nukrai, Mokady and Globerson}]{nukrai2022text}
\bibinfo{author}{Nukrai, D.}, \bibinfo{author}{Mokady, R.},
  \bibinfo{author}{Globerson, A.}, \bibinfo{year}{2022}.
\newblock \bibinfo{title}{Text-only training for image captioning using
  noise-injected clip}.
\newblock \bibinfo{journal}{arXiv preprint arXiv:2211.00575} .
\bibitem[{Nwoye et~al.(2022)Nwoye, Yu, Gonzalez, Seeliger, Mascagni, Mutter,
  Marescaux and Padoy}]{nwoye2021rendezvous}
\bibinfo{author}{Nwoye, C.I.}, \bibinfo{author}{Yu, T.},
  \bibinfo{author}{Gonzalez, C.}, \bibinfo{author}{Seeliger, B.},
  \bibinfo{author}{Mascagni, P.}, \bibinfo{author}{Mutter, D.},
  \bibinfo{author}{Marescaux, J.}, \bibinfo{author}{Padoy, N.},
  \bibinfo{year}{2022}.
\newblock \bibinfo{title}{Rendezvous: Attention mechanisms for the recognition
  of surgical action triplets in endoscopic videos}.
\newblock \bibinfo{journal}{Medical Image Analysis} \bibinfo{volume}{78},
  \bibinfo{pages}{102433}.
\bibitem[{Oord et~al.(2018)Oord, Li and Vinyals}]{oord2018representation}
\bibinfo{author}{Oord, A.v.d.}, \bibinfo{author}{Li, Y.},
  \bibinfo{author}{Vinyals, O.}, \bibinfo{year}{2018}.
\newblock \bibinfo{title}{Representation learning with contrastive predictive
  coding}.
\newblock \bibinfo{journal}{arXiv preprint arXiv:1807.03748} .
\bibitem[{Owens and Efros(2018)}]{owens2018audio}
\bibinfo{author}{Owens, A.}, \bibinfo{author}{Efros, A.A.},
  \bibinfo{year}{2018}.
\newblock \bibinfo{title}{Audio-visual scene analysis with self-supervised
  multisensory features}, in: \bibinfo{booktitle}{Proceedings of the European
  conference on computer vision (ECCV)}, pp. \bibinfo{pages}{631--648}.
\bibitem[{Padoy et~al.(2012)Padoy, Blum, Ahmadi, Feussner, Berger and
  Navab}]{padoy2012statistical}
\bibinfo{author}{Padoy, N.}, \bibinfo{author}{Blum, T.},
  \bibinfo{author}{Ahmadi, S.A.}, \bibinfo{author}{Feussner, H.},
  \bibinfo{author}{Berger, M.O.}, \bibinfo{author}{Navab, N.},
  \bibinfo{year}{2012}.
\newblock \bibinfo{title}{Statistical modeling and recognition of surgical
  workflow}.
\newblock \bibinfo{journal}{Medical image analysis} \bibinfo{volume}{16},
  \bibinfo{pages}{632--641}.
\bibitem[{Papineni et~al.(2002)Papineni, Roukos, Ward and
  Zhu}]{papineni2002bleu}
\bibinfo{author}{Papineni, K.}, \bibinfo{author}{Roukos, S.},
  \bibinfo{author}{Ward, T.}, \bibinfo{author}{Zhu, W.J.},
  \bibinfo{year}{2002}.
\newblock \bibinfo{title}{Bleu: a method for automatic evaluation of machine
  translation}, in: \bibinfo{booktitle}{Proceedings of the 40th annual meeting
  of the Association for Computational Linguistics}, pp.
  \bibinfo{pages}{311--318}.
\bibitem[{Pfeiffer et~al.(2019)Pfeiffer, Funke, Robu, Bodenstedt, Strenger,
  Engelhardt, Ro{\ss}, Clarkson, Gurusamy, Davidson
  et~al.}]{pfeiffer2019generating}
\bibinfo{author}{Pfeiffer, M.}, \bibinfo{author}{Funke, I.},
  \bibinfo{author}{Robu, M.R.}, \bibinfo{author}{Bodenstedt, S.},
  \bibinfo{author}{Strenger, L.}, \bibinfo{author}{Engelhardt, S.},
  \bibinfo{author}{Ro{\ss}, T.}, \bibinfo{author}{Clarkson, M.J.},
  \bibinfo{author}{Gurusamy, K.}, \bibinfo{author}{Davidson, B.R.}, et~al.,
  \bibinfo{year}{2019}.
\newblock \bibinfo{title}{Generating large labeled data sets for laparoscopic
  image processing tasks using unpaired image-to-image translation}, in:
  \bibinfo{booktitle}{Medical Image Computing and Computer Assisted
  Intervention--MICCAI 2019: 22nd International Conference, Shenzhen, China,
  October 13--17, 2019, Proceedings, Part V 22},
  \bibinfo{organization}{Springer}. pp. \bibinfo{pages}{119--127}.
\bibitem[{Poria et~al.(2016)Poria, Cambria, Howard, Huang and
  Hussain}]{poria2016fusing}
\bibinfo{author}{Poria, S.}, \bibinfo{author}{Cambria, E.},
  \bibinfo{author}{Howard, N.}, \bibinfo{author}{Huang, G.B.},
  \bibinfo{author}{Hussain, A.}, \bibinfo{year}{2016}.
\newblock \bibinfo{title}{Fusing audio, visual and textual clues for sentiment
  analysis from multimodal content}.
\newblock \bibinfo{journal}{Neurocomputing} \bibinfo{volume}{174},
  \bibinfo{pages}{50--59}.
\bibitem[{Radford et~al.(2021)Radford, Kim, Hallacy, Ramesh, Goh, Agarwal,
  Sastry, Askell, Mishkin, Clark et~al.}]{radford2021learning}
\bibinfo{author}{Radford, A.}, \bibinfo{author}{Kim, J.W.},
  \bibinfo{author}{Hallacy, C.}, \bibinfo{author}{Ramesh, A.},
  \bibinfo{author}{Goh, G.}, \bibinfo{author}{Agarwal, S.},
  \bibinfo{author}{Sastry, G.}, \bibinfo{author}{Askell, A.},
  \bibinfo{author}{Mishkin, P.}, \bibinfo{author}{Clark, J.}, et~al.,
  \bibinfo{year}{2021}.
\newblock \bibinfo{title}{Learning transferable visual models from natural
  language supervision}, in: \bibinfo{booktitle}{International Conference on
  Machine Learning}, \bibinfo{organization}{PMLR}. pp.
  \bibinfo{pages}{8748--8763}.
\bibitem[{Radford et~al.(2023)Radford, Kim, Xu, Brockman, McLeavey and
  Sutskever}]{radford2022robust}
\bibinfo{author}{Radford, A.}, \bibinfo{author}{Kim, J.W.},
  \bibinfo{author}{Xu, T.}, \bibinfo{author}{Brockman, G.},
  \bibinfo{author}{McLeavey, C.}, \bibinfo{author}{Sutskever, I.},
  \bibinfo{year}{2023}.
\newblock \bibinfo{title}{Robust speech recognition via large-scale weak
  supervision}, in: \bibinfo{booktitle}{International Conference on Machine
  Learning}, \bibinfo{organization}{PMLR}. pp. \bibinfo{pages}{28492--28518}.
\bibitem[{Raffel et~al.(2020)Raffel, Shazeer, Roberts, Lee, Narang, Matena,
  Zhou, Li and Liu}]{raffel2020exploring}
\bibinfo{author}{Raffel, C.}, \bibinfo{author}{Shazeer, N.},
  \bibinfo{author}{Roberts, A.}, \bibinfo{author}{Lee, K.},
  \bibinfo{author}{Narang, S.}, \bibinfo{author}{Matena, M.},
  \bibinfo{author}{Zhou, Y.}, \bibinfo{author}{Li, W.}, \bibinfo{author}{Liu,
  P.J.}, \bibinfo{year}{2020}.
\newblock \bibinfo{title}{Exploring the limits of transfer learning with a
  unified text-to-text transformer}.
\newblock \bibinfo{journal}{The Journal of Machine Learning Research}
  \bibinfo{volume}{21}, \bibinfo{pages}{5485--5551}.
\bibitem[{Ramesh et~al.(2023)Ramesh, Srivastav, Alapatt, Yu, Murali, Sestini,
  Nwoye, Hamoud, Sharma, Fleurentin et~al.}]{ramesh2022dissecting}
\bibinfo{author}{Ramesh, S.}, \bibinfo{author}{Srivastav, V.},
  \bibinfo{author}{Alapatt, D.}, \bibinfo{author}{Yu, T.},
  \bibinfo{author}{Murali, A.}, \bibinfo{author}{Sestini, L.},
  \bibinfo{author}{Nwoye, C.I.}, \bibinfo{author}{Hamoud, I.},
  \bibinfo{author}{Sharma, S.}, \bibinfo{author}{Fleurentin, A.}, et~al.,
  \bibinfo{year}{2023}.
\newblock \bibinfo{title}{Dissecting self-supervised learning methods for
  surgical computer vision}.
\newblock \bibinfo{journal}{Medical Image Analysis} \bibinfo{volume}{88},
  \bibinfo{pages}{102844}.
\bibitem[{Reed et~al.(2016)Reed, Akata, Yan, Logeswaran, Schiele and
  Lee}]{reed2016generative}
\bibinfo{author}{Reed, S.}, \bibinfo{author}{Akata, Z.}, \bibinfo{author}{Yan,
  X.}, \bibinfo{author}{Logeswaran, L.}, \bibinfo{author}{Schiele, B.},
  \bibinfo{author}{Lee, H.}, \bibinfo{year}{2016}.
\newblock \bibinfo{title}{Generative adversarial text to image synthesis}, in:
  \bibinfo{booktitle}{International conference on machine learning},
  \bibinfo{organization}{PMLR}. pp. \bibinfo{pages}{1060--1069}.
\bibitem[{Rivoir et~al.(2019)Rivoir, Bodenstedt, von Bechtolsheim, Distler,
  Weitz and Speidel}]{rivoir2019unsupervised}
\bibinfo{author}{Rivoir, D.}, \bibinfo{author}{Bodenstedt, S.},
  \bibinfo{author}{von Bechtolsheim, F.}, \bibinfo{author}{Distler, M.},
  \bibinfo{author}{Weitz, J.}, \bibinfo{author}{Speidel, S.},
  \bibinfo{year}{2019}.
\newblock \bibinfo{title}{Unsupervised temporal video segmentation as an
  auxiliary task for predicting the remaining surgery duration}, in:
  \bibinfo{booktitle}{International Workshop on OR 2.0 Context-Aware Operating
  Theaters}, \bibinfo{organization}{Springer}. pp. \bibinfo{pages}{29--37}.
\bibitem[{Rivoir et~al.(2022)Rivoir, Funke and Speidel}]{rivoir2022pitfalls}
\bibinfo{author}{Rivoir, D.}, \bibinfo{author}{Funke, I.},
  \bibinfo{author}{Speidel, S.}, \bibinfo{year}{2022}.
\newblock \bibinfo{title}{On the pitfalls of batch normalization for end-to-end
  video learning: A study on surgical workflow analysis}.
\newblock \bibinfo{journal}{arXiv preprint arXiv:2203.07976} .
\bibitem[{Rivoir et~al.(2021)Rivoir, Pfeiffer, Docea, Kolbinger, Riediger,
  Weitz and Speidel}]{rivoir2021long}
\bibinfo{author}{Rivoir, D.}, \bibinfo{author}{Pfeiffer, M.},
  \bibinfo{author}{Docea, R.}, \bibinfo{author}{Kolbinger, F.},
  \bibinfo{author}{Riediger, C.}, \bibinfo{author}{Weitz, J.},
  \bibinfo{author}{Speidel, S.}, \bibinfo{year}{2021}.
\newblock \bibinfo{title}{Long-term temporally consistent unpaired video
  translation from simulated surgical 3d data}, in:
  \bibinfo{booktitle}{Proceedings of the IEEE/CVF International Conference on
  Computer Vision}, pp. \bibinfo{pages}{3343--3353}.
\bibitem[{Rojas-Mu{\~n}oz et~al.(2020)Rojas-Mu{\~n}oz, Couperus and
  Wachs}]{rojas2020daisi}
\bibinfo{author}{Rojas-Mu{\~n}oz, E.}, \bibinfo{author}{Couperus, K.},
  \bibinfo{author}{Wachs, J.}, \bibinfo{year}{2020}.
\newblock \bibinfo{title}{Daisi: Database for ai surgical instruction}.
\newblock \bibinfo{journal}{arXiv preprint arXiv:2004.02809} .
\bibitem[{Rombach et~al.(2022)Rombach, Blattmann, Lorenz, Esser and
  Ommer}]{rombach2022high}
\bibinfo{author}{Rombach, R.}, \bibinfo{author}{Blattmann, A.},
  \bibinfo{author}{Lorenz, D.}, \bibinfo{author}{Esser, P.},
  \bibinfo{author}{Ommer, B.}, \bibinfo{year}{2022}.
\newblock \bibinfo{title}{High-resolution image synthesis with latent diffusion
  models}, in: \bibinfo{booktitle}{Proceedings of the IEEE/CVF conference on
  computer vision and pattern recognition}, pp. \bibinfo{pages}{10684--10695}.
\bibitem[{Ross et~al.(2018)Ross, Zimmerer, Vemuri, Isensee, Wiesenfarth,
  Bodenstedt, Both, Kessler, Wagner, M{\"u}ller et~al.}]{ross2018exploiting}
\bibinfo{author}{Ross, T.}, \bibinfo{author}{Zimmerer, D.},
  \bibinfo{author}{Vemuri, A.}, \bibinfo{author}{Isensee, F.},
  \bibinfo{author}{Wiesenfarth, M.}, \bibinfo{author}{Bodenstedt, S.},
  \bibinfo{author}{Both, F.}, \bibinfo{author}{Kessler, P.},
  \bibinfo{author}{Wagner, M.}, \bibinfo{author}{M{\"u}ller, B.}, et~al.,
  \bibinfo{year}{2018}.
\newblock \bibinfo{title}{Exploiting the potential of unlabeled endoscopic
  video data with self-supervised learning}.
\newblock \bibinfo{journal}{International journal of computer assisted
  radiology and surgery} \bibinfo{volume}{13}, \bibinfo{pages}{925--933}.
\bibitem[{Sain et~al.(2023)Sain, Bhunia, Chowdhury, Koley, Xiang and
  Song}]{sain2023clip}
\bibinfo{author}{Sain, A.}, \bibinfo{author}{Bhunia, A.K.},
  \bibinfo{author}{Chowdhury, P.N.}, \bibinfo{author}{Koley, S.},
  \bibinfo{author}{Xiang, T.}, \bibinfo{author}{Song, Y.Z.},
  \bibinfo{year}{2023}.
\newblock \bibinfo{title}{Clip for all things zero-shot sketch-based image
  retrieval, fine-grained or not}, in: \bibinfo{booktitle}{Proceedings of the
  IEEE/CVF Conference on Computer Vision and Pattern Recognition}, pp.
  \bibinfo{pages}{2765--2775}.
\bibitem[{Sanghi et~al.(2022)Sanghi, Chu, Lambourne, Wang, Cheng, Fumero and
  Malekshan}]{sanghi2022clip}
\bibinfo{author}{Sanghi, A.}, \bibinfo{author}{Chu, H.},
  \bibinfo{author}{Lambourne, J.G.}, \bibinfo{author}{Wang, Y.},
  \bibinfo{author}{Cheng, C.Y.}, \bibinfo{author}{Fumero, M.},
  \bibinfo{author}{Malekshan, K.R.}, \bibinfo{year}{2022}.
\newblock \bibinfo{title}{Clip-forge: Towards zero-shot text-to-shape
  generation}, in: \bibinfo{booktitle}{Proceedings of the IEEE/CVF Conference
  on Computer Vision and Pattern Recognition}, pp.
  \bibinfo{pages}{18603--18613}.
\bibitem[{Seenivasan et~al.(2022)Seenivasan, Islam, Krishna and
  Ren}]{seenivasan2022surgical}
\bibinfo{author}{Seenivasan, L.}, \bibinfo{author}{Islam, M.},
  \bibinfo{author}{Krishna, A.K.}, \bibinfo{author}{Ren, H.},
  \bibinfo{year}{2022}.
\newblock \bibinfo{title}{Surgical-vqa: Visual question answering in surgical
  scenes using transformer}, in: \bibinfo{booktitle}{International Conference
  on Medical Image Computing and Computer-Assisted Intervention},
  \bibinfo{organization}{Springer}. pp. \bibinfo{pages}{33--43}.
\bibitem[{Seide et~al.(2011)Seide, Li, Chen and Yu}]{seide2011feature}
\bibinfo{author}{Seide, F.}, \bibinfo{author}{Li, G.}, \bibinfo{author}{Chen,
  X.}, \bibinfo{author}{Yu, D.}, \bibinfo{year}{2011}.
\newblock \bibinfo{title}{Feature engineering in context-dependent deep neural
  networks for conversational speech transcription}, in:
  \bibinfo{booktitle}{2011 IEEE Workshop on Automatic Speech Recognition \&
  Understanding}, \bibinfo{organization}{IEEE}. pp. \bibinfo{pages}{24--29}.
\bibitem[{Shah et~al.(2023)Shah, Sikder, Vedula and Patel}]{shah2023glsformer}
\bibinfo{author}{Shah, N.A.}, \bibinfo{author}{Sikder, S.},
  \bibinfo{author}{Vedula, S.S.}, \bibinfo{author}{Patel, V.M.},
  \bibinfo{year}{2023}.
\newblock \bibinfo{title}{Glsformer: Gated-long, short sequence transformer for
  step recognition in surgical videos}.
\newblock \bibinfo{journal}{arXiv preprint arXiv:2307.11081} .
\bibitem[{Shen et~al.(2022)Shen, Li, Hu, Xie, Yang, Zhang, Gan, Wang, Yuan, Liu
  et~al.}]{shen2022k}
\bibinfo{author}{Shen, S.}, \bibinfo{author}{Li, C.}, \bibinfo{author}{Hu, X.},
  \bibinfo{author}{Xie, Y.}, \bibinfo{author}{Yang, J.},
  \bibinfo{author}{Zhang, P.}, \bibinfo{author}{Gan, Z.},
  \bibinfo{author}{Wang, L.}, \bibinfo{author}{Yuan, L.}, \bibinfo{author}{Liu,
  C.}, et~al., \bibinfo{year}{2022}.
\newblock \bibinfo{title}{K-lite: Learning transferable visual models with
  external knowledge}.
\newblock \bibinfo{journal}{Advances in Neural Information Processing Systems}
  \bibinfo{volume}{35}, \bibinfo{pages}{15558--15573}.
\bibitem[{Soomro et~al.(2012)Soomro, Zamir and Shah}]{soomro2012ucf101}
\bibinfo{author}{Soomro, K.}, \bibinfo{author}{Zamir, A.R.},
  \bibinfo{author}{Shah, M.}, \bibinfo{year}{2012}.
\newblock \bibinfo{title}{Ucf101: A dataset of 101 human actions classes from
  videos in the wild}.
\newblock \bibinfo{journal}{arXiv preprint arXiv:1212.0402} .
\bibitem[{Tang et~al.(2019)Tang, Ding, Rao, Zheng, Zhang, Zhao, Lu and
  Zhou}]{tang2019coin}
\bibinfo{author}{Tang, Y.}, \bibinfo{author}{Ding, D.}, \bibinfo{author}{Rao,
  Y.}, \bibinfo{author}{Zheng, Y.}, \bibinfo{author}{Zhang, D.},
  \bibinfo{author}{Zhao, L.}, \bibinfo{author}{Lu, J.}, \bibinfo{author}{Zhou,
  J.}, \bibinfo{year}{2019}.
\newblock \bibinfo{title}{Coin: A large-scale dataset for comprehensive
  instructional video analysis}, in: \bibinfo{booktitle}{Proceedings of the
  IEEE/CVF Conference on Computer Vision and Pattern Recognition}, pp.
  \bibinfo{pages}{1207--1216}.
\bibitem[{Touvron et~al.(2023)Touvron, Lavril, Izacard, Martinet, Lachaux,
  Lacroix, Rozi{\`e}re, Goyal, Hambro, Azhar et~al.}]{touvron2023llama}
\bibinfo{author}{Touvron, H.}, \bibinfo{author}{Lavril, T.},
  \bibinfo{author}{Izacard, G.}, \bibinfo{author}{Martinet, X.},
  \bibinfo{author}{Lachaux, M.A.}, \bibinfo{author}{Lacroix, T.},
  \bibinfo{author}{Rozi{\`e}re, B.}, \bibinfo{author}{Goyal, N.},
  \bibinfo{author}{Hambro, E.}, \bibinfo{author}{Azhar, F.}, et~al.,
  \bibinfo{year}{2023}.
\newblock \bibinfo{title}{Llama: Open and efficient foundation language
  models}.
\newblock \bibinfo{journal}{arXiv preprint arXiv:2302.13971} .
\bibitem[{Twinanda et~al.(2016)Twinanda, Shehata, Mutter, Marescaux,
  De~Mathelin and Padoy}]{twinanda2016endonet}
\bibinfo{author}{Twinanda, A.P.}, \bibinfo{author}{Shehata, S.},
  \bibinfo{author}{Mutter, D.}, \bibinfo{author}{Marescaux, J.},
  \bibinfo{author}{De~Mathelin, M.}, \bibinfo{author}{Padoy, N.},
  \bibinfo{year}{2016}.
\newblock \bibinfo{title}{Endonet: a deep architecture for recognition tasks on
  laparoscopic videos}.
\newblock \bibinfo{journal}{IEEE transactions on medical imaging}
  \bibinfo{volume}{36}, \bibinfo{pages}{86--97}.
\bibitem[{Vondrick et~al.(2016)Vondrick, Pirsiavash and
  Torralba}]{vondrick2016anticipating}
\bibinfo{author}{Vondrick, C.}, \bibinfo{author}{Pirsiavash, H.},
  \bibinfo{author}{Torralba, A.}, \bibinfo{year}{2016}.
\newblock \bibinfo{title}{Anticipating visual representations from unlabeled
  video}, in: \bibinfo{booktitle}{Proceedings of the IEEE conference on
  computer vision and pattern recognition}, pp. \bibinfo{pages}{98--106}.
\bibitem[{Vondrick et~al.(2018)Vondrick, Shrivastava, Fathi, Guadarrama and
  Murphy}]{vondrick2018tracking}
\bibinfo{author}{Vondrick, C.}, \bibinfo{author}{Shrivastava, A.},
  \bibinfo{author}{Fathi, A.}, \bibinfo{author}{Guadarrama, S.},
  \bibinfo{author}{Murphy, K.}, \bibinfo{year}{2018}.
\newblock \bibinfo{title}{Tracking emerges by colorizing videos}, in:
  \bibinfo{booktitle}{Proceedings of the European conference on computer vision
  (ECCV)}, pp. \bibinfo{pages}{391--408}.
\bibitem[{Wang et~al.(2022)Wang, Long, Fan and Dou}]{wang2022neural}
\bibinfo{author}{Wang, Y.}, \bibinfo{author}{Long, Y.}, \bibinfo{author}{Fan,
  S.H.}, \bibinfo{author}{Dou, Q.}, \bibinfo{year}{2022}.
\newblock \bibinfo{title}{Neural rendering for stereo 3d reconstruction of
  deformable tissues in robotic surgery}, in: \bibinfo{booktitle}{Medical Image
  Computing and Computer Assisted Intervention--MICCAI 2022: 25th International
  Conference, Singapore, September 18--22, 2022, Proceedings, Part VII},
  \bibinfo{organization}{Springer}. pp. \bibinfo{pages}{431--441}.
\bibitem[{Ward et~al.(2021)Ward, Mascagni, Ban, Rosman, Padoy, Meireles and
  Hashimoto}]{ward2021computer}
\bibinfo{author}{Ward, T.M.}, \bibinfo{author}{Mascagni, P.},
  \bibinfo{author}{Ban, Y.}, \bibinfo{author}{Rosman, G.},
  \bibinfo{author}{Padoy, N.}, \bibinfo{author}{Meireles, O.},
  \bibinfo{author}{Hashimoto, D.A.}, \bibinfo{year}{2021}.
\newblock \bibinfo{title}{Computer vision in surgery}.
\newblock \bibinfo{journal}{Surgery} \bibinfo{volume}{169},
  \bibinfo{pages}{1253--1256}.
\bibitem[{Websurg(2023)}]{WebSurgt99:online}
\bibinfo{author}{Websurg}, \bibinfo{year}{2023}.
\newblock \bibinfo{title}{Websurg, the online university of ircad}.
\newblock \URLprefix \url{https://websurg.com/en/}.
\bibitem[{Wu et~al.(2014)Wu, Bondugula, Luisier, Zhuang and
  Natarajan}]{wu2014zero}
\bibinfo{author}{Wu, S.}, \bibinfo{author}{Bondugula, S.},
  \bibinfo{author}{Luisier, F.}, \bibinfo{author}{Zhuang, X.},
  \bibinfo{author}{Natarajan, P.}, \bibinfo{year}{2014}.
\newblock \bibinfo{title}{Zero-shot event detection using multi-modal fusion of
  weakly supervised concepts}, in: \bibinfo{booktitle}{Proceedings of the IEEE
  conference on computer vision and pattern recognition}, pp.
  \bibinfo{pages}{2665--2672}.
\bibitem[{Wu et~al.(2018)Wu, Xiong, Yu and Lin}]{wu2018unsupervised}
\bibinfo{author}{Wu, Z.}, \bibinfo{author}{Xiong, Y.}, \bibinfo{author}{Yu,
  S.X.}, \bibinfo{author}{Lin, D.}, \bibinfo{year}{2018}.
\newblock \bibinfo{title}{Unsupervised feature learning via non-parametric
  instance discrimination}, in: \bibinfo{booktitle}{Proceedings of the IEEE
  conference on computer vision and pattern recognition}, pp.
  \bibinfo{pages}{3733--3742}.
\bibitem[{Xie et~al.(2023)Xie, Sun, Xiong, Zheng, Zhao and Zhou}]{xie2023ra}
\bibinfo{author}{Xie, C.W.}, \bibinfo{author}{Sun, S.}, \bibinfo{author}{Xiong,
  X.}, \bibinfo{author}{Zheng, Y.}, \bibinfo{author}{Zhao, D.},
  \bibinfo{author}{Zhou, J.}, \bibinfo{year}{2023}.
\newblock \bibinfo{title}{Ra-clip: Retrieval augmented contrastive
  language-image pre-training}, in: \bibinfo{booktitle}{Proceedings of the
  IEEE/CVF Conference on Computer Vision and Pattern Recognition}, pp.
  \bibinfo{pages}{19265--19274}.
\bibitem[{Xu et~al.(2021a)Xu, Ghosh, Huang, Okhonko, Aghajanyan, Metze,
  Zettlemoyer and Feichtenhofer}]{xu2021videoclip}
\bibinfo{author}{Xu, H.}, \bibinfo{author}{Ghosh, G.}, \bibinfo{author}{Huang,
  P.Y.}, \bibinfo{author}{Okhonko, D.}, \bibinfo{author}{Aghajanyan, A.},
  \bibinfo{author}{Metze, F.}, \bibinfo{author}{Zettlemoyer, L.},
  \bibinfo{author}{Feichtenhofer, C.}, \bibinfo{year}{2021}a.
\newblock \bibinfo{title}{Videoclip: Contrastive pre-training for zero-shot
  video-text understanding}.
\newblock \bibinfo{journal}{arXiv preprint arXiv:2109.14084} .
\bibitem[{Xu et~al.(2015)Xu, Ba, Kiros, Cho, Courville, Salakhudinov, Zemel and
  Bengio}]{xu2015show}
\bibinfo{author}{Xu, K.}, \bibinfo{author}{Ba, J.}, \bibinfo{author}{Kiros,
  R.}, \bibinfo{author}{Cho, K.}, \bibinfo{author}{Courville, A.},
  \bibinfo{author}{Salakhudinov, R.}, \bibinfo{author}{Zemel, R.},
  \bibinfo{author}{Bengio, Y.}, \bibinfo{year}{2015}.
\newblock \bibinfo{title}{Show, attend and tell: Neural image caption
  generation with visual attention}, in: \bibinfo{booktitle}{International
  conference on machine learning}, \bibinfo{organization}{PMLR}. pp.
  \bibinfo{pages}{2048--2057}.
\bibitem[{Xu et~al.(2021b)Xu, Islam, Lim and Ren}]{xu2021class}
\bibinfo{author}{Xu, M.}, \bibinfo{author}{Islam, M.}, \bibinfo{author}{Lim,
  C.M.}, \bibinfo{author}{Ren, H.}, \bibinfo{year}{2021}b.
\newblock \bibinfo{title}{Class-incremental domain adaptation with smoothing
  and calibration for surgical report generation}, in:
  \bibinfo{booktitle}{Medical Image Computing and Computer Assisted
  Intervention--MICCAI 2021: 24th International Conference, Strasbourg, France,
  September 27--October 1, 2021, Proceedings, Part IV 24},
  \bibinfo{organization}{Springer}. pp. \bibinfo{pages}{269--278}.
\bibitem[{Xu et~al.(2022)Xu, Islam and Ren}]{xu2022rethinking}
\bibinfo{author}{Xu, M.}, \bibinfo{author}{Islam, M.}, \bibinfo{author}{Ren,
  H.}, \bibinfo{year}{2022}.
\newblock \bibinfo{title}{Rethinking surgical captioning: End-to-end
  window-based mlp transformer using patches}, in:
  \bibinfo{booktitle}{International Conference on Medical Image Computing and
  Computer-Assisted Intervention}, \bibinfo{organization}{Springer}. pp.
  \bibinfo{pages}{376--386}.
\bibitem[{Yengera et~al.(2018)Yengera, Mutter, Marescaux and
  Padoy}]{yengera2018less}
\bibinfo{author}{Yengera, G.}, \bibinfo{author}{Mutter, D.},
  \bibinfo{author}{Marescaux, J.}, \bibinfo{author}{Padoy, N.},
  \bibinfo{year}{2018}.
\newblock \bibinfo{title}{Less is more: Surgical phase recognition with less
  annotations through self-supervised pre-training of cnn-lstm networks}.
\newblock \bibinfo{journal}{arXiv preprint arXiv:1805.08569} .
\bibitem[{YouTube(2023)}]{1YouTube5:online}
\bibinfo{author}{YouTube}, \bibinfo{year}{2023}.
\newblock \bibinfo{title}{Youtube}.
\newblock \URLprefix \url{https://www.youtube.com/}.
\bibitem[{Yuan et~al.(2021)Yuan, Holden, Gao and Lee}]{yuan2021surgical}
\bibinfo{author}{Yuan, K.}, \bibinfo{author}{Holden, M.}, \bibinfo{author}{Gao,
  S.}, \bibinfo{author}{Lee, W.S.}, \bibinfo{year}{2021}.
\newblock \bibinfo{title}{Surgical workflow anticipation using instrument
  interaction}, in: \bibinfo{booktitle}{Medical Image Computing and Computer
  Assisted Intervention--MICCAI 2021: 24th International Conference,
  Strasbourg, France, September 27--October 1, 2021, Proceedings, Part IV 24},
  \bibinfo{organization}{Springer}. pp. \bibinfo{pages}{615--625}.
\bibitem[{Zellers et~al.(2021)Zellers, Lu, Hessel, Yu, Park, Cao, Farhadi and
  Choi}]{zellers2021merlot}
\bibinfo{author}{Zellers, R.}, \bibinfo{author}{Lu, X.},
  \bibinfo{author}{Hessel, J.}, \bibinfo{author}{Yu, Y.},
  \bibinfo{author}{Park, J.S.}, \bibinfo{author}{Cao, J.},
  \bibinfo{author}{Farhadi, A.}, \bibinfo{author}{Choi, Y.},
  \bibinfo{year}{2021}.
\newblock \bibinfo{title}{Merlot: Multimodal neural script knowledge models}.
\newblock \bibinfo{journal}{Advances in Neural Information Processing Systems}
  \bibinfo{volume}{34}, \bibinfo{pages}{23634--23651}.
\bibitem[{Zhukov et~al.(2019)Zhukov, Alayrac, Cinbis, Fouhey, Laptev and
  Sivic}]{zhukov2019cross}
\bibinfo{author}{Zhukov, D.}, \bibinfo{author}{Alayrac, J.B.},
  \bibinfo{author}{Cinbis, R.G.}, \bibinfo{author}{Fouhey, D.},
  \bibinfo{author}{Laptev, I.}, \bibinfo{author}{Sivic, J.},
  \bibinfo{year}{2019}.
\newblock \bibinfo{title}{Cross-task weakly supervised learning from
  instructional videos}, in: \bibinfo{booktitle}{Proceedings of the IEEE/CVF
  Conference on Computer Vision and Pattern Recognition}, pp.
  \bibinfo{pages}{3537--3545}.

\end{thebibliography}
\end{document}